\documentclass{article}

\usepackage{arxiv}
\usepackage{graphicx}%
\usepackage{multirow}%
\usepackage{amsmath,amssymb,amsfonts, mathtools}%
\usepackage{amsthm}%
\usepackage{mathrsfs}%
\usepackage{xcolor}%
\usepackage{textcomp}%
\usepackage{manyfoot}%
\usepackage{booktabs}%
\usepackage{algorithm}%
\usepackage{algorithmicx}%
\usepackage{algpseudocode}%
\usepackage{listings}%
\usepackage{xspace}
\usepackage{nicefrac}       %
\usepackage{microtype}      %
\usepackage{glossaries}
\usepackage[square,numbers]{natbib}
\renewcommand{\arraystretch}{1.5}
\usepackage{bookmark}
\usepackage{cleveref}
\usepackage{tikz}
\usepackage{setspace}

\hypersetup{hidelinks}

\crefname{figure}{Fig.}{Figs.}
\crefname{table}{Tab.}{Tabs.}
\crefname{appfig}{Supplementary Fig.}{Supplementary Figs.}
\crefname{apptab}{Supplementary Tab.}{Supplementary Tabs.}
\crefname{section}{Section}{Sections}
\crefname{equation}{Eq.}{Eqs}
\crefname{appendix}{Supplementary Note}{Supplementary Notes}
\glsdisablehyper %
\newacronym{ai}{AI}{Artificial Intelligence}
\newacronym{amax}{ActMax}{Activation Maximization}
\newacronym{cone}{$\boldsymbol{\vartheta}$}{semantic embedding}
\glsunset{cone}
\newacronym{pcone}{$\boldsymbol{\vartheta}_\text{probe}$}{probing embedding}
\glsunset{pcone}
\newacronym{ours}{{\includegraphics[scale=0.22]{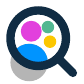}}\,\textsc{SemanticLens}}{\textsc{SemanticLens}}
\glsunset{ours}
\newacronym{ch}{CH}{Clever Hans}
\newacronym{crp}{CRP}{Concept Relevance Propagation}
\newacronym{cnn}{CNN}{Convolutional Neural Network}
\newacronym{llm}{LLM}{Large Language Model}
\newacronym{crc}{CRP}{Concept Relevance Propagation}
\newacronym{dl}{DL}{Deep Learning}
\newacronym{dnn}{DNN}{Deep Neural Network}
\newacronym{ga}{GA}{Gradient Ascent}
\newacronym{gan}{GAN}{Generative Adversarial Network}
\newacronym{hitl}{HITL}{Human in the Loop}
\newacronym{lrp}{LRP}{Layer-wise Relevance Propagation}
\newacronym{ml}{ML}{Machine Learning}
\newacronym{mlp}{MLP}{Multilayer Perceptron}
\newacronym{vit}{ViT}{Vision Transformer}
\newacronym{rmax}{RelMax}{Relevance Maximization}
\newacronym{rnn}{RNN}{Recurrent Neural Network}
\newacronym{tcav}{TCAV}{Testing With Activation Vectors}
\newacronym{xai}{XAI}{eXplainable Artificial Intelligence}
\newacronym{sae}{SAE}{Sparse Autoencoder}
\newacronym{invert}{\textsc{INVERT}}{Inverse Recognition}
\glsunset{invert}
\newacronym{cosy}{\textsc{CoSy}}{Concept Synthesis}
\glsunset{cosy}
\newacronym{imagenet}{ImageNet}{ImageNet}
\glsunset{imagenet}
\newacronym{clip-dissect}{{CLIP-Dissect}}{{CLIP-Dissect}}
\glsunset{clip-dissect}
\newacronym{broden}{{Broden}}{{Broad and Densely Labeled Dataset}}
\newacronym{paco}{{Paco}}{Parts and Attributes of Common Objects}
\newacronym{clip}{{CLIP}}{Contrastive Language-Image Pre-Training}
\newacronym{cav}{{CAV}}{Concept Activation Vector}

\newacronym{classspecific}{CS}{class-specific}
\newacronym{samplespecific}{SS}{sample-specific}
\newacronym{localized}{L}{localized}
\newacronym{examples}{E}{examples}
\newacronym{partially}{PP}{partially}
\newacronym{features}{F}{features}
\newacronym{architecture}{A}{architecture} 
\newacronym{dataNlabels}{D\&L}{data and labels}
\newacronym{training}{T}{training}

\newcommand{\mbf}[1]{\mathbf{#1}}
\newcommand{\ie}{{i.e.}\xspace}
\newcommand{\sm}{Supplementary Note}
\newcommand{\sms}{Supplementary Notes}

\newcommand{\wrt}{{wrt.}\xspace}
\newcommand{\eg}{{e.g.}\xspace}

\newcommand{\etc}{{etc.}\xspace}

\newcommand{\comment}[1]{}

\newcommand*\circled[1]{\tikz[baseline=(char.base)]{
            \node[shape=circle,draw,inner sep=1pt,fill=black] (char) {\color{white}\small{#1}};}}
\newcommand*\circledsmall[1]{\tikz[baseline=(char.base)]{
            \node[shape=circle,draw,inner sep=1pt,fill=black] (char) {\color{white}\footnotesize{#1}};}}
\newcommand*\circledw[1]{\tikz[baseline=(char.base)]{
            \node[shape=circle,draw,inner sep=1pt,fill=lightgray!60,color=lightgray!60] (char) {\color{black}\small{#1}};}}
\newcommand*\circledgraysmall[1]{\tikz[baseline=(char.base)]{
            \node[shape=circle,draw,inner sep=1pt,fill=lightgray!60,color=lightgray!60] (char) {\color{black}\footnotesize{#1}};}}
\newcommand*\circledrsmall[1]{\tikz[baseline=(char.base)]{
            \node[shape=circle,draw,inner sep=1pt,fill=red!60,color=red!60] (char) {\color{white}\footnotesize{#1}};}}

\DeclarePairedDelimiterX{\norm}[1]{\lVert}{\rVert}{#1}
\DeclarePairedDelimiterX{\skp}[1]{\langle}{\rangle}{#1}

\title{Mechanistic understanding and validation of large AI models with SemanticLens}

\author{%
Maximilian Dreyer$^{1}$\thanks{The authors contributed equally.} \quad Jim Berend$^{1*}$ \quad Tobias Labarta$^1$ \quad Johanna Vielhaben$^1$ \\ \textbf{Thomas Wiegand$^{1,2,3}$ \quad Sebastian Lapuschkin$^1$ \quad Wojciech Samek$^{1,2,3}$}\\
$^1$Fraunhofer Heinrich Hertz Institute \quad $^2$Technische Universität Berlin \\
$^3$BIFOLD -- Berlin Institute for the Foundations of Learning and Data\\
\texttt{\{wojciech.samek,sebastian.lapuschkin\}@hhi.fraunhofer.de}\\
}

\date{}

\renewcommand{\shorttitle}{Mechanistic understanding and validation of large AI models}

\hypersetup{
pdftitle={Mechanistic understanding and validation of large AI models with SemanticLens},
pdfsubject={q-bio.NC, q-bio.QM},
pdfauthor={Maximilian Dreyer, Jim Berend},
pdfkeywords={First keyword, Second keyword, More},
}

\begin{document}
\maketitle
\begin{spacing}{1.2}
\begin{abstract}
Unlike human-engineered systems such as aeroplanes, where each component's role and dependencies are well understood, the inner workings of AI models remain largely opaque, hindering verifiability and undermining trust. This paper introduces \gls{ours}, a universal  explanation method for neural networks that maps hidden knowledge encoded by components (\eg, individual neurons) into the semantically structured, multimodal space of a foundation model such as CLIP. In this space, unique operations become possible, including (\textbf{i}) textual search to identify neurons encoding specific concepts, (\textbf{ii}) systematic analysis and comparison of model representations, (\textbf{iii}) automated labelling of neurons and explanation of their functional roles, and (\textbf{iv}) audits to validate decision-making against requirements. Fully scalable and operating without human input, \gls{ours} is shown to be effective for debugging and validation, summarizing model knowledge, aligning reasoning with expectations (\eg, adherence to the ABCDE-rule in melanoma classification), and detecting components tied to spurious correlations and their associated training data. By enabling component-level understanding and validation, the proposed approach helps bridge the ``trust gap'' between AI models and traditional engineered systems.
We provide code for \gls{ours} on \url{https://github.com/jim-berend/semanticlens} and a demo on \url{https://semanticlens.hhi-research-insights.eu}.
\end{abstract}


\section{Introduction}\label{sec:introduction}
Technical systems designed by humans are constructed step by step, with each component serving a specific, well-understood function. For instance, an aeroplane’s wings and wheels have clear roles, and an edge detection algorithm applies defined signal processing steps like high-pass filtering. Such a construction by synthesis not only helps to understand the system’s overall behaviour, but also simplifies the validation of its safety. In contrast, neural networks are developed holistically through optimization, often using datasets of unprecedented scale. While this process yields models with impressive capabilities that increasingly outperform engineered systems, it has a major drawback: it does not provide semantic descriptions of each neuron’s function. Especially in high-stakes applications such as medicine or autonomous driving, the sole reliance on the output of the black-box AI model is often unacceptable as faulty or Clever Hans-type behaviours \cite{LapNCOMM19, kauffmann2024clever,borys2023explainable} may go unnoticed but have serious consequences. Recent regulations, such as the EU AI Act and the U.S. President’s Executive Order on AI underline the need for transparency and conformity assessment. What is urgently needed, therefore, is the ability to understand and validate the inner workings and individual components of AI models~\cite{tegmark2023provably,hernandez2017measure}, as we do for human-engineered systems. 

Despite progress in fields such as \gls{xai} \cite{SamXAI19,gunning2019xai} and mechanistic interpretability \cite{bricken2023towards}, the automated explanation and validation of model components at scale remains infeasible. Current approaches are limited in several ways: Firstly, they often strongly depend on human intervention \cite{miller2019explanation}, \eg, manual investigation of individual components~\cite{bereska2024mechanistic,ramaswamy2023overlooked} or predictions~\cite{friedrich2023typology}, preventing scaling to large modern architectures and datasets. Secondly, current explanatory methods focus mostly on isolated aspects of the model behaviour and lack a holistic perspective, \ie, do not enlighten the relations between the data, representation and prediction. It is, for example, not enough to only measure \emph{that} specific knowledge (\eg, a concept) has been learned~\cite{nguyen2019understanding,bau2017network}, but also necessary to understand \emph{how} it is actually used~\cite{achtibat2023attribution,fel2023craft,fel2024holistic,ahn2024www} and \emph{where} in the training dataset it is coming from \cite{koh2017understanding}. Further, available tools are suited to probe for expected concepts~\cite{kim2018interpretability}, but miss the part of a model that encodes for other unexpected concepts, which may interact with the former in non-trivial ways and thus influence model behaviour. Finally, methods that are applicable for ensuring compliance with legal/real-world requirements are scarce~\cite{li2024making,anwar2024foundational}. Holistic approaches are needed that quantify which parts of a model align with expectation and which not, thereby revealing spurious and potentially harmful components along with related training data. 

\begin{figure}[t]
    \centering
    \includegraphics[width=0.99\textwidth]{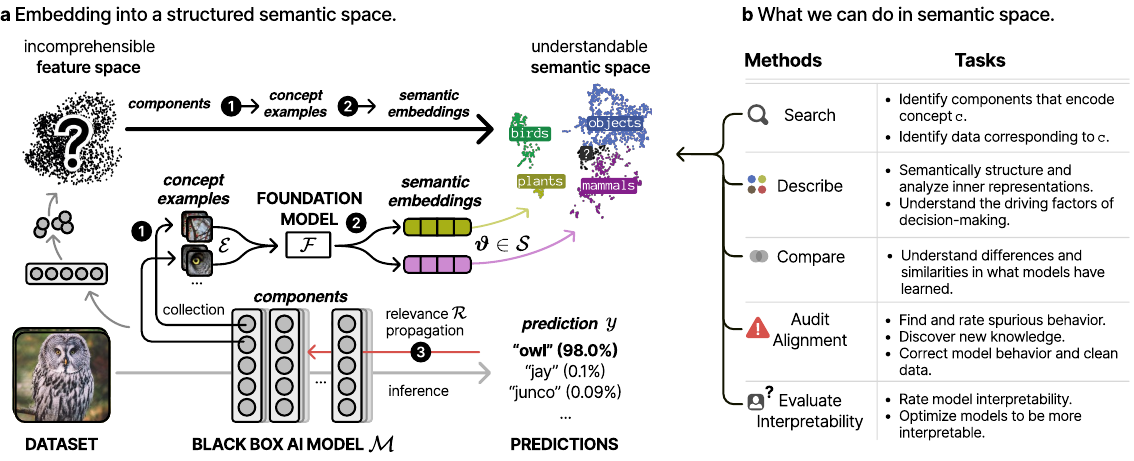}
    \caption{Embedding the model components in an understandable semantic space allows to systematically and more easily understand the inner workings of large neural networks.
    \textbf{a})
    In order to turn the incomprehensible latent feature space (hidden knowledge) into an understandable representation, we leverage a foundation model $\mathcal{F}$ that serves as a semantic expert. Concretely, for each component of the analysed model $\mathcal{M}$, \protect\circledsmall{1} concept examples $\mathcal{E}$ are extracted from the dataset, representing samples that induce high stimuli (\ie, activate the component), and \protect\circledsmall{2} embedded in the latent space of the foundation model resulting in a semantic representation $\boldsymbol{\vartheta}$.
    Further, \protect\circledsmall{3} relevance scores $\mathcal{R}$ for all components are collected, which illustrate their role in decision-making.
    \textbf{b})
    This new understandable model representation (\ie, set of $\boldsymbol{\vartheta}$'s, potentially linked to $\mathcal{E}$'s and $\mathcal{R}$'s) enables to systematically search, describe, structure, and compare internal knowledge of AI models. It further allows to audit alignment to human expectation and opens-up ways to evaluate and optimize human-interpretability.
    }
    \label{fig:introduction:overview}
\end{figure}

To address these shortcomings, we propose \gls{ours}, a novel method that embeds hidden knowledge encoded by individual components of an AI model into the semantically structured, multimodal space of a foundation model such as CLIP~\cite{radford2021learning}. Our approach not only enables understanding, but also allows measuring how knowledge is used for inference, which internal representations are encoding the knowledge, and which training data are relevant. This embedding is realized by two mappings:\\[+6px]
\circled{1} \textbf{\textit{components}} $\to$ \textbf{\textit{concept examples}}: For each component (\eg, neuron) of model $\mathcal{M}$, we collect a set of examples $\mathcal{E}$ (\eg, highly activating image patches) representing the concepts encoded by this component.  
\\[+6px]
\circled{2} \textbf{\textit{concept examples}} $\to$ \textbf{\textit{semantic space}}: We embed each set of examples $\mathcal{E}$ into the semantic space $\mathcal{S}$ of a multimodal foundation model $\mathcal{F}$ such as CLIP~\cite{radford2021learning}. As a result, each single component of model $\mathcal{M}$ is represented by a vector $\boldsymbol{\vartheta}\in\mathcal{S}$ in the semantic space of model $\mathcal{F}$.\\[+6px]
In addition, we use \gls{crp}~\cite{achtibat2023attribution} to identify relevant components and circuits for an individual model prediction, forming a third mapping:
\\[+6px]
\circled{3} \textbf{\textit{prediction}} $\to$ \textbf{\textit{components}}: Relevance scores $\mathcal{R}$ quantify the contributions of model components to individual predictions $\mbf{y} = \mathcal{M}(\mbf{x})$ on data points $\mbf{x}$.
\\[+6px]
By mutually connecting the model representation ($\mathcal{M}$), the relevant (training) data ($\mathcal{E})$, the semantic interpretation ($\mathcal{F}$) and the model prediction ($\mbf{y}$), \gls{ours} offers a holistic approach, which enables one to systematically analyse the internals of AI models and their prediction behaviours in a scalable manner without the need of having a human-in-the-loop \cite{de2024evaluating}, as illustrated in \cref{fig:introduction:overview}.

\noindent The multimodal foundation model $\mathcal{F}$ serves as a ``semantic expert'' for the data domain under consideration, effectively representing the model $\mathcal{M}$ as a comparable and searchable vector database, \ie, as a set of $\boldsymbol{\vartheta}$ (one vector $\boldsymbol{\vartheta}$ for every neuron), potentially linked to sets of $\mathcal{E}$ and $\mathcal{R}$. This enables new capabilities to answer questions about $\mathcal{M}$:
\\[+4px]
\textbf{Search} efficiently via text or other modalities for encoded knowledge, pinpointing corresponding components and data samples (see \cref{sec:results:explore:search,sec:app:exp:search}).
\\[+3px]
\textbf{Describe} at scale what concepts the model has learned, which are missing, and how it is using its knowledge for inference (see \cref{sec:results:explore:describe,sec:app:exp:describe}).
\\[+3px]
\textbf{Compare} learned concepts across models, varying architectures, or training procedures (see \cref{sec:results:explore:compare,app:exp:compare}).
\\[+3px]
\textbf{Audit alignment} of the model's encoded knowledge with expected human-defined concepts (see \cref{sec:results:audit,sec:results:medical,app:exp:audit}).
\\[+3px]
\textbf{Evaluate human-interpretability} of the hidden network components in terms of ``clarity'', ``polysemanticity'' and ``redundancy'' (see \cref{sec:results:evaluate,app:interpretability}).
\\[+4px]
More details and specific example questions are summarized in \cref{tab:questions}. Ultimately, the transformation of the model into a semantic representation, which not only reveals what and where knowledge is encoded but also connects it to the (training) data and decision-making, enables systematic validation and opens up new possibilities for more robust and trustworthy AI.
\crefname{appfig}{Suppl. Fig.}{Suppl. Figs.}
\crefname{apptab}{Suppl. Table}{Suppl. Tables}
\begin{table}[t!]
    \caption{Overview of questions which can be answered by \gls{ours}. The workflow to answer each question is provided in \cref{app:tab:questions_workflow}.}
    \centering 
    \begin{tabular}{l|p{10cm}|p{2.6cm}}
        Type & Question to the model $\mathcal{M}$ & Results\\
        \toprule
        \multirow{4}{*}{search} & \textit{``Has my model learned to encode a specific concept?''} via convenient  ``search-engine''-like text or image descriptions & \cref{fig:results:describe}a and \cref{fig:app:explore:search:examples:1,fig:app:explore:search:examples:2,fig:app:explore:search:comp_models} \\
         & \textit{``Which components have encoded a concept, how is it used, and which data is related?''}  & 
        \cref{fig:results:describe}d \\
        \hline
        \multirow{7}{*}{describe} & \textit{``What concepts has my model learned?''} in a structured, condensed and understandable manner via textual descriptions & \cref{fig:results:describe}b and \cref{fig:app:explore:describe:umap_resnet50v1,fig:app:explore:describe:umap_resnet50v2,fig:app:explore:describe:umap_resnet50_timm,fig:app:explore:describe:dissection,fig:app:explore:describe:dissection_ox}\\
         & \textit{``What and how are concepts contributing to a decision?''} by visualizing concept interactions throughout the model & \cref{fig:results:describe}c and \cref{fig:app:explore:describe:attribution_graph_resnet50v2_ox}\\
         & \textit{``What do I \emph{not} yet understand of my model?''}, offering to understand the unexpected concepts and their role for the model and origin in data & \cref{fig:results:describe}d and \cref{fig:app:explore:audit:audit_background_1,fig:app:explore:audit:audit_background_2,fig:app:explore:audit:audit_background_3,fig:app:explore:audit:audit_background_4,fig:app:explore:audit:audit_background_5,fig:app:explore:audit:audit_background_6,fig:app:explore:audit:audit_background_7,fig:app:explore:audit:audit_background_8}\\
        \hline
        \multirow{5}{*}{compare} & \textit{``What concepts are shared between two models, and which are unique to each one?''} by comparing learned concepts qualitatively and quantitatively & \cref{fig:app:explore:compare:compare_qualitatively,fig:app:explore:compare:quantitatively}\\
         & \textit{``How do my model's concepts change when changing the architecture or training?''} by comparing and tracking semantics of components & \cref{fig:app:explore:describe:dissection,fig:app:explore:compare:compare_qualitatively}\\
        \hline
        \multirow{2}{*}{audit} & \textit{``Is my model relying on valid information only?''} by separating learned concepts into \emph{valid}, \emph{spurious} or \emph{unexpected} knowledge & \cref{fig:results:audit,fig:results:medical,fig:app:explore:audit:audit}\\
        \hline
        \multirow{4}{*}{evaluate} & \textit{``How interpretable is my model?''} with easy to compute measures & \cref{fig:results:eval}b\\
         & \textit{``How can I improve interpretability of my model?''} by evaluating interpretability measures when changing model architecture or training procedure & \cref{fig:results:eval}c and \cref{tab:app:interptretability:optimizing_sparsity,tab:app:interptretability:optimizing_augmentation,tab:app:interptretability:optimizing_complexity,tab:app:interptretability:optimizing_drop_out,tab:app:interptretability:optimizing_numer_of_epochs}\\
        \bottomrule
    \end{tabular}
    \label{tab:questions}
\end{table}
\crefname{appfig}{Supplementary Fig.}{Supplementary Figs.}
\crefname{apptab}{Supplementary Table}{Supplementary Tables}

\section{Related Work}
\gls{ours} is a holistic framework that enables a systematic concept-level understanding of large AI models.
Its core elements rely on previous research advances related to concept visualization, labelling, attribution, comparison, discovery, and audits, as well as human-interpretability measures.

\paragraph{Feature Visualization}
To describe the role of individual components of a neural network, input images (referred to as ``concept examples'' in this work) are commonly sought that maximize their activation~\cite{zhou2015object, olah2017feature, radford2017learning, bau2017network,fel2023craft,yeh2020completeness}.
Concept examples can either be \emph{generated synthetically} using gradient-based approaches~\cite{mahendran2015understanding,olah2017feature,nguyen2016synthesizing,yin2020dreaming,fel2024unlocking} or diffusion-models~\cite{augustin2024dig}, or, alternatively, \emph{selected} from a test dataset by collecting neuron activations during predictions~\cite{fel2023craft,achtibat2023attribution,yeh2020completeness,bau2017network}. 
As synthetic concept examples often result in data samples that are out of the training data distribution,
we \emph{select} examples from the original test dataset.
Notably,
as multiple distracting features can be present in test data samples,
we further use \gls{crp}~\cite{achtibat2023attribution} to crop full samples to more meaningful image patches with less irrelevant features.

\paragraph{Concept Labelling}
Various methods are invested in labelling the concept of individual neurons,
which allows for easier interpretation of concepts and their corresponding examples, as well as quantitative evaluations.
One group of approaches is purely based on activation patterns,
such as Network Dissection~\cite{bau2017network} or INVERT~\cite{bykov2024labeling},
which require a large set of data annotations.
Notably,
CLIP-Dissect~\cite{oikarinen2022clip} circumvents the requirement for costly concept annotation by using a multimodal foundation model to generate soft data labels.
Other methods, such as ours, operate on the set of maximally activating images for a neuron,
hereby relying on other vision-language models~\cite{hernandez2021natural,oikarinen2022clip,ahn2024www,bai2024describe,kalibhat2023identifying}.

\paragraph{Concept Importance Scores}
To not only understand \emph{what} concepts have been learned, but also \emph{how} concepts are used,
importance scores \wrt predicted outputs (or upper-level component activations) can be computed.
Here,
various traditional feature attributions can be extended to compute importance scores of concepts~\cite{dreyer2023understanding,fel2024holistic}.
We adhere to the \gls{crp} framework for computing relevance scores of singular components or groups thereof.

\paragraph{Concept Comparison}
Various popular approaches exist that measure alignment between representation spaces of neural networks,
including Centered Kernel Alignment~\cite{Kornblith2019SimilarityON},
attention (map) patterns~\cite{Walmer2022TeachingMI, Raghu2021DoVT,Park2023WhatDS} or ``concept embeddings'' (\ie, weights for neuron activations to detect specific concepts in the data) as in Net2Vec~\cite{fong2018net2vec}.
The approaches above only provide a single scalar for the overall alignment between two representation spaces. In contrast,
other works (including ours) also enable for similarity analysis between single concepts,
allowing, \eg, to identify which concepts models share and in which concepts they differ.
Similarities between single concepts can be based on activation patterns~\cite{vielhaben2024beyond,bykov2023dora,li2024can}, relevance pattern~\cite{achtibat2023attribution} or concept example embeddings~\cite{park2023concept} as in \gls{ours}.

\paragraph{Concept Discovery}
Whereas early works show that neurons often encode for human-understandable concepts~\cite{bau2017network,olah2017feature},
other works argue that linear directions (or subspaces) in latent feature space are more interpretable and disentangled~\cite{fong2018net2vec,graziani2023uncovering,vielhaben2023multi}.
In fact,
neurons can be redundant and polysemantic (encoding for multiple concepts),
which directions might be less prone to~\cite{o2023disentangling,colin2024local}.
Recent research focuses on \glspl{sae}~\cite{huben2023sparse} or activation factorization~\cite{fel2024holistic} to receive more disentangled representations, for which, again, concept examples and concept relevance scores can be computed.
Whereas we focus in this work on the neural basis,
\gls{ours} is thus also applicable to \glspl{sae} or factorized activations.

\paragraph{Human-Interpretability Measures}
The work of Network Dissection
\cite{bau2017network} evaluates interpretability indirectly by the degree to which neurons align to a large set of expected concepts.
Later works leverage feature spaces of large models, where the concept examples of individual neurons are encoded.
Specifically, 
\cite{dreyer2024pure,foote2024tackling} introduce measures related to polysemanticity,
 \cite{kalibhat2023identifying,li2024evaluating,dreyer2024pure} related to clarity (or coherence),
and \cite{parekh2024concept} related to redundancy.
Recently,
measures to capture concept complexity have also been introduced~\cite{fel2024understanding}.
The semantic embedding of concept examples forms an integral part of \gls{ours} and allows us to provide a set of interpretability measures related to concept clarity, polysemanticity and redundancy in \cref{sec:methods:interpretability}.

\paragraph{Concept Audits}
Established methods for evaluating and auditing latent feature spaces of neural networks are
TCAV~\cite{kim2018interpretability} or linear probes~\cite{conneau2018you}.
Both are based on trying to detect a signal (linear direction) in the latent activations that can be associated with a specific user-defined concept of interest.
Whereas linear probes only detect that a certain concept is encoded by a model, 
TCAV also allows to quantify whether it is actually used~\cite{schrouff2021best}.
However,
the part of the model that is not covered by the (set of) expected concept(s) is not studied,
which could also incorporate various other spurious concepts.
\gls{ours} fills this gap and provides quantification of which concepts are valid, spurious, and not yet identified (unexpected).

\paragraph{Explanation Frameworks}
Instead of focusing on individual aspects, 
explanation frameworks combine multiple interpretability aspects and enable a more holistic understanding of model and data.
For example,
\gls{crp}~\cite{achtibat2023attribution} or CRAFT~\cite{fel2023craft} combine feature visualization and attribution,
but do not include labelling.
CLIP-Dissect~\cite{oikarinen2022clip} on the other hand,
leverages foundation models such as CLIP~\cite{radford2021learning} to label neurons,
but does not investigate how concepts are actually used during inference.
Based on the semantic embedding of model components,
\gls{ours} represents a more comprehensive and holistic framework compared to previous works that enables to systematically search, label, compare, describe and evaluate the inner mechanics of large AI models.
In \cref{app:related_works} we provide a detailed overview over other frameworks including NetDissect~\cite{bau2017network}, Net2Vec~\cite{fong2018net2vec}, TCAV~\cite{kim2018interpretability}, Summit~\cite{hohman2019s}, CLIP-Dissect~\cite{oikarinen2022clip}, CRP~\cite{achtibat2023attribution}, CRAFT \cite{fel2023craft}, PCX~\cite{dreyer2023understanding}, FALCON~\cite{kalibhat2023identifying}, ConceptEvo~\cite{park2023concept}, SpuFix~\cite{neuhaus2023spurious}, WWW~\cite{ahn2024www} and MAIA~\cite{shaham2024maia}.

\section{Methods}\label{sec:methods}

\gls{ours} embeds each component of a neural network into a semantic space. This embedding is realized in two steps as described in the following subsections.

\subsection{Describing the Role of Neurons via Concept Examples}
\label{sec:methods:examples}
To describe the role of a neuron, highly activating data samples are retrieved from the (training) database. Since the concept represented by the neuron can only occur in a small part of a large input sample, we facilitate the \gls{crp} framework \cite{achtibat2023attribution} to identify the relevant part of the input and crop each data sample to exclude input features with less than 1\,\% of the highest attribution value, as illustrated in \cref{fig:app:methods:methods_in_brief}a. For \glspl{vit} the \gls{crp} method is not available yet, therefore we approximate attributions by up-sampled spatial maps, as discussed in \cref{app:methods}. Thus, concept examples for neuron $k$ in layer $\ell$ are retrieved as
\begin{equation}
    \mathcal{E}_{k,\ell} = \left\{ 
    \operatorname{CRP} (\mathbf{x}): \mathbf{x}\in  \operatorname{top}_{m}(\mathcal{M}_k^\ell, \mathcal{D})  \right\}~,
\end{equation}
 where the latent activations at layer $\ell\in\{1,...,n\}$ with $k_{\ell}\in\mathbb{N}^+$ neurons are given by $\mathcal{M}^{\ell}\colon X\to Z^{\ell}\in \mathbb{R}^{k_{\ell}}$, $\operatorname{CRP}$ denotes the cropping operation, and $\operatorname{top}_m$ selects the $m$ maximally activating samples of dataset $\mathcal{D}\subset X$.

\subsection{Transformation into a Semantic Space}
\label{sec:methods:transformation}
In the second step, \gls{ours} generates a universal semantic representation for each model component based on the concept examples. To this end, we employ a foundation model $\mathcal{F}$ that serves as a semantic expert of the data domain, operating on the set of concept examples $\mathcal{E}_{k}$. As illustrated in \cref{fig:introduction:overview}a for step \circled{2}, we obtain the semantic representation of the $k$-th neuron in layer $\ell$ as a single vector $\boldsymbol{\vartheta}_{k}$ in the latent space $\mathcal{S}$ of foundation model $\mathcal{F}$ (index $\ell$ omitted for the sake of clarity):
\begin{equation}
    \boldsymbol{\vartheta}_k %
    \coloneqq 
    \mathbb{E}_{x\sim\mathcal{E}_k}\left[\mathcal{F}(x) \right]
    \approx
    \frac{1}{|\mathcal{E}_k| }\sum_{\mathbf{x} \in \mathcal{E}_k} \mathcal{F}(\mathbf{x}) \in \mathcal{S}\subseteq\mathbb{R}^d~.
\end{equation}
Computing the mean over individual feature vectors $\{\mathcal{F}(\mbf{x})\}_{\mbf{x}\in \mathcal{E}_k}$ (as also proposed in \cite{park2023concept}) is usually more meaningful than using individual vectors (\eg, for labelling as in \cref{sec:app:exp:describe}).
Averaging embeddings can be viewed as a smoothing operation, where noisy background signals are reduced, resulting in a better representation of the overall semantic meaning. Setting $|\mathcal{E}_k|=30$ results in converged $\boldsymbol{\vartheta}_k$ throughout ImageNet experiments, as detailed in \cref{sec:app:exp:describe}.

\paragraph{From Semantic Space to Model, Predictions and Data}
The semantic space representation is inherently connected to the model components, that are, themselves, linked to model predictions and the data, as illustrated in \cref{fig:app:methods:methods_in_brief}c. We can thus identify all neurons that correspond to a concept (via search as detailed in \cref{sec:methods:search}), filter this selection to the ones relevant for a decision output using \gls{crp} (via neuron-specific relevance scores $\mathcal{R}$ per data point, see step \circled{3} in \cref{fig:introduction:overview}a), and lastly, identify all data (\ie, $\mathcal{E}$) which highly activate the corresponding group of model components.

\subsection{Concept Search, Labelling and Comparison}
\label{sec:methods:search}
As semantic embeddings \gls{cone} are elements in a vector space, we measure similarity $s$ directly via cosine similarity, as is also the design choice of CLIP~\cite{radford2021learning}:
\begin{equation}
\label{eq:methods:similarity_measure}
    s_\text{cos}\colon \mathbb{R}^d\times \mathbb{R}^d\to [-1,1],~(x,y)\mapsto \frac{\skp{x, y}}{\norm{x}_2 \norm{y}_2}~.
\end{equation}\\
\textbf{Search:} Given a set of semantic embeddings of model components $\mathcal{V}_\mathcal{M}=\{\boldsymbol{\vartheta}_1,...,\boldsymbol{\vartheta}_k\}$ and an additional probing embedding  \gls{pcone} representing a sought-after concept, we can now \emph{search} for model components encoding the concept via 
\begin{equation}\label{eq:methods:concept_probing}
    \boldsymbol{\vartheta}^\ast =\operatornamewithlimits{argmax}_{\boldsymbol{\vartheta} \in\mathcal{V}_\mathcal{M}}\big\{
    s(\boldsymbol{\vartheta}_\text{probe},\boldsymbol{\vartheta}) - s(\boldsymbol{\vartheta}_{\texttt{<>}},\boldsymbol{\vartheta})\big\}~,
\end{equation}
where we additionally subtract the similarity to a ``null'' embedding $\boldsymbol{\vartheta}_{\texttt{<>}}$ representing background (noise) present in the concept examples if available. For text, \eg, it is common to subtract the embedding of the empty template to remove its influence~\cite{ahn2024www}, leading to more faithful labelling in \cref{sec:app:explore:describe:evaluation}.\\[+6px]
\textbf{Label:} In order to label the model representation $\mathcal{V}_\mathcal{M}$, a set of predefined concepts is embedded, resulting in $\mathcal{V}_\text{probe}\coloneqq \{\boldsymbol{\vartheta}^\text{probe}_1,...,\boldsymbol{\vartheta}^\text{probe}_l\}\subset\mathbb{R}^d$. Analogously to \cref{eq:methods:concept_probing} each neuron is assigned the most aligned label from the pre-defined set, or none if the similarity falls below a certain threshold.\\[+6px]
\textbf{Compare:}
Two models $\mathcal{N}$ and $\mathcal{M}$ may be quantitatively compared via the number of neurons that were assigned to concept labels as introduced by NetDissect~\cite{bau2017network}, or measuring set similarity $S_{{\mathcal{V}_\mathcal{M}} \rightarrow \mathcal{V}_\mathcal{N}}$ based on average maximal pairwise similarity:
\begin{equation}\label{eq:methods:concept_comparison}
    S_{{\mathcal{V}_\mathcal{M}} \rightarrow \mathcal{V}_\mathcal{N}} = \frac{1}{|\mathcal{V}_\mathcal{M}|}
    \sum_{\boldsymbol{\vartheta} \in\mathcal{V}_\mathcal{M}}\max_{\boldsymbol{\vartheta}' \in\mathcal{V}_\mathcal{N}}
    s(\boldsymbol{\vartheta},\boldsymbol{\vartheta}'),
\end{equation}
that quantifies the degree to which the knowledge (semantics) encoded in model $\mathcal{M}$ is also encoded in model $\mathcal{N}$. Notably, another means for comparison constitute the interpretability measures detailed in \cref{sec:methods:interpretability}. 

\subsection{Auditing Concept Alignment}
\label{sec:methods:audit}
As outlined in \cref{sec:results:audit}, it is important to measure how well the used concepts of a model are aligned with expected behaviour. In order to compute concept alignment, we require a set of model embeddings $\mathcal{V}_\mathcal{M}$, and a set of expected valid and spurious semantic embeddings $\mathcal{V}_\text{valid}$ and $\mathcal{V}_\text{spur}$, respectively. For each model component $k$, we then compute the alignment scores
\begin{align*}
    a_k^\text{valid} {=} \max_{\boldsymbol{\vartheta}_\text{v} \in\mathcal{V}_\text{valid}}\big\{
    s(\boldsymbol{\vartheta}_\text{v},\boldsymbol{\vartheta}_k) {-} s(\boldsymbol{\vartheta}_{\texttt{<>}},\boldsymbol{\vartheta}_k)\big\},
    \quad
    a_k^\text{spur} {=} \max_{\boldsymbol{\vartheta}_\text{s} \in\mathcal{V}_\text{spur}}\big\{
    s(\boldsymbol{\vartheta}_\text{s},\boldsymbol{\vartheta}_k) {-} s(\boldsymbol{\vartheta}_{\texttt{<>}},\boldsymbol{\vartheta}_k)\big\}.
\end{align*}
Additionally, it is important to take into account \emph{how} the components are used. We thus propose to retrieve the relevance of each model component during inference, \eg, the relevance for predictions of a specific class. Optimally, all relevant components are aligned to valid concepts only, \ie, $a^\text{valid}>0$ and $a^\text{spur}<0$. A high spurious alignment score $a^\text{spur}>0$ indicates potential harmful model behaviour. Neurons that aligned to neither should be examined more closely, representing unexpected concepts. 

\subsection{Human-Interpretability Measures for Concepts}
\label{sec:methods:interpretability}
We now introduce measures to capture the human-interpretability of concepts.

\subsubsection{Concept Clarity}
The \texttt{clarity} measure aims to represent how easy it is to understand the role of a model component, \ie, how easy it is to grasp the common theme of concept examples. Intuitively, clarity is low, when there is a lot of distracting (background) elements in the concept examples. Further, clarity is low when a concept is very abstract and many, at first glance, unrelated elements are shown throughout examples. To measure \texttt{clarity}, we compute semantic similarities in the set of concept examples, inspired by~\cite{dreyer2024pure,kalibhat2023identifying,li2024evaluating}. Cosine similarity serves here as a measure of how semantically similar two samples are in the latent space of the used foundation model. For the overall \texttt{clarity} score of neuron $k$, we compute the average pair-wise semantic similarity of the individual feature vectors $V_k = \left\{\mathbf{v}_{k, i}\right\}_i$:

\begin{align}
    I_{\texttt{clarity}}(V_k)
    &\coloneqq 
    \tfrac{1}{|V_k|(|V_k|-1)}
    \sum_{i=1}^{|V_k|} 
    \sum_{j\neq i} 
    s_{\text{cos}}(\mbf{v}_{k,i},\mbf{v}_{k,j})
    \label{eq:methods:efficient-clarity:line1}
    \\
    &= 
    \tfrac{|V_k|}{|V_k|{-}1}
    \Big(
    \big\|
    \tfrac{1}{|V_k|}
    \textstyle\sum_{i=1}^{|V_k|} \tfrac{\mathbf{v}_{k,i}}{\norm{\mathbf{v}_{k,i}}_2}\big\|^2_2 
    - 
     \frac{1}{|V_k|}
    \Big) \in [-\tfrac{1}{|V_k|-1}, 1] \label{eq:methods:efficient-clarity}
\end{align}
where the last expression is a formulation that is computationally less expensive, and circumvents the need to compute large similarity matrices.

\subsubsection{Concept Similarity and Redundancy}
The semantic representation allows conducting comparisons across arbitrary sets of neurons without being restricted to neurons from identical layers or model architectures. In particular, it allows us to assess the degree of \texttt{similarity} between the concepts of two neurons $k$ and $j$, which we define as 

\begin{equation}
I_\texttt{sim}(\boldsymbol{\vartheta}_k, \boldsymbol{\vartheta}_j)\coloneqq
    s_\text{cos}(\boldsymbol{\vartheta}_k, \boldsymbol{\vartheta}_j) =  \frac{\langle \boldsymbol{\vartheta}_k, \boldsymbol{\vartheta}_j \rangle}{\|\boldsymbol{\vartheta}_k\|_2 \|\boldsymbol{\vartheta}_j\|_2} \in [-1,1]
\end{equation}
via cosine similarity. Based on similarity, we can further assess the degree of \texttt{redundancy} across the concepts of $m$ neurons with the \gls{cone} set $\mathcal{V} = \{\boldsymbol{\vartheta}_1, \dots, \boldsymbol{\vartheta}_m \}$ which we define as 

\begin{equation}
    I_{\texttt{red}}(\mathcal{V}) \coloneqq
    \frac{1}{m} \sum_{k=1}^m \max_{j\neq k} \left\{ I_\texttt{sim}(\boldsymbol{\vartheta}_k, \boldsymbol{\vartheta}_j) \right\} \in [-1,1]~.
\end{equation}

Notably, \emph{semantic} redundancy might not imply \emph{functional} redundancy. Even though two semantics are similar, they might correspond to different input features. For example, the stripes of a zebra or striped boarfish are semantically similar, but might be functionally very different for a model that discerns both animals.

\subsubsection{Concept Polysemanticity}
A neuron is considered polysemantic if multiple semantic directions exist in the concept example set. Formally, we define a neuron as polysemantic if subsets of $\mathcal{E}_k$ can be identified that provide diverging \glspl{cone}. The \texttt{polysemanticity} measure is defined as

\begin{align}
I_{\texttt{poly}}({V}_k^{(1)}{,}{...}{,}{V}_k^{(h)}) 
    &\coloneqq 1 - I_\texttt{clarity}\Big(\Big\{ 
     \textstyle\sum_{\mbf{v}\in {V}_k^{(i)}} \mbf{v} \mid i=1,...,h
     \Big\}\Big)~,
\end{align}
where ${V}^{(i)}_k \subseteq {V}_k$ for $i=1{,...},h$ is a subset of the embedded concept examples, generated by an off-the-shelf clustering method, where we use $h=2$ throughout experiments. Alternatively, as proposed by \cite{dreyer2024pure}, polysemanticity can be measured as an increase in the clarity of each set of concept examples, which, however, performs worse in the user study evaluation as detailed in \cref{app:interpretability:study}.

\section{Results}
\label{sec:results}
We begin in \cref{sec:results:explore} with demonstrating how to understand the internal knowledge of AI models by searching and describing the semantic space. These functionalities provide the basis for effectively auditing alignment of the model's reasoning \wrt human-expectation in \cref{sec:results:audit}. We demonstrate how to spot flaws in medical models and improve robustness and safety in \cref{sec:results:medical}. Lastly, computable measures for human-interpretability of model components are introduced, enabling to rate and improve interpretability at scale in \cref{sec:results:evaluate}. 

The different sets of experiments reported in this paper were conducted on a variety of models, including convolutional neural networks with ResNet and VGG architectures as well as different \glspl{vit}. Additionally, we used two large vision datasets, namely ImageNet \cite{deng2009imagenet} and ISIC 2019 \cite{tschandl2018ham10000}, along with several foundation models, including Mobile-CLIP~\cite{vasu2024mobileclip}, DINOv2~\cite{oquab2023dinov2} and WhyLesionCLIP~\cite{yang2024textbook}. Further details about the experimental setting can be found in \cref{app:experimental-settings}. Additional analyses are reported in \cref{sec:app:exp:search,sec:app:exp:describe,app:exp:compare,app:exp:audit,app:interpretability}.

\subsection{Understanding the Inner Knowledge of AI Models}
\label{sec:results:explore}

In the following, \gls{ours} is used to systematically analyse the knowledge encoded by neural network components of ResNet50v2~\cite{he2016deep} trained on the ImageNet classification task \cite{deng2009imagenet}. The individual components of the model are embedded as vectors \gls{cone} into the multimodal and semantically organized space of the Mobile-CLIP foundation model~\cite{vasu2024mobileclip}, as illustrated in \cref{fig:introduction:overview} and described in \cref{sec:methods}.

\subsubsection{Search: Finding the Needle in the Haystack}
\label{sec:results:explore:search}

The first capability of \gls{ours} that we demonstrate is its \emph{search} capability, allowing one to quickly browse through all neurons of the ResNet50v2 model and identify concepts that a user is interested in, such as potential biases (\eg, gender or racial), data artefacts (\eg, watermarks) or specific knowledge. The search is based on (cosine) similarity comparison between a probing vector \gls{pcone}, representing the concept we are looking for (\eg, the concept \texttt{person}), and the set of embedded neurons (\ie, \gls{cone}'s) of the ResNet model. The shared vision-text embedding space of Mobile-CLIP allows us to query concepts described through images (image of a person) as well as concepts described by text (textual input ``person''). More details about the creation of the probing vectors and the retrieval process can be found in \cref{sec:methods}.

\begin{figure}[t]
    \centering
    \includegraphics[width=0.95\textwidth]{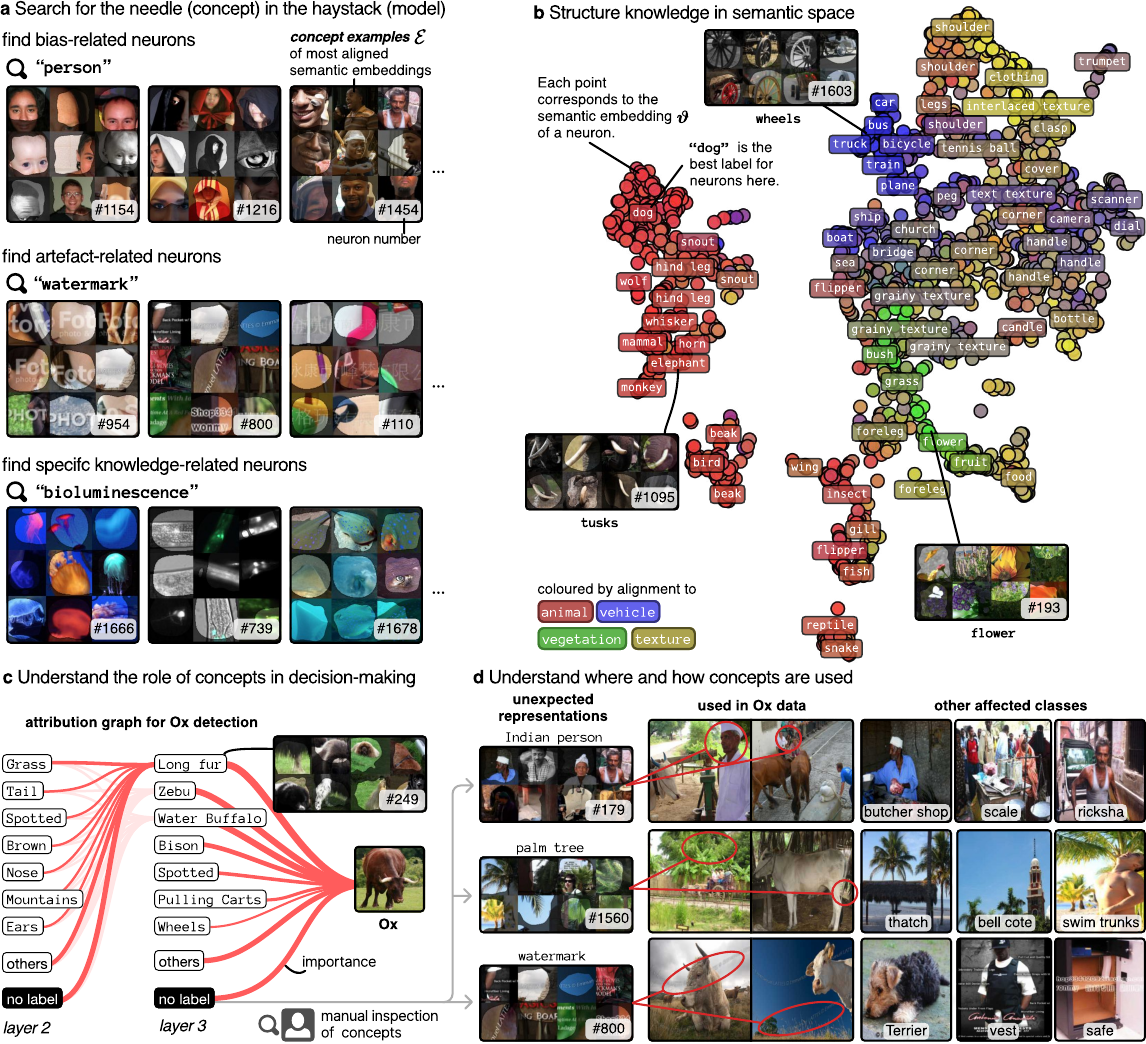}
    \caption{
    \gls{ours} allows to systematically understand the internal knowledge and inference of neural networks. 
    \textbf{a}) Via search engine-like queries, one can probe for knowledge referring to, \eg, (racial) biases, data artefacts, or specific knowledge of interest. 
    \textbf{b}) A low-dimensional UMAP projection of the semantic embeddings provides a structured overview of the model's knowledge, where each point corresponds to the encoded concept of a model component. By searching for human-defined concepts, we can add descriptions to all parts of the semantic space.
    \textbf{c}) Having grouped the knowledge into concepts,
    attribution graphs reveal where  concepts are encoded in the model and how they are utilized (and interconnected) for inference. For predicting Ox, we learn that ox-cart related background concepts are used. Importantly, we can also identify relevant knowledge that could not be labelled, and should be manually inspected by the user.
    \textbf{d}) The set of unexpected concepts includes \texttt{Indian person}, \texttt{palm tree}, and \texttt{watermark} concepts, which correlate in the dataset with Ox. We can further find other affected output classes, \eg, ``butcher shop'', ``scale'' and ``ricksha'' for the \texttt{Indian person} concept.
    }
    \label{fig:results:describe}
\end{figure}
As illustrated in \cref{fig:results:describe}a, 
neurons of the ResNet50v2 model can be identified that encode for \texttt{person}-related concepts. Two embedded neurons, which are most similar to the probing vector represent different, non-obvious and potentially discriminative aspects of a person, such as ``hijab'' (neuron \texttt{\#1216}) and ``dark skin'' (neuron \texttt{\#1454}). It is in principle a valid strategy to represent different object subgroups sharing certain visual features by specialized neurons. However, if these ``sensitive attribute''-encoding neurons are used for other purposes, \eg, the ``dark skin''-\texttt{person} neuron is used for classification of ``steel drum'' (see \cref{fig:results:audit}b), then this may hint at potential fairness issues. 

We also query the model for the concept \texttt{watermark}. The retrieved neurons encode watermarks and other text superimposed on an image. Such data artefacts may become part of the model's prediction strategy, known as shortcut learning \cite{friedrich2023typology,kuhn2025efficient} or Clever Hans phenomenon \cite{achtibat2023attribution}, and massively undermine its trustworthiness (\ie, the model predicts right but for the wrong reason \cite{stammer2021right}). While previous works have unmasked such \texttt{watermark}-encoding neurons more or less by chance \cite{achtibat2023attribution,pahde2023reveal}, \gls{ours} allows one to intentionally query the model for the presence of such neurons.

In addition to searching for bias- or artefact-related neurons, we can also query the model for specific knowledge, \eg, the concept \texttt{bioluminescence}. The results show that this concept has been learned by the ResNet50v2 model. Such specific knowledge queries can help ensure that the model has learned all the relevant concepts needed to solve a task, as demonstrated in the ABCDE-rule for melanoma detection in Section \ref{sec:results:audit}. Notably, \gls{ours} not only allows to query the model for specific concepts, but also to further identify the output classes for which concepts are used and the respective (training) data, as later shown in \cref{fig:results:describe}d. Additional examples, comparisons between models, and details are provided in \cref{sec:app:exp:search}.

\subsubsection{Describe: What Knowledge Exists and How Is It Used?}
\label{sec:results:explore:describe}

Another feature of \gls{ours} is its ability to \emph{describe} and
systematically analyse \emph{what} knowledge the model has learned and \emph{how} it is used. \cref{fig:results:describe}b provides an overview of the ResNet50v2 model's internal knowledge (penultimate layer components) as a UMAP projection of the semantic embeddings \gls{cone}. Here, \eg, searching for \texttt{animal} results in aligned embeddings on the left (indicated by red colour), whereas \texttt{transport}-related embeddings are located in the centre (blue coloured). Even more insights can be gained when systematically searching and annotating semantic embeddings, as described in the following.

\paragraph{Labelling and Categorizing Knowledge}
To structure the learned knowledge systematically, we assign a text-form concept label (from a user-defined set) to a neuron embedding if its alignment exceeds the alignment with a baseline which is an empty text label. The labelled embeddings can then be grouped according to their annotation, \eg, all embeddings matching \texttt{dog} are grouped together, which reduces complexity, especially if many neurons with similar semantic embeddings exist. In fact, the ResNet results in over a hundred neurons related to \texttt{dog}, as illustrated in \cref{fig:results:describe}b, where the overall top-aligned label from the expected set for clusters of semantic embeddings \gls{cone} are provided. Further details (including labels) and examples are provided in \cref{app:explore:describe:labels,app:exp:describe:umap}, respectively.

It is further possible to ``dissect''~\cite{bau2017network} a model's knowledge at different levels of complexity, ranging from broad categories such as ``objects'' and ``animals'' to more fine-grained concepts such as ``bicycle'' or ``elephant''. For instance, in \cref{sec:app:exp:describe}, we categorize the model components relevant to the ``Ox'' class into ``breeds'' like \texttt{Water Buffalo}, ``work''-related concepts such as \texttt{ploughing}, and ``physical attributes'' such as \texttt{horns}. Importantly, labelling not only facilitates the assessment of what the model has learned but also identifies gaps in its knowledge, \ie, cases where no neuron aligns with a user-defined concept. In the studied ResNet model, for instance, no neuron encodes the Ox breeds \texttt{Angus} and \texttt{Hereford}, indicating areas where additional training data could enhance model performance. Notably, faithfulness of labels is important~\cite{kopf2024cosy}, which is evaluated in \cref{sec:app:explore:describe:evaluation}.

\paragraph{Understanding How Knowledge is Used}
Understanding \emph{how} the model uses the learned knowledge is as crucial as knowing \emph{what} knowledge exists. For example, while \texttt{wheels} can be a valid concept to detect sports cars, it should not be relevant for detecting an Ox, which is, however, measurable for the ResNet. \cref{fig:results:describe}c shows the attribution graph for the class Ox. The graph is constructed from the conditioned relevance scores computed with \gls{crp}~\cite{achtibat2023attribution} and reveals associations between neuron groups with the same concept-label. For the class Ox, the attribution graph in \cref{fig:results:describe}c, \eg, reveals next to the \texttt{wheel} concept another highly relevant \texttt{long fur} concept encoded by neuron \texttt{\#179} in \textit{layer 3}, which in turn relies in the next lower-level layer on a \texttt{grass} concept, indicating that neuron \texttt{\#179} is encoding long-furred animals on green grass. Attribution graphs thus not only describe \emph{what} and \emph{how} concepts are used, but also enhance our understanding of sub-graphs (``circuits'') within the model. A full attribution graph is detailed in \cref{app:explore:describe:attribution_graph}.

\paragraph{The Link Between Knowledge, Data and Predictions}
Notably, some components did not align with any of the pre-defined text-based concepts, yielding embedding similarities that were equal to or lower than those obtained using an empty text prompt as a baseline. As shown in \cref{fig:results:describe}d, manual inspection of these unexpected concepts reveals associations to \texttt{Indian person}, \texttt{palm tree} and \texttt{watermark}, traced to neurons \texttt{\#179}, \texttt{\#1569} and \texttt{\#800} in \textit{layer 3}, respectively. All three concepts correspond to spurious correlations in the dataset, \eg, farmers using Ox to plough a field, palm trees in the background or a watermark overlaid over images, where the responsible training data can be generally identified by retrieving highly activating samples $\mathcal{E}$. The plot further shows other ImageNet classes for which the neurons are highly relevant. Affected classes include ``butcher shop'', ``scale'', and ``rickshaw'' for \texttt{Indian person}; ``thatch'', ``bell cote'', and ``swim trunk'' for \texttt{palm tree}; and ``Lakeland Terrier'', ``bulletproof vest'', and ``safe'' for \texttt{watermark}. By inherently connecting data, model components, and predictions, \gls{ours} constitutes an effective and actionable tool for model debugging, further described in \cref{sec:results:medical}.

\subsubsection{Compare: Identify Common and Unique Knowledge}
\label{sec:results:explore:compare}
So far, we have investigated a \emph{single} model in semantic space. However, the semantic space serves as a unified space, where \emph{multiple} models of different architectures, different layers or model parts can be embedded and compared. As such, the influence on learned concepts when changing the network architecture or training hyperparameters, such as the training duration, can be studied. 

In \cref{app:exp:compare} two ResNet50 models trained on ImageNet, where one (ResNet50v2) is trained more extensively and results in higher test accuracy, are compared using
\gls{ours}. As illustrated in \cref{fig:app:explore:compare:compare_qualitatively}, both models share common knowledge, \eg, bird-related concepts. However, whereas the better trained ResNet50v2 has learned more specific concepts, \eg, specific fur textures of dogs, the other has learned more abstract concepts that are shared throughout classes. For the dog breed ``Komondor'' which has a white mop-like coat, for example, the ResNet50 has learned a mop-like concept that is used to detect ``Komondor'' as well as ``mop'', whereas the ResNet50v2 learned a class-specific concept. This is in line with works that study generalization of neural networks for long training regimes, observing that latent model components become more structured and class-specific~\cite{liu2022towards}. We further provide quantitative comparisons via network dissection in \cref{app:explore:describe:dissection}.
Alternatively, \gls{ours} allows to compare models also quantitatively \emph{without} access to concept-labels by evaluating the similarity between the models' knowledge. In \cref{app:exp:compare}, we discuss the alignment of various pre-trained neural networks across layers and architectures.

\subsection{Audit Alignment: Do Models Reason as Expected?}
\label{sec:results:audit}
\begin{figure}[t]
    \centering
    \includegraphics[width=0.99\textwidth]{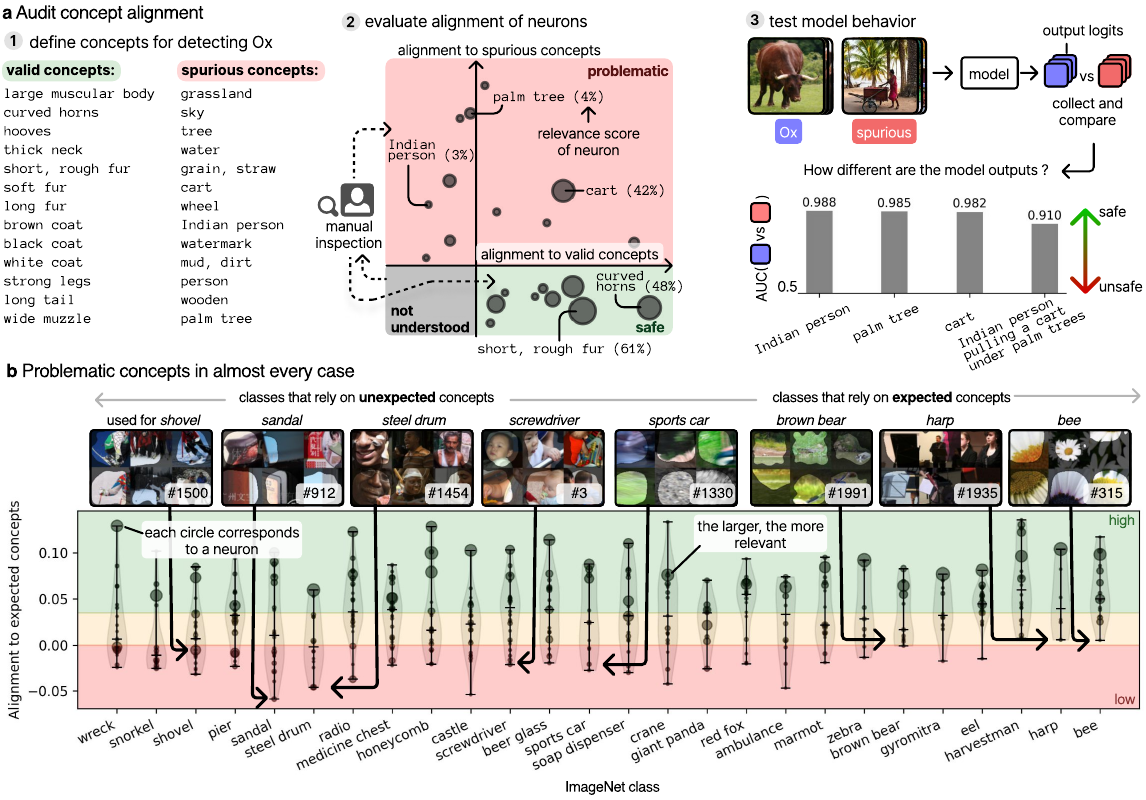}
    \caption{Using \gls{ours} to audit models and check if their reasoning aligns with human expectation.
    \textbf{a}) \protect\circledgraysmall{1} In a first step, a set of valid and spurious concepts  is defined via text descriptions, \eg, \texttt{curved horns} or \texttt{palm tree} for ``Ox'' detection, respectively. 
    \protect\circledgraysmall{2} Afterwards,
    we check which model components encode for either spurious or valid concepts, both or neither.
    The size of each dot in the chart represents the importance of a component for ``Ox'' detections.
    We learn, that the ResNet50v2 relies on \texttt{Indian person}, \texttt{palm tree} and \texttt{cart} concepts.
    \protect\circledgraysmall{3} Lastly,
    we can test our model, and try to distinguish the ``Ox'' output logits on ``Ox'' images (from the test dataset) and diffusion-based images with spurious features only.
    When multiple spurious features are present, as for \texttt{Indian person pulling a cart under palm trees},
    model outputs become more difficult to separate, indicated by a lower AUC score.
    \textbf{b})
    When auditing the ResNet's alignment to valid concepts for 26 ImageNet classes, we find that in \emph{all} cases,
    spurious or background concepts are used.
    }
    \label{fig:results:audit}
\end{figure}

The analyses introduced in \cref{sec:results:explore} enable the quantification of a model's alignment with human expectations. Specifically, they allow assessment of a model's reliance on valid, spurious, or unexpected concepts. The steps of an alignment audit, outlined in \cref{fig:results:audit}a, include \circledw{1} defining concepts, \circledw{2} evaluating concept alignment, and \circledw{3} testing model behaviour.

\circledw{1} \textbf{Defining a set of expected concepts:} First, a set of valid and spurious concepts is defined, utilized to compare against the concepts actually employed by the model. For illustration, we revisit the Ox example where valid concepts include \texttt{curved horns}, \texttt{wide muzzle} and \texttt{large muscular body}, as shown in \cref{fig:results:audit}a (\emph{left}). On the other hand, we are also aware of spurious correlations, such as \texttt{palm tree}, \texttt{Indian person} and \texttt{watermark}. Notably, all of these concepts can be defined within the modality of the model's data domain (\ie, via example images), or, as demonstrated here, simply via text-prompts when utilizing a multimodal foundation model such as CLIP for concept encoding.

\circledw{2} \textbf{Evaluating alignment to valid and spurious concepts:} The alignment of the model's knowledge with user-defined spurious or valid concepts is visualized in the scatter plot in \cref{fig:results:audit}a (\emph{middle}) for ``Ox'' detection. Concretely, we calculate the maximum alignment between an embedding \gls{cone} and all probing embeddings \gls{pcone} within a set (valid or spurious), with mathematical formulations detailed in \cref{sec:methods:audit}. Each dot in the plot represents a neuron of the penultimate layer, with its size indicating its highest importance (shown in parentheses) during inference on the test set.

Several spurious concepts such as \texttt{palm tree}, \texttt{Indian person} or \texttt{cart} are identified besides valid concepts such as \texttt{short, rough fur} or \texttt{curved horns}. Notably, neurons that do not align to any user-defined concept can be manually inspected as done in \cref{fig:results:describe}d, and incorporated into the set of spurious or valid concepts. As discussed for a VGG~\cite{simonyan2014very} model in \cref{app:exp:audit}, lower overall alignment scores can also result for neurons that encode for highly abstract concepts, or that exhibit ``polysemantic'' behaviour, encoding for multiple concepts simultaneously.

\circledw{3} \textbf{Testing models for spurious behaviour:} While \gls{ours} enables quantification of a model's reliance on valid or spurious features (\eg, via the share of spuriously aligned components), it is equally important to assess the actual impact of identified spurious features on inference. Here we use a model test~\cite{neuhaus2023spurious} evaluating the separability of two sets of outputs: one generated from images containing valid features (associated with the ``Ox'' class) and the other from images with spurious features, as illustrated in \cref{fig:results:audit}a (\emph{right}). When testing the model on images (generated with Stable Diffusion) for a single concept (\texttt{Indian person}, \texttt{palm tree} or \texttt{cart}), the model output logits for ``Ox'' are clearly distinguishable from those attained from ``Ox'' images, achieving AUC scores above $0.98$. However, when multiple spurious features are presented simultaneously, and we test the model on images combining all three concepts, the ``Ox''  output logits are further amplified. Specifically, the ``Ox'' class ranks among the top-5 predictions in over half of the spurious samples, resulting in an AUC of 0.91,
as further detailed in \cref{app:exp:audit}.

\paragraph{Problematic concept reliance everywhere}
The previous example highlights the presence of unexpected spurious correlations, such as the association of palm trees with ``Ox''. Expanding on this, we evaluate the alignment of model components with valid concepts across 26 additional ImageNet classes, including ``shovel'', ``steel drum'' and ``screwdriver''. \cref{fig:results:audit}b presents the resulting highest alignment scores with a valid concept for each model component, where size again indicates relevance for ``Ox''. Remarkably, no class shows complete alignment of all relevant model components with valid concepts. In every case, spurious or background features are relevant, including \texttt{snow} for ``shovel'', \texttt{Afro-American person} for ``steel drum'', and \texttt{child} for ``screwdriver''. A comprehensive overview over the utilized concepts by the model is provided in \cref{app:exp:audit}.

\paragraph{Unaligned models are often challenging to interpret}
When analysing popular pre-trained models on the ImageNet dataset, we observe strong variations \wrt their alignment to valid concepts. The reason often lies in the share of knowledge that is neither aligned to valid or spurious concepts, as demonstrated for the VGG-16 in \cref{app:exp:audit}. For instance, the VGG-16 contains several polysemantic components that perform multiple roles in decision-making, which generally reduces alignment. On the other hand, more performant and wider models tend to have more specialized (\eg, class-specific) and monosemantic model components, later quantified in \cref{sec:results:evaluate}. Overall, higher-performing models with larger feature spaces (more neurons per layer) show thus greater alignment scores throughout experiments detailed in \cref{app:exp:audit}. Interpretability and trustworthiness are closely tied, underscoring the importance of optimizing models for interpretability.

\subsection{Towards Robust and Safe Medical Models}
\label{sec:results:medical}
\begin{figure}[t]
    \centering
    \includegraphics[width=0.99\textwidth]{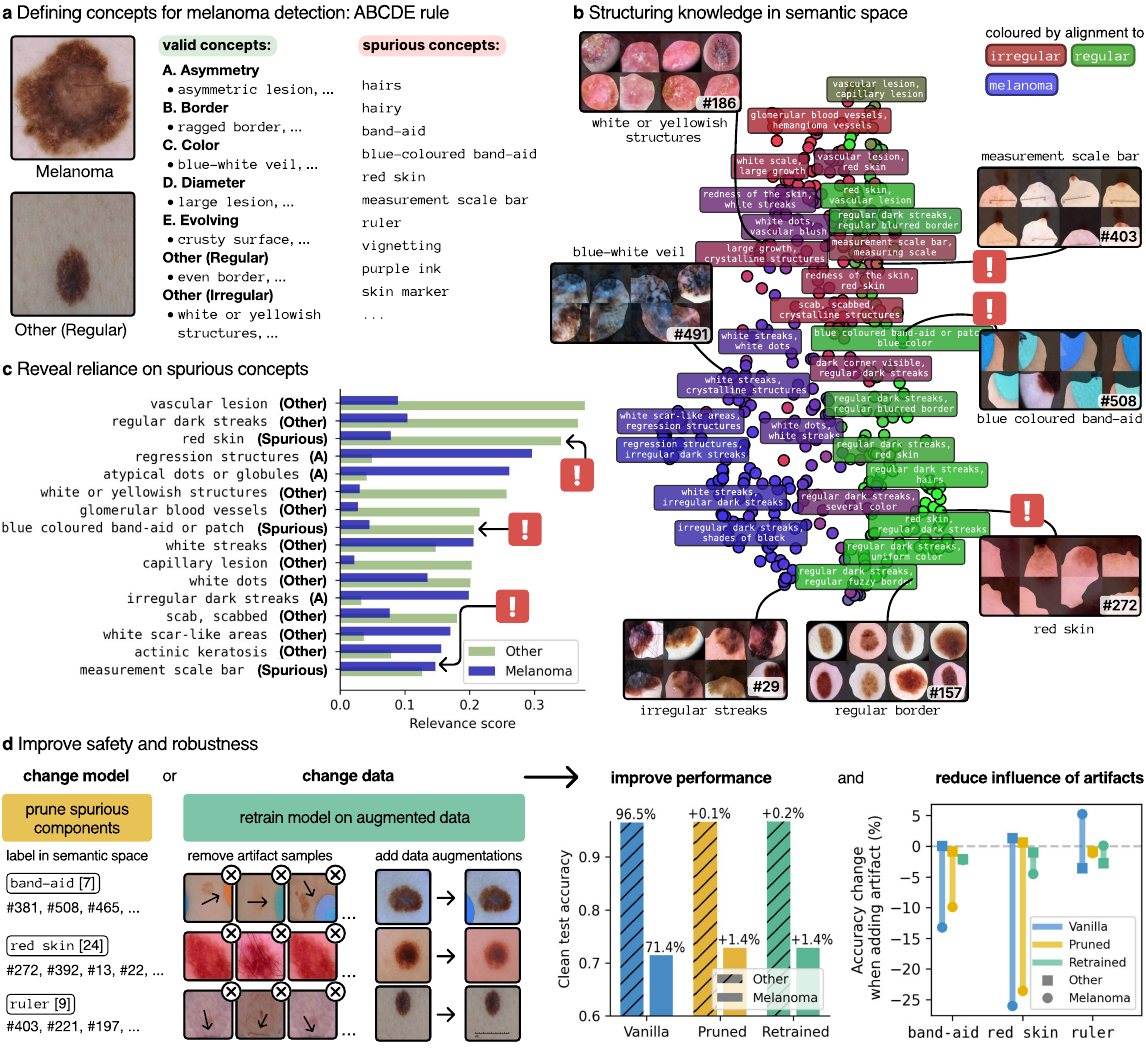}
    \caption{
    Using \gls{ours} to find and correct bugs in medical models that detect melanoma skin cancer.
    \textbf{a})
    The ABCDE-rule is a popular guide for visual melanoma clues. We expect models to learn several concepts corresponding to the ABCDE-rule, as well as other melanoma-unrelated indications (such as \texttt{regular border}) or spurious concepts, including \texttt{hairs} or \texttt{band aid}.
    \textbf{b})
    In semantic space visualized via a UMAP projection, we can identify valid concepts, such as \texttt{blue white veil} for ``melanoma'', but also spurious ones such as \texttt{red skin} or \texttt{ruler}.
    \textbf{c})
    When investigating the importance of concepts, we find that \texttt{red skin} or \texttt{band-aid} concepts are strongly used for the ``other'' (non-melanoma) class. Also \texttt{ruler} concepts are used with slightly higher relevance for ``melanoma''.
    \textbf{d})
    We can improve safety and robustness of our model by either changing the model and remove spurious components, or retrain the model on augmented data. Whereas both approaches lead to improved clean performance, the influence of artefacts is only significantly reduced via re-training.
    }
    \label{fig:results:medical}
\end{figure}

One of the most popular medical use cases for AI is melanoma detection in dermoscopic images, as shown in \cref{fig:results:medical}a. In the following, we demonstrate how to debug a VGG-16 model with \gls{ours} that is trained to discern melanoma from other irregular or benign (referred to as ``other'') samples in a public benchmark dataset~\cite{tschandl2018ham10000,codella2018skin,hernandez2024bcn20000}.

\subsubsection{ABCDE-Rule for Melanoma Detection}
Dermatologists have created guidelines for visual melanoma detection, such as the ABCDE-rule, short for Asymmetry, Border, Colour, Diameter and Evolving \cite{duarte2021-md}. We will use \gls{ours} to check whether the model has learned concepts regarding the ABCDE-rule, such as \texttt{asymmetric lesion} (A), \texttt{ragged border} (B), \texttt{blue-white veil} (C), \texttt{large lesion} (D), and \texttt{crusty surface} (E). In addition, we also define concepts for benign and other skin diseases as well as several spurious concepts that have been reported in previous works~\cite{cassidy2022analysis,kim2024transparent}, corresponding to hairs, band-aids, red-hued skin, rulers, vignetting and skin markings. Please refer to \cref{sec:app:medical:labels} for a full list of concepts.

\subsubsection{Finding Bugs in Medical Models}
To embed the VGG's components into a semantic space, we leverage a recently introduced CLIP model trained on skin lesion data \cite{yang2024textbook}. As shown in \cref{fig:results:medical}b, the semantic embeddings are structured, with concepts aligning to \texttt{irregular} in the top (red colour), \texttt{melanoma} in the bottom left (blue colour), and \texttt{regular} in the bottom right (green colour). Here, we can identify several valid concepts such as \texttt{blue-white veil} and \texttt{irregular streaks} for detecting melanoma, and \texttt{regular border} for benign samples. On the other hand, spurious model components are also revealed, such as neuron \texttt{\#403} encoding for \texttt{measurement scale bar}, \texttt{\#508} for \texttt{blue coloured band-aid}, and \texttt{\#272} for \texttt{red skin} (visually red-coloured skin).

To quantify \emph{how} concepts are used by the model, we compute their highest importance for predicting the ``melanoma'' or ``other'' class using \gls{crp} on the test set, as shown in \cref{fig:results:medical}c. Alarmingly, we find the previously found spurious concepts to be highly relevant: \texttt{red skin} and \texttt{blue-coloured band-aid} are strongly used for ``other'', whereas \texttt{measurement scale bar} is slightly stronger used for ``melanoma''.

\subsubsection{Model Correction and Evaluation}
In application, the background features of red-coloured skin, plasters and rulers should not influence a detection. \gls{ours} helps identifying model components and data associated with spurious concepts. To debug the model \cite{spinner2024innspector}, we apply two approaches, namely pruning without retraining and retraining on augmented data. For pruning, we label corresponding neurons, resulting in overall 40 out of 512 neurons in the penultimate layer that are pruned. On the other hand, we remove data samples that incorporate the artefacts, identified through studying the highly activating samples of our labelled components. In order to become insensitive towards the artefacts, we randomly augment data samples during training by overlaying hand-crafted artefacts, as illustrated in \cref{fig:results:medical}d (\emph{left}).

The results in \cref{fig:results:medical}d (right) show that both strategies, pruning and retraining, lead to increased accuracy on a clean test set (without artefact samples), especially for melanoma (from 71.4\,\% to 72.8\,\%). We further ``poison'' data with artificially inserted artefacts by cropping out ruler and plasters from real test samples and inserting them as an overlay into clean test samples as done in~\cite{pahde2023reveal}, or, for \texttt{red skin}, add a reddish hue, as detailed in \cref{sec:app:medical:evaluation}. Interestingly, the pruned model decreases artefact sensitivity only slightly, still remaining highly sensitive. When adding red colour, for example, test accuracy still drops by over 20\,\% for non-melanoma samples for the pruned model. Although computationally more expensive, only retraining leads to a strong reduction in artefact sensitivity. Further details and discussions are provided in \cref{sec:app:medical:evaluation}.

\subsection{Evaluating Human-Interpretability of Model Components}
\label{sec:results:evaluate}
Deciphering the meaning of concept examples $\mathcal{E}$ can be particularly challenging, especially when neurons are polysemantic and encode for multiple concepts, as observed in \cref{sec:results:audit}. We introduce a set of easily computable measures that assess how ``clear'', ``similar'' and ``polysemantic'' concepts are perceived by humans, as inferred from their concept examples $\mathcal{E}$. Additionally, we introduce a measure to quantify the ``redundancies'' present within a set of concepts. All measures are based on evaluating similarities of concept examples $\mathcal{E}$ in semantic space $\mathcal{S}$, with mathematical definitions given in \cref{sec:methods:interpretability}.

\begin{figure}[t]
    \centering
    \includegraphics[width=0.99\textwidth]{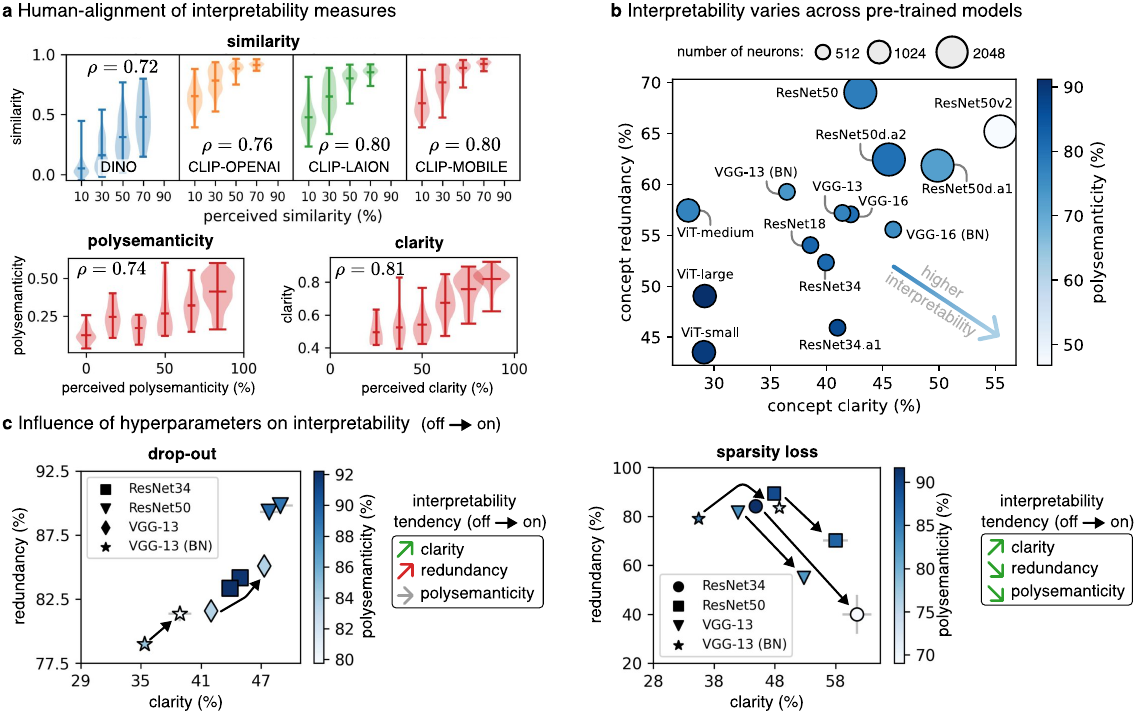}
    \caption{We introduce computable human-interpretability measures that are useful to rate and improve model interpretability: ``clarity'' for how clear and easy it is to understand the common theme of concept examples, ``polysemanticity'' describes if multiple distinct semantics are present in the concept examples, ``similarity'' for the similarity of concepts,
    and ``redundancy'' which describes the degree of redundancies in a set of concepts.
    \textbf{a})
    Our computable measures align with human perception in user studies, resulting in correlation scores above 0.74. Generally, more recent and performant foundation models lead to higher correlation scores.
    \textbf{b})
    Interpretability differs strongly for common pre-trained models. Usually, \glspl{vit} or smaller and less performant convolutional models show lower interpretability.
    \textbf{c})
    We can optimize model interpretability \wrt hyperparameter choices, such as drop-out or activation sparsity regularization during training. Whereas drop-out leads to more redundancies besides improved clarity of concepts, applying a sparsity loss improves interpretability overall.
    }
    \label{fig:results:eval}
\end{figure}

\subsubsection{Alignment of Interpretability Measures with Human Perception}
Aiming to assess \emph{human}-interpretability, we first evaluate the alignment between human judgments and our proposed measures (similarity, clarity and polysemanticity) through user studies.  Specifically, we recruited over 218 participants via Amazon Mechanical Turk to engage in 15-minute tasks. In these studies, participants were presented with concept examples drawn from the ImageNet object detection task, similar to those shown in~\cref{sec:results:explore,sec:results:audit}. For each interpretability measure, we designed an independent study consisting of both qualitative and quantitative experiments. Further details regarding the study design, the models used, and the data filtering procedures can be found in \cref{app:interpretability:study}.
    
All in all, we obtain a high alignment between computed measures and human perception, indicated by high correlation scores above 0.74, as shown in \cref{fig:results:eval}a, which recent works using textual concept examples also reflect~\cite{li2024evaluating}. Regarding concept similarity, human-alignment varies across foundation models, namely the DINOv2~\cite{oquab2023dinov2} (uni-modal), CLIP-OpenAI~\cite{radford2021learning}, CLIP-LAION~\cite{schuhmann2022laion}, and the most recent CLIP-Mobile~\cite{vasu2024mobileclip} (specific variants reported in \cref{app:interpretability:study}). Our results indicate that more recent and more performant CLIP models are also more aligned with human perception. Other hyperparameter choices such as the used similarity measure are compared in \cref{sec:methods}. We further performed an odd-one-out task, where participants are asked to detect the outlier concept (of three concepts). Interestingly, our measures often outperform the human participants, indicating that computational measures can even be more reliable than humans. Participants from Amazon Mechanical Turk are, however, often motivated to complete studies quickly to maximize their pay rate, which may not result in optimal performance.

\subsubsection{Rating and Improving Interpretability}
\label{sec:results:evaluate:optimize}
The difficulty of understanding the role of components in standard pre-trained models can vary strongly, as, \eg, previously observed in \cref{sec:results:explore,sec:results:audit}. This is further confirmed by evaluating various popular neural networks trained on ImageNet using our newly introduced measures for penultimate layer neurons, detailed in \cref{fig:results:eval}b. Larger and broader models, such as ResNet101, show higher degrees of redundancy, which can be expected as more neurons per layer allow more redundancies to form, \eg, in order to increase robustness. For narrow models such as the ResNet18, on the other hand, the effective neural basis might be too small, leading to superimposed signals and a higher polysemanticity (neurons are more likely to fulfil multiple tasks)~\cite{elhage2022superposition}.
    
The convolution-based ResNet architecture shows higher concept clarity compared to the more recent transformer-based \gls{vit}. Whereas the ResNet consists of ReLU non-linearities that allow to associate a high neuronal activation with a specific active input pattern, \glspl{vit} often refrain from ReLUs, which enables to superimpose signals (concepts) throughout model components, ultimately leading to high polysemanticity and low interpretability~\cite{scherlis2022polysemanticity}. Interestingly, recent efforts are being made in \gls{llm} interpretability to extend the transformer architecture post-hoc with \glspl{sae} based on ReLUs to again receive a more interpretable neuronal basis~\cite{huben2023sparse}. Moreover, our analysis shows that more extensively trained models have clearer and overall more interpretable components, as is the case for the ResNet50v2 compared to the ResNet50. This observation raises the question on how we can influence training parameters to gain higher latent interpretability, which we inspect in the following:\\[+4px]
\emph{Drop-out:} Drop-out regularization is effective for reducing overfitting, preventing high reliance on few features by randomly setting a fraction of component activations to zero during training. Our results shown in \cref{fig:results:eval}c indicate that VGG-13 model components become more redundant, but also clearer when drop-out is applied during training on a subset of ImageNet (standard error given by gray error bars for eight runs each). It can be expected that more redundancies form, as redundancies make predictions more robust when components are randomly pruned. On the other hand, neurons are measured to become more class-specific and thus clearer. Notably, architectures might react differently in terms of interpretability, as indicated by the ResNet-34 and ResNet-50 which are not strongly affected by drop-out. Qualitative examples of concepts, detailed training procedures and results are provided in \cref{app:interpretability:training}.\\[+4px]
\emph{Sparsity regularization:} Secondly, we apply L1 sparsity regularization during training on the neuron activations, as is, \eg, common for \glspl{sae}. Our experiments indicate that sparsity regularization improves interpretability in all measured aspects, resulting in more specific, less polysemantic and semantically redundant neurons. We further investigate the effect of task complexity, number of training epochs and data augmentation on latent interpretability in \cref{app:interpretability:training}.

\section{Discussion}\label{sec:discussion}

With \gls{ours}, we propose to transfer components of large machine learning models into an understandable semantic representation that allows one to understand and evaluate their inner workings in a holistic manner. This transfer is made possible through recent foundation models that serve as domain experts, taking the human out of interpretation loops, that otherwise would be cognitively infeasible to process due to the sheer amount of components of modern deep neural networks. Especially useful are \emph{multimodal} foundation models that allow to search, annotate and label network components via textual descriptions. Foundation models improve constantly, becoming more efficient and applicable in scarcer data domains such as medical data, or other data modalities including audio and video~\cite{laionclap2023,xu2021videoclip}. 

These new capabilities offered by \gls{ours} allow to comprehensively audit the internal components of AI models. A multitude of spurious behaviours of popular pre-trained models are hereby revealed, stressing the need to understand every part of a model in order to ensure fairness, safety and robustness in application. To understand and audit models, we are dependent on the interpretability of the model components themselves. While some models demonstrate higher interpretability, progress is still needed to develop truly interpretable models, especially regarding recent transformer architectures. However, post-hoc architecture modifications or training regularizations are promising ongoing endeavours to achieve also high interpretability in modern architectures. Our newly introduced human-interpretability measures are an effective tool for optimizing and understanding model architecture choices without relying on expensive user studies for evaluation. There are still many other hyperparameters that we leave for future work, including training with pretrained models, adversarial training, weight decay regularization, and \glspl{sae}. 

Trust and safety go hand in hand with a verification of the internal components, as is the case with traditional engineered systems such as aeroplanes.  In order to close this ``trust gap'', holistic approaches such as \gls{ours} are needed, that allow to understand and quantify the validity of latent components, as well as offer ways to increase their interpretability and reduce potential spurious behaviours. However, various future work remains with post-hoc component-level \gls{xai} approaches such as \gls{ours}, including the need for further, meaningful evaluation metrics~\cite{nauta2023anecdotal}, application to generative models \cite{amara2024challenges}, and potential limitations regarding ``post-hoc'' vs. ``ante-hoc'' interpretability \cite{rudin2019stop}, leaving enough room for innovation by the next generation of \gls{xai} researchers \cite{longo2024explainable}.

\subsection*{Code Availability}
We provide an open-source toolbox for the scientific community written in Python and based on PyTorch \cite{paszke2019pytorch}, Zennit-CRP~\cite{achtibat2023zennit} and Zennit \cite{anders2021software}. The GitHub repository containing our implementations of \gls{ours} is publicly available on \url{https://github.com/jim-berend/semanticlens}. All experiments were conducted with Python 3.10.12, zennit-crp 0.6, Zennit 0.4.6 and PyTorch 2.2.2.

\subsection*{Acknowledgements}
We would like to express our gratitude to Oleg Hein for his work and fruitful discussions for developing a public demo of \gls{ours} on \url{https://semanticlens.hhi-research-insights.eu}.

\end{spacing}
\bigskip

\newpage

\bibliographystyle{unsrtnat}

\bibliography{arxiv}

\newpage

\begin{appendix}
\appendix
\counterwithout{figure}{section}
\setcounter{figure}{0}
\setcounter{table}{0}\setcounter{equation}{0}
\renewcommand{\figurename}{Supplementary Fig.}
\renewcommand{\tablename}{Supplementary Tab.}
\crefalias{figure}{appfig}
\crefalias{table}{apptab}
\renewcommand{\thefigure}{\thesection.\arabic{figure}}
\renewcommand{\thetable}{\thesection.\arabic{table}}
\section*{Supplementary Materials}

This article has supplementary files providing additional details and information, descriptions, experiments and figures. 
\cref{app:related_works} offers a detailed survey of related work important to our contribution, contrasting several related explainability techniques to our proposed technical contribution regarding different criteria. 
\cref{app:experimental-settings} includes a description of  the datasets and models used in our experiments.
\sms~\ref{sec:app:exp:search},~\ref{sec:app:exp:describe} and~\ref{app:exp:compare}
include additional experiments and explanations for the search, describe and compare functionalities of \gls{ours}.
Subsequently,
\cref{app:exp:audit} provide additional details for auditing and debugging models with \gls{ours}.
In \cref{app:interpretability}, details on the user study of \cref{sec:results:evaluate} and more experiments for optimizing latent interpretability are provided.
\cref{app:methods} describes our technical contribution, the \gls{ours}, in increased detail and provides additional background. 
The proposed interpretability measures are detailed in \cref{app:methods:interpretability}.
In \sm~\ref{sec:app:methods:workflow} we summarize the computational steps involved in answering the questions presented in \cref{tab:questions}.
Current challenges and an outlook to future work are discussed in \cref{app:limitations}.

\section{Extended Related Work}
\label{app:related_works}

\definecolor{OliveGreen}{rgb}{0.23,0.49,0.15} %
\newcommand{\yes}{{\color{OliveGreen}\checkmark}} %
\newcommand{\kinda}{{\color{orange}$\bigcirc$}}
\newcommand{\redx}{{\color{red}$\boldsymbol \times$}}
\newcommand{\softx}{{\color{orange}$\boldsymbol \times$}}
\gls{ours} is a holistic framework that enables a systematic concept-level understanding of large AI models.
Its core elements rely on previous research advances related to concept visualization, labelling, attribution, comparison, discovery, audits, and human-interpretability measures, as detailed in the following.
In \cref{app:tab:related_work:overview},
we compare \gls{ours} with other popular \gls{xai} frameworks.

\paragraph{Concept Examples (Feature Visualization)}
Most feature visualization techniques rely on maximizing activation values of single neurons or a linear combination thereof~\cite{zhou2015object, olah2017feature, radford2017learning, bau2017network,fel2023craft,yeh2020completeness}, where in its simplest form, input images are sought that produce the highest activation value of a specific unit.
In this work, 
the set of images is referred to as ``concept examples''.
Concept examples can be generated synthetically using gradient ascent, or alternatively found from a sample dataset by collecting neuron activations during predictions. 
Regarding synthetic examples, preventing the emergence of adversarial patterns became a main research area. 
Several priors were proposed to guide optimization into more realistic looking images~\cite{mahendran2015understanding,olah2017feature,nguyen2016synthesizing,yin2020dreaming,fel2024unlocking}.
Recently,
diffusion models are being applied to also help in generating more realistic concept examples~\cite{augustin2024dig}.

Alternatively,
natural concept examples can be collected on the training or test data,
where it is favourable to collect patches of the input data~\cite{fel2023craft,achtibat2023attribution,yeh2020completeness,bau2017network}, as whole inputs can incorporate a lot of distracting background features.
We follow the \gls{crp} approach and crop full data samples to the actual relevant part using neuron-specific attributions~\cite{achtibat2023attribution}.
Other approaches facilitate upsampled spatial activation maps~\cite{bau2017network}, that are only available for convolutional layers, or transformers (through spatial token information).

\paragraph{Encoding Concepts of Neurons: Activation Pattern or Feature Space}
There are two approaches in literature to encode the concept of neurons: (1) via activation patterns~\cite{bau2017network,bykov2023dora,oikarinen2022clip} on data with concept annotations (\eg, binary labels or segmentation mask) or (2) by embedding concept examples into another feature space~\cite{bai2024describe,park2023concept,ahn2024www,kalibhat2023identifying}.
Activation patterns are a very direct measure, but often only correspond to a singular (pooled) activation score per data point.
Usually, data points incorporate multiple features, which can lead to wrong conclusions due to unexpected correlations when working with singular activation scores.
For example,
two neurons that encode for \texttt{nose} and \texttt{eyes} will activate very similarly for data with human faces, but encode for different concepts.
It is thus important to have a qualitative and meaningful set of concept data.
Alternatively,
concept examples (cropped to the relevant part, see \cref{sec:methods})  aim to communicate the semantic role of neurons more directly.
Then, 
in order to encode a concept,
the concept examples are embedded in the feature space of a model: either the same model~\cite{park2023concept} or a foundation model~\cite{bai2024describe,ahn2024www}.
Notably, generating the concept examples and encodings is algorithmically and computationally more involved compared to activation pattern computation.
Whereas using the same model for encoding is convenient as it does not require a foundation model (that might need to be trained first),
the latent space of the investigated might not be as semantically structured.
The work of \cite{vielhaben2024beyond} shows that self-supervised foundation models have a more semantically structured latent space than models trained on a classification task.
Especially multimodal foundation models are interesting as they allow to also interact and describe the embedding space more flexible.
However,
it is to note that describing concepts through concept examples is assuming that the concept is precisely and well-defined via these examples,
which might not always be the case~\cite{kalibhat2023identifying}.

\paragraph{Neuron Labelling}
Various methods are invested in labelling the concept which an individual neuron represents.
Some are purely based on activation patterns,
such as Network Dissection~\cite{bau2017network} or \gls{invert}~\cite{bykov2024labeling},
which require a large set of data annotations.
Notably,
CLIP-Dissect~\cite{oikarinen2022clip} circumvents the requirement for costly concept annotation by using a multimodal foundation model for annotation.
Other methods, such as ours, operate on the set of maximally activating images (concept examples) for a neuron,
hereby relying on other vision-language models~\cite{hernandez2021natural,oikarinen2022clip,ahn2024www,bai2024describe,kalibhat2023identifying}.

\paragraph{Concept Importance Scores}
In order to understand \emph{how} concepts or components are used, we need to compute their importance during inference \wrt the output or other components.
Here,
various traditional feature attributions can be used to compute importance scores of latent representations~\cite{dreyer2023understanding,fel2024holistic}.
We adhere to the \gls{crp} framework for computing relevance scores of singular components (or groups thereof) \wrt to the output prediction and/or specific model parts, further detailed in \cref{app:methods}.

\paragraph{Concept Discovery}
Whereas early works show that neurons often encode for human-understandable concepts~\cite{bau2017network,olah2017feature},
other works argue that linear directions (or subspaces) in latent feature space are more interpretable and disentangled~\cite{fong2018net2vec,graziani2023uncovering,vielhaben2023multi}.
In fact,
neurons can be redundant and polysemantic (encoding for multiple concepts),
which directions might be less prone to~\cite{o2023disentangling,colin2024local}.
Recent research focuses on \glspl{sae}~\cite{huben2023sparse} or activation factorization~\cite{fel2024holistic} to receive more disentangled representations, for which, again, concept examples and concept relevance scores can be computed.
Whereas we focus in this work on the neural basis,
\gls{ours} is thus also applicable to \glspl{sae} or factorized activations.

\paragraph{Concept Comparison in Models}
Various popular approaches exist that measure alignment between representation spaces of neural networks,
including Centered Kernel Alignment~\cite{Kornblith2019SimilarityON},
attention (map) patterns~\cite{Walmer2022TeachingMI, Raghu2021DoVT,Park2023WhatDS} or ``concept embeddings'' (\ie, weights for neuron activations to detect specific concepts in data) as in Net2Vec~\cite{fong2018net2vec}.
The approaches above only provide a single scalar value for the overall alignment between two representation spaces. In contrast,
other works (including ours) also enable for similarity analysis between single concepts,
allowing, \eg, to identify which concepts models share and in which concepts they differ.
Similarities between concepts can be based on activation patterns~\cite{vielhaben2024beyond,bykov2023dora,li2024can}, relevance pattern~\cite{achtibat2023attribution} or concept example embeddings~\cite{park2023concept} as in \gls{ours}.

\paragraph{Concept-level Audits}
Established methods for evaluating and auditing latent feature spaces of neural networks are
TCAV~\cite{kim2018interpretability} or linear probes~\cite{conneau2018you}.
Both are based on trying to detect a signal (linear direction) in the latent activations that can be associated with a specific user-defined concept of interest.
Contrary to \gls{ours},
where a description of a concept is given through a set of concept examples (in the form of images or text),
TCAV, \eg, requires additionally a set of negative examples without the concept.
Originally,
linear probes only detect that a certain concept is encoded by a model,
but not \emph{how} it used or how relevant it is.
TCAV uses latent gradients collected on a dataset to estimate the sensitivity of the model \wrt a concept.
However,
sensitivity does not fully reflect the degree to which a concept contributes during inference,
as the contribution also depends on the concept activation (magnitude).
The work of \cite{schrouff2021best} extends TCAV to also gain information in terms of concept importances for local predictions.
Further,
the part of the model that is not covered by the (set of) expected concept(s) is not studied,
which could also incorporate various other spurious concepts.

Popular methods to evaluate model behaviour \wrt model outputs are, besides test set performance and worst-group accuracies on subsets of the test set~\cite{pahde2023reveal,friedrich2023typology},
also other more direct measures evaluating concept sensitivity~\cite{neuhaus2023spurious,pahde2023reveal}.
Concretely,
whereas \cite{neuhaus2023spurious} evaluates the separability of model outputs on samples solely with and without a spurious feature,
\cite{pahde2023reveal} directly inserts artefacts into clean samples and measure the effect on model prediction.

\paragraph{Auto-Interpretability}
The field of automated interpretability
aims to combine the flexibility of human experimentation
with the scalability of automated techniques (usually by using deep models themselves), \eg, for labelling of neurons~\cite{hernandez2021natural,ahn2024www,oikarinen2022clip,kalibhat2023identifying}.
Automated interpretability can be more lightweight, relying on investigated models themselves or foundation models\cite{hernandez2021natural,ahn2024www,oikarinen2022clip,kalibhat2023identifying}, or more involved by solving complex interpretability tasks using \glspl{llm} agents~\cite{shaham2024maia}.

\paragraph{Human-Interpretability Measures}
The work of Network Dissection
\cite{bau2017network} evaluates interpretability indirectly by the degree to which neurons align to a large set of expected concepts.
Later works leverage features spaces of large models, where the concept examples of individual neurons are encoded.
Specifically, 
\cite{dreyer2024pure,foote2024tackling} introduce measures related to polysemanticity,
 \cite{kalibhat2023identifying,li2024evaluating,dreyer2024pure} related to clarity,
and \cite{parekh2024concept} related to redundancy.
Recently,
measures to capture concept complexity have also been introduced~\cite{fel2024understanding}.

\paragraph{Explanation Frameworks}
Instead of focusing on individual aspects, 
explanation frameworks combine multiple interpretability aspects and enable a more holistic understanding of model and data.
For example,
\gls{crp}~\cite{achtibat2023attribution} or CRAFT~\cite{fel2023craft} combine feature visualization and attribution,
but do not include labelling.
CLIP-Dissect~\cite{oikarinen2022clip} on the other hand,
leverages foundation models such as CLIP~\cite{radford2021learning} to label neurons,
but does not investigate how concepts are actually used during inference.
Based on the semantic embedding of model components,
\gls{ours} represents a more comprehensive and holistic framework compared to previous works that enables to systematically search, label, compare, describe and evaluate the inner mechanics of large AI models.
In the following,
several explanatory frameworks that allow to gain insights into deep neural networks are presented and compared to \gls{ours},
as summarized in \cref{app:tab:related_work:overview}.
\begin{table}[h]
    \centering
    \caption{A comparison of selected \gls{xai} frameworks at a glance, considering the explanatory insight they provide for model components. As explanatory capabilities it is considered whether they include concept examples, concept labelling, concept relevances, concept audit capabilities, concept comparison tools, or interpretability evaluation metrics. The table indicates if an explainer exhibits specific explanatory capabilities partially (\kinda) or fully (\yes).}
    \begin{tabular}{ l ||  c|c|c|c|c |  c }
            \hline
            & \multicolumn{6}{c}{Explaining Capabilities for Components} \\
            \hline
              Method & examples   & labels     &  relevances  &  comparison & audit  & interpretability  \\
            \hline
            \hline
                CRP~\cite{achtibat2023attribution} & \yes &  & \yes & \kinda  &        & \\
                CRAFT~\cite{fel2023craft} & \yes &  & \yes &   &        & \\
                PCX~\cite{dreyer2023understanding} & \yes &  & \yes &   &   \kinda     & \\
                Summit~\cite{hohman2019s} & \yes &  & \kinda &  \kinda &  \\
                NetDissect \cite{bau2017network}  & \yes & \yes  &  & \kinda  &  \kinda   & \kinda \\
                CLIP-Dissect \cite{oikarinen2022clip,li2024can}  &\yes & \yes &  & \yes &   \kinda    &  \kinda\\
                FALCON~\cite{kalibhat2023identifying} &\yes & \yes &  & \kinda &   \kinda    &  \kinda\\
                TCAV + IG \cite{schrouff2021best,kim2018interpretability} &  &  & \yes & \kinda &  \kinda     &   \\
                WWW \cite{ahn2024www} & \yes & \yes & \yes &   &                            &   \\
                ConceptEvo \cite{park2023concept}& \yes &  &  &  \yes &           &  \\ 
                SpuFix \cite{neuhaus2023spurious}& \yes &  &  & \kinda  &              \kinda             &  \\ 
                MAIA \cite{shaham2024maia}& \yes & \yes & \kinda &   &              \yes             & \\ 
                \textbf{Ours} & \yes & \yes & \yes & \yes & \yes               & \yes \\
            \hline
        \end{tabular}
    \label{app:tab:related_work:overview}
\end{table}

\textbf{CRP}:
\glsdesc{crp}~\cite{achtibat2023attribution} is a local concept-based explainability approach,
that combines feature visualization techniques with local feature attribution,
thus enabling a much deeper understanding of the decision-making of neural networks than with traditional local feature attribution alone.
Concretely,
the role of a neuron is described by collecting either the most activating samples or the samples where a neuron is most relevant for.
For all neurons and single prediction outcome,
the feature attribution method \gls{lrp}~\cite{bach2015pixel} is extended to compute concept relevance scores and neuron-specific heatmaps.
In their work,
comparisons between neurons are computed based on similarity of neuron relevance patterns.

\textbf{CRAFT}:
Similarly to CRP,
CRAFT~\cite{fel2023craft} combines feature visualization and concept attributions to enable local concept-based explanations.
However,
CRAFT further proposes to perform activation factorization which reduces the high dimensionality of the neural basis (used by CRP).

\textbf{PCX}:
PCX~\cite{dreyer2023understanding} extends CRP by collecting local concept-based explanations and clustering them to extract ``prototypes'' that summarize the model behaviour on the whole (training) dataset.
As such,
PCX enables to reduce the workload of a user for debugging a model as only few prototypes need to be studied instead of thousands of explanations.
However,
PCX does not provide any neuron labels.

\textbf{Summit}:
The Summit framework combines local relevance scores and feature visualization techniques into compact visualizations, aiming to guide and facilitate the manual inspection of convolutional neurons and their roles within a network. 
Class relevances are derived by aggregating neuron activations over data samples associated with specific classes, while conditional neuron-to-neuron relevances are computed by aggregating the product of peak activations and connecting weights.
For visualization, class relevance scores for each neuron are combined into a vector and visualized using a UMAP projection to provide an overview of class specificity within the layer. 
The conditional relevance scores, on the other hand, are combined into an ``attribution graph'' that illustrates the interactions and roles of individual neurons.
Both the UMAP projections and attribution graphs are accompanied by sampled and generated concept examples for each neuron.
While not explored in the original paper, the stacked class-relevance vectors can also be used to compare components across layers or architectures, offering further insights into network behaviour.

\textbf{NetDissect}:
Network Dissection~\cite{bau2017network} is one of the first explanatory frameworks that aims to quantitatively analyse the latent representations of deep neural networks.
In a first step,
channels of convolutional layers are labelled, by matching their upsampled spatial activation maps with densely annotated labelled data.
The labelled representations allow now to compare models by what the models have learned and how well they match certain labels.
In order to compare,
however,
labels need to be available.
In principle,
Network Dissection also allows to audit models by checking if they align to expected labels (but without indication how they are used).
Also,
latent interpretability can be evaluated,
with the assumption that low alignments indicate low interpretability.

\textbf{Net2Vec}:
Net2Vec~\cite{fong2018net2vec} is a framework in which (user-defined) concepts are mapped to vector embeddings based on corresponding component activations.
Concretely,
for each concept,
neuron activations are collected on a reference (``probe'') dataset,
and subsequently weights for each neuron are estimated that correspond to the usefulness to detect the concept.
As such, these vector embeddings allow to show that in most cases, multiple filters are required to code for a concept, and that often neurons are not concept-specific and
encode multiple concepts.
In their work,
they use NetDissect to visualize the function of single neurons.
The work observes that, compared to activation patterns, their Net2Vec embeddings are able to better characterize the meaning of a representation and its relationship to other concepts.
Further,
the Net2Vec embeddings allow to compare whole feature spaces of different models.
Notably,
Net2Vec aims to compare how concepts are represented in \emph{whole} feature spaces, and is not meant for component-level analysis.

\textbf{CLIP-Dissect}:
CLIP-Dissect~\cite{oikarinen2022clip} extends Network Dissection by integrating a multimodal foundation model such as CLIP into the labelling pipeline.
In principle,
CLIP is used here to soft-label the dataset with expected concepts.
The activation pattern of each neuron is compared to the soft labels of CLIP in order to label.
The activation patterns and labels can further be used to compare models in~\cite{li2024can}.

\textbf{TCAV + IG}:
TCAV~\cite{kim2018interpretability} is a popular framework for evaluating the existence of expected concepts and the model's concept sensitivity.
Concretely,
in a first step,
a linear direction in the latent space of some layer is estimated
using latent activations on data samples with and without concept.
In order to test the model,
the latent gradient is used, 
which~\cite{schrouff2021best} extended to the more stable Integrated Gradients (IG)~\cite{sundararajan2017axiomatic} method.
As such,
we can compare models and their sensitivity \wrt expected concepts.
However,
there is no indication on how much the model actually relies on unexpected concepts.

\textbf{FALCON}:
Similarly to \gls{ours}, FALCON~\cite{kalibhat2023identifying} is based on using CLIP models to annotate representations.
FALCON uses therefore not only most activating samples, but also (visually) similar but lowly activating samples to further improve labelling.
The work of~\cite{kalibhat2023identifying} further estimates interpretability of representations based on semantic similarities of concept examples, as proposed for the clarity measure in \cref{sec:methods:interpretability}.

\textbf{WWW}:
The WWW framework~\cite{ahn2024www} combines feature visualization and attribution techniques with neuron labelling to explain networks' decision-making on a local and global level. Their proposed labelling pipeline is most similar to the one we describe in \cref{sec:app:methods:workflow}. To label a neuron, they collect data samples that maximize the neuron's activation and embed them into the latent space of a CLIP model alongside a set of predefined text labels. Based on the pairwise cosine similarities between the image and text embeddings, a set of labels for the examined neuron is selected using an adaptive threshold.
Our labelling procedure differs in two key ways: we crop the collected samples to the image portions that include the most relevant pixels with respect to the inspected neuron, reducing the influence of background features not relevant to the neuron. Additionally, we average the image embeddings before measuring the cosine similarity, which also helps reduce the influence of noise in the data.

\textbf{ConceptEvo}:
ConceptEvo~\cite{park2023concept} focuses on interpreting and comparing deep neural networks during training. 
Similar to our approach, it employs a unified semantic space to embed model components for comparison and interpretation; however, ConceptEvo constructs this space from scratch. 
This is done by learning neuron embeddings based on co-activation relationships in a base model and aligning image embeddings by minimizing their distance to neurons they strongly activate. 
Network components from other models are then represented by averaging the embeddings of their strongly activating images.
Whereas  ConceptEvo proposes a novel way to measure importance of a concept evolution, no relevance measures for individual concepts are presented.

\textbf{SpuFix}:
In \cite{neuhaus2023spurious}, \citeauthor{neuhaus2023spurious} extensively analyse the representations of a robustly trained ResNet50 ImageNet classifier by applying activation factorization, similar to CRAFT. They manually inspect and label the resulting activation directions as encoding either spurious or valid features for the corresponding class.
Building on this analysis, they propose the SpuFix method to identify spurious directions in other ImageNet classifiers without requiring manual labelling. 
SpuFix aligns the spurious directions identified in the ResNet50 model with directions in the target classifier by maximizing co-activation, and prunes the matched spurious components to mitigate reliance on spurious features.
Additionally, the manually labelled spurious directions are used to construct \emph{Spurious ImageNet}, a dataset containing only spurious features for 100 ImageNet classes. 
This dataset enables the evaluation of a classifier's reliance on spurious features by assessing overall and per-class accuracy metrics.

\textbf{MAIA}:
The Multimodal Automated Interpretability Agent (MAIA) utilizes a pre-trained vision-language model and a set of tools, like collecting highly activating samples for a given neuron, cropping images, \etc to derive an interpretability agent. 
Presented with a question like ``What does neuron \texttt{\#42} in layer 5 encode?'' the vision-language model autonomously queries a provided API of interpretability tools, and runs multiple hypothesis and validation iterations before providing an answer.

\setcounter{figure}{0}
\setcounter{table}{0}\setcounter{equation}{0}
\section{Experimental Settings}
\label{app:experimental-settings}
The following section outlines the experimental settings used throughout this work.

\subsection{Architectures and Models}
    We evaluate multiple pre-trained models from the torchvision~\cite{torchvision2016} and hugging face model zoo as detailed in the following.
    \paragraph{ResNet}
        The ResNet is a convolutional neural network architecture consisting of four layer blocks and one fully connected layer.
        For all experiments,
        we collect activation and relevance scores after each layer block.
        Concretely,
        we use the ResNet~\cite{he2016deep} architectures: 
        ResNet18, ResNet32, ResNet50, ResNet50v2, ResNet101, ResNet101v2 provided by torchvision~\cite{torchvision2016}.
        We further evaluate 
        ResNet18 with identifier ``resnet18.a1\_in1k'',
        ResNet34 with identifier ``resnet34.a1\_in1k'',
        ResNet50s with identifiers ``resnet50.a1\_in1k'', ``resnet50d.a1\_in1k'', ``resnet50d.a2\_in1k'', and
        ResNet101 with identifier ``resnet101.a1\_in1k''
        from the timm model zoo~\cite{rw2019timm}.
    
    \paragraph{VGG}
    The VGG~\cite{simonyan2014very} is a convolutional neural network architecture consisting of a set of convolutional layers and three fully connected layers.
    In our experiments,
    the VGG-13, VGG-16 and VGG-19 with and without Batch Normalization~\cite{ioffe2015bn} (BN) layers are used from the torchvision model zoo.
    
    \paragraph{Vision Transformer}
    The vision transformer utilizes attention and fully connected layers which are applied to the input image after it was split  into tiles and projected into a sequence of tokens. 
    In our experiments, we use the model with identifiers 
    \begin{itemize}
    \item ``vit\_small\_patch16\_224.augreg\_in21k\_ft\_in1k'', 
    \item ``vit\_mediumd\_patch16\_reg4\_gap\_256.sbb\_in12k\_ft\_in1k''
        \item 
    ``vit\_large\_patch16\_224.augreg\_in21k\_ft\_in1k'', 
    
    \end{itemize}
    from timm model zoo~\cite{rw2019timm}.
    Notably,
    the \gls{vit}'s last linear layer does not operate on activations that are preprocessed by a ReLU non-linearity.
    As such,
    activations can be negative or positive.
    Thus,
    we apply both activation maximization and minimization to generate concept examples.
    Concretely,
    we duplicate the feature dimension (\ie, double the amount of neurons),
    and collect most activating samples for the first half,
    and minimal activating samples for the second half of neurons.

    \paragraph{Foundation Models}
    We use Mobile-CLIP~\cite{vasu2024mobileclip} throughout all ImageNet experiments in \cref{sec:results:explore,sec:results:audit},
    DINOv2~\cite{oquab2023dinov2} for the interpretability experiments in \cref{sec:results:evaluate},
    and WhyLesionCLIP~\cite{yang2024textbook} for the medical experiments per default.
    Overall, we used five foundation models in this work: 
    CLIP-OpenAI~\cite{radford2021learning} (CLIP-ViT-base-patch32), CLIP-LAION~\cite{schuhmann2022laion} (CLIP-ViT-B-32-laion2B-s34B-b79K), Mobile-CLIP~\cite{vasu2024mobileclip} (MobileCLIP-S2), DINOv2~\cite{oquab2023dinov2} (DINOv2-base), and WhyLesionCLIP~\cite{yang2024textbook}.

\subsection{Datasets}
We use two datasets throughout our experiments, which are presented in the following.

\subsubsection{ImageNet}
ImageNet~\cite{deng2009imagenet} (specifically ImageNet-1k) is a visual object classification dataset with 1000 object classes and contains 1,281,167 training images and 50,000 validation images.
ImageNet is used throughout \cref{sec:results:explore,sec:results:audit,sec:results:evaluate}.

All data is, per default,
loaded as follows:
(1) resizing images to a resolution of $256\times256$,
(2) centre crop to $224\times224$ pixels,
(3) normalization to range [0, 1] (RGB values are divided by 255), and
(4) subtracting [0.485, 0.456, 0.406] followed by a normalization of standard deviation values [0.229, 0.224, 0.225] over the red, green and blue colour channels respectively.

\subsubsection{ISIC Challenge}
The medical dataset used in \cref{sec:results:medical} is taken from the ISIC challenge of 2019~\cite{tschandl2018ham10000,codella2018skin,hernandez2024bcn20000},
including 25,331 dermoscopic images among nine different diagnostic categories:
``Melanoma'',
``Melanocytic nevus'',
``Basal cell carcinoma'',
``Actinic keratosis'',
``Benign keratosis'',
``Dermatofibroma'',
``Vascular lesion'',
and
``Squamous cell carcinoma''.
For simplicity,
we group all categories other than melanoma into one group ``other'' to receive a binary classification setting.

All data is, per default,
loaded as follows:
(1) resizing images to a resolution of $224\times224$,
(3) normalization to range [0, 1] (RGB values are divided by 255), and
(4) subtracting [0.5, 0.5, 0.5] followed by a normalization of standard deviation values [0.5, 0.5, 0.5] over the red, green and blue colour channels respectively.

\setcounter{figure}{0}
\setcounter{table}{0}\setcounter{equation}{0}
\section{Search in Semantic Space}
\label{sec:app:exp:search}

In this section, 
we provide additional details and examples of the \emph{search} tool within \gls{ours}, as presented in \cref{sec:results:explore:search}. A quantitative analysis of \gls{ours}' labelling capabilities—building on \emph{search}—and the effects of selected hyperparameters are discussed in \cref{sec:app:explore:describe:evaluation}.

\begin{figure}[t]
    \centering
    
    \includegraphics[width=\textwidth]{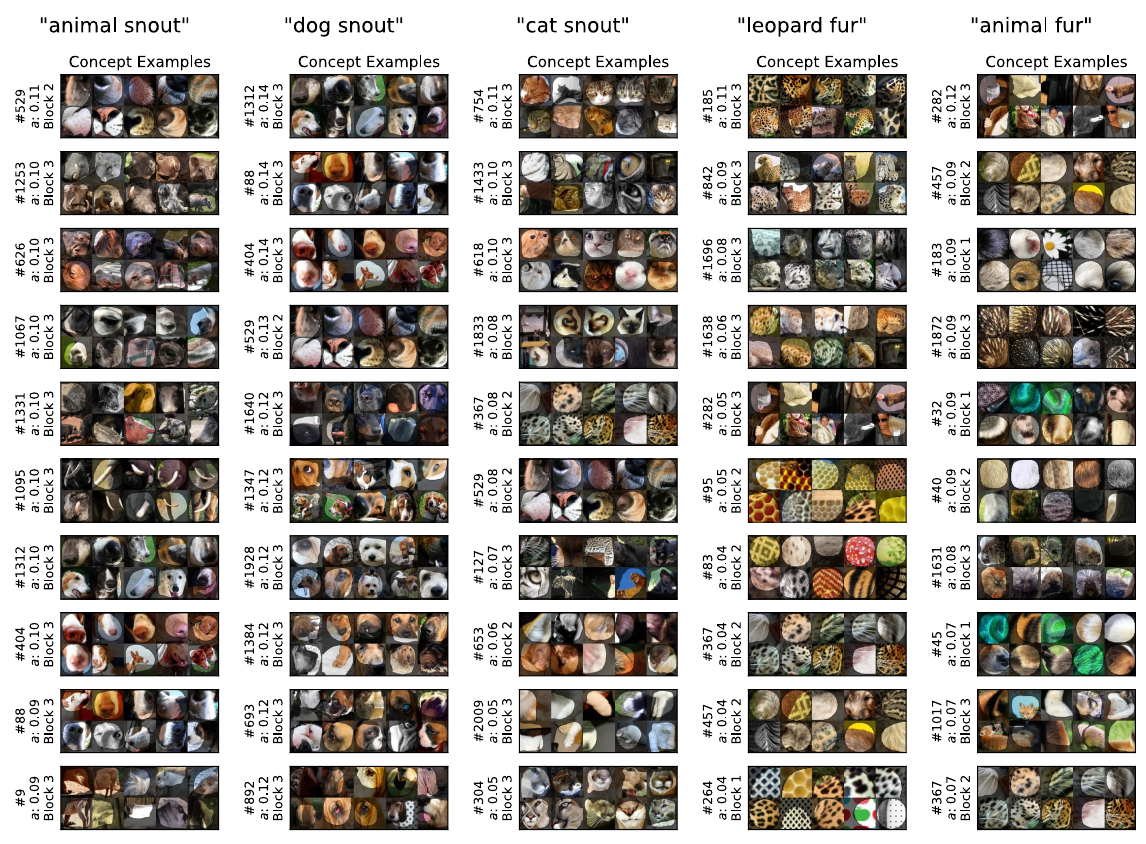}
    \caption{    
    \emph{Search} results among all neurons in the final layers of each of the four ResNet blocks in the ResNet50v2 for animal-related queries: ``animal snout'', ``dog snout'', ``cat snout'', ``leopard fur'' and ``animal fur'' (from left to right). 
    We show the top-10 most aligned neurons, sorted in descending order from top to bottom. For each neuron, we provide the specific alignment score $a$ (as defined in \cref{eq:app:explore:search}), along with the neuron index and layer name.
    The scores are derived using the CLIP-Mobile-S2 foundation model with the three prompt templates: 
    ``\texttt{<concept>}'',
    ``a \texttt{<concept>}'' 
    and ``an image of a close up of \texttt{<concept>}''.
    The ``null'' embedding is obtained from the empty templates.
    The results demonstrate \gls{ours}' ability to identify neurons specific to a given query and differentiate between similar concepts, such as the general term ``animal snout'' and more specific terms like ``dog snout'' or ``cat snout.''
    }
    \label{fig:app:explore:search:examples:1}
\end{figure}

The \emph{search} tool of \gls{ours} enables to identify neurons that have learned concepts similar to a given search prompt.
As described in \cref{sec:methods:search}, the \emph{search} process operates on the semantic embeddings $\mathcal{V}_\mathcal{M} = \{\boldsymbol{\vartheta}_1, \dots, \boldsymbol{\vartheta}_k\}$ of the components of a model $\mathcal{M}$. The process involves the following steps:
\begin{enumerate}
    \item \textbf{Probe Selection} 
    The searched concept can be specified in any modality supported by the foundation model within \gls{ours}, or as combinations thereof. 
    For instance, the search prompt may take the form of
    one or multiple samples from the model's data domain (\eg, images of green objects), 
    or one or multiple text descriptions (\eg, ``a photo of a green car'', ``a photo of a green wall'', \etc).
    \item \textbf{Probe Embedding}  
    The selected search prompt is embedded into the semantic space. If multiple prompts are provided, their embeddings are aggregated (\eg, via average pooling) to produce a single probe embedding, $\boldsymbol{\vartheta}_\text{probe}$.
    \item \textbf{Similarity Search}  
    The final step is a similarity search within the set of semantic embeddings $\mathcal{V}_\mathcal{M}$ \wrt $\boldsymbol{\vartheta}_\text{probe}$: 
    \begin{equation}
    \label{eq:app:explore:search}
        \boldsymbol{\vartheta}^\ast = \operatornamewithlimits{argmax}_{\boldsymbol{\vartheta}\in\mathcal{V}_\mathcal{M}} 
        \{
        a(\boldsymbol{\vartheta}, \boldsymbol{\vartheta}_\text{probe})
        \}
        \quad \text{with} \quad
        a(\boldsymbol{\vartheta}, \boldsymbol{\vartheta}_\text{probe}) \coloneqq 
        s(\boldsymbol{\vartheta}, \boldsymbol{\vartheta}_\text{probe}) {-} s(\boldsymbol{\vartheta}, \boldsymbol{\vartheta}_\texttt{<>}).
    \end{equation}
    Here, $\boldsymbol{\vartheta}_\texttt{<>}$ is a "null" embedding used to mitigate the influence of background noise, such as background objects in image prompts, or irrelevant words in text prompts (\eg, "photo" in the prompt template ``\texttt{a photo of <concept>}'').
    The similarity score $s$ is defined in \cref{eq:methods:similarity_measure}.
\end{enumerate}

\begin{figure}[t]
    \centering
    
    \includegraphics[width=\textwidth]{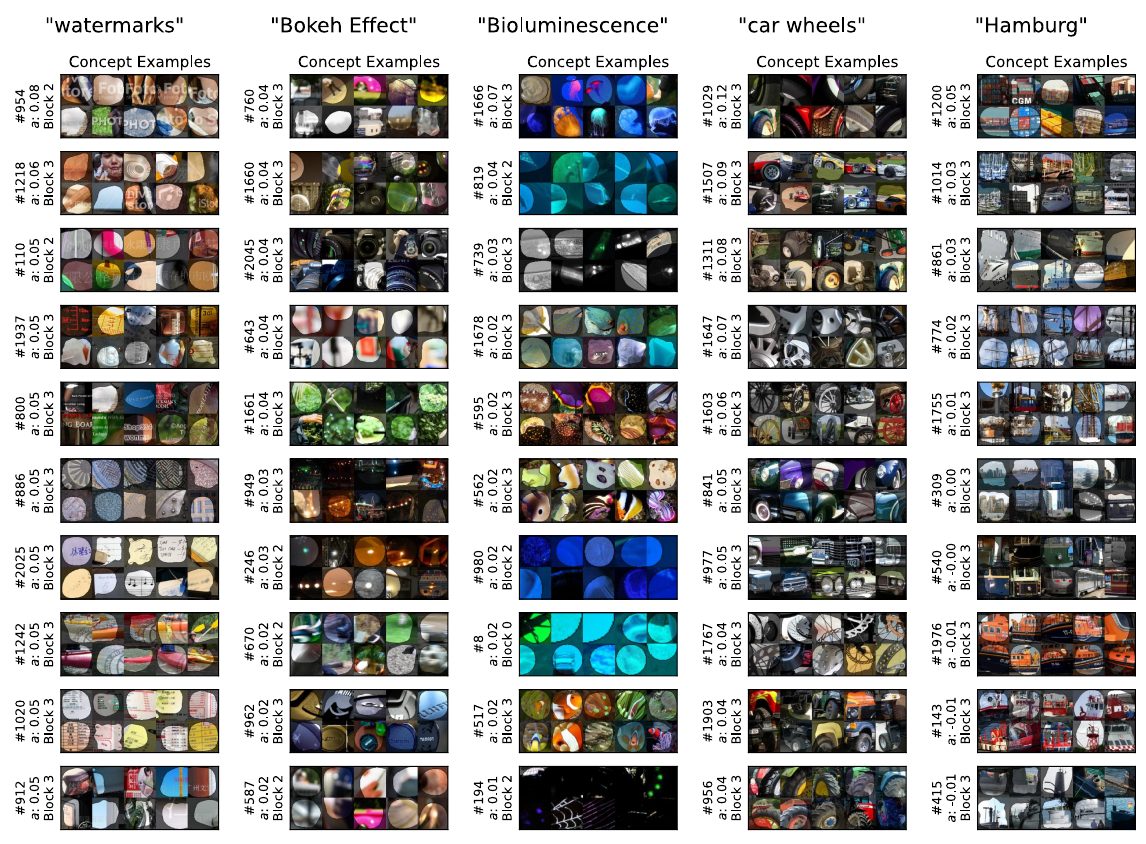}
    \caption{    
    \emph{Search} results among all neurons in the final layers of each of the four ResNet blocks in the ResNet50v2 for the queries: 
    ``watermarks'', 
    ``Bokeh Effect'', 
    ``Bioluminescence'', 
    ``car wheels'' and 
    ``Hamburg'' 
    (from left to right). 
    We show the top-10 most aligned neurons, sorted in descending order from top to bottom. For each neuron, we provide the specific alignment score $a$ (as defined in \cref{eq:app:explore:search}), along with the neuron index and layer name.
    The scores are derived using the CLIP-Mobile-S2 foundation model with the three prompt templates: 
    ``\texttt{<concept>}'',
    ``a \texttt{<concept>}'' 
    and ``an image of a close up of \texttt{<concept>}''.
    The ``null'' embedding is obtained from the empty templates.
    The results showcase the diverse range of learned representations that can be discovered using \gls{ours}. These include spurious concepts, such as \emph{watermarks}, specialized neurons that activate for unique phenomena like \emph{bioluminescence}, and even abstract representations, such as an entire city, like \emph{Hamburg}, encompassing maritime and Harbor-like features.
    }
    \label{fig:app:explore:search:examples:2}
\end{figure}

In \cref{fig:app:explore:search:examples:2,fig:app:explore:search:examples:1}, we present \emph{search} results for different text-prompts, carried out across the four ResNet blocks in the ResNet50v2 \gls{imagenet} classifier using the CLIP-Mobile-S2 foundation model. These examples highlight \gls{ours}' ability to identify and differentiate between similar yet distinct representations based on the selected search probes (\cref{fig:app:explore:search:examples:1}) and demonstrate the range of concept specificity that can be uncovered (\cref{fig:app:explore:search:examples:2}).

\begin{figure}[t]
    \centering
    
    \includegraphics[width=\textwidth]{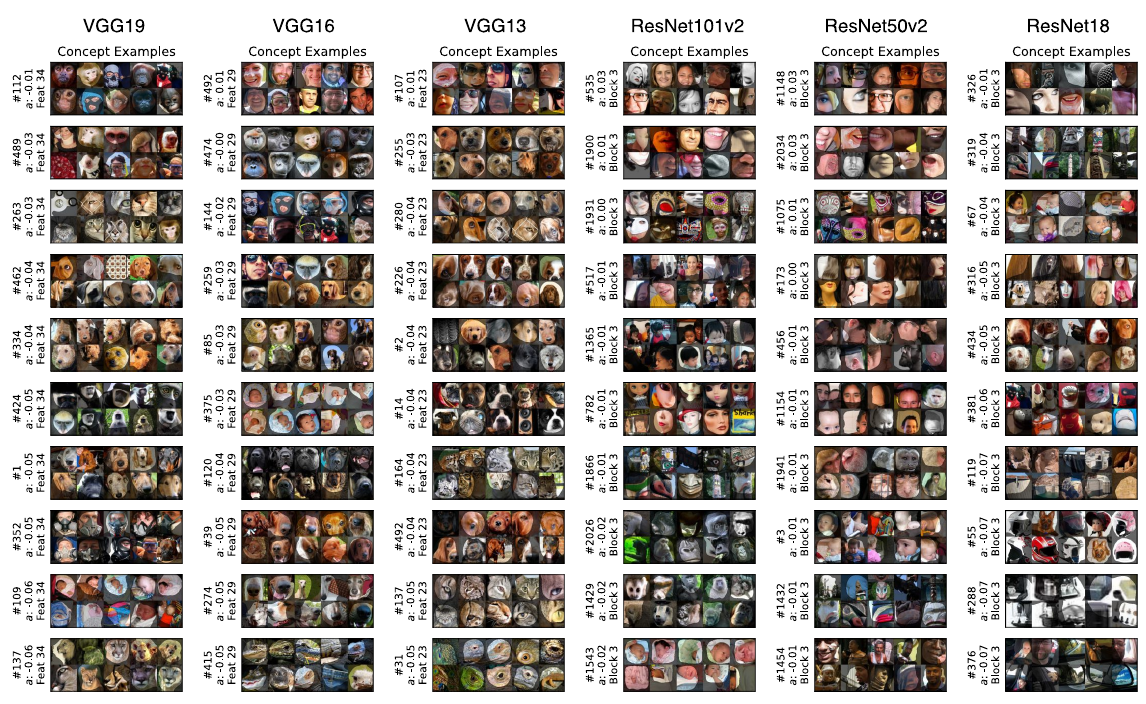}
    \caption{    
    Comparison of \emph{search} results among neurons in the final feature layer of VGG19, VGG16, VGG13, ResNet101v2, ResNet50v2 and ResNet18 (from left to right) for the search query ``human face''. 
    We present the top-10 most aligned neurons, sorted in descending order from top to bottom. For each neuron, we provide the specific alignment score $a$ (as defined in \cref{eq:app:explore:search}), along with the neuron index and layer name. The scores are derived using the CLIP-Mobile-S2 model with the following three prompt templates:
    ``\texttt{<concept>}'',
    ``a \texttt{<concept>}'' 
    and ``an image of a close up of \texttt{<concept>}''.
    }
    \label{fig:app:explore:search:comp_models}
\end{figure}

The ability to search for concepts naturally facilitates comparisons between models, providing insights into a model's general and relative expertise with respect to the queried concepts. In \cref{fig:app:explore:search:comp_models}, we display the neurons most aligned with the text probe ``human face'' across six distinct \gls{imagenet} classifiers. \gls{ours}' comparative capabilities extend beyond visual inspection and are examined in further detail in \cref{app:exp:compare}.

\setcounter{figure}{0}
\setcounter{table}{0}\setcounter{equation}{0}
\section{Describe in Semantic Space}
\label{sec:app:exp:describe}
This section provides more details on \cref{sec:results:explore:describe},
where it is demonstrated how \gls{ours} can be utilized to describe the internal representations and encoded knowledge.
Concretely,
\cref{app:explore:describe:labels} provides an overview of used concept labels,
and \cref{app:exp:describe:umap,app:explore:describe:attribution_graph,app:explore:describe:dissection} include details and additional experiments for UMAP embeddings, attribution graphs, and network dissection, respectively.
Lastly,
\cref{sec:app:explore:describe:evaluation} provides faithfulness evaluations for labels.

\subsection{Concept Set}
\label{app:explore:describe:labels}
The following concept labels are used for annotation if not stated otherwise:
\small{
\paragraph{Colors}
\texttt{red color}, \texttt{blue color}, \texttt{green color}, \texttt{yellow color}, \texttt{black color}, \texttt{white color}, \texttt{gray color}, \texttt{brown color}, \texttt{orange color}, \texttt{pink color}, \texttt{purple color}, \texttt{cyan color}, \texttt{magenta color}, \texttt{teal color}, \texttt{maroon color}, \texttt{navy color}, \texttt{olive color}, \texttt{lime color}, \texttt{aqua color}, \texttt{coral color}, \texttt{turquoise color}, \texttt{gold color}, \texttt{silver color}, \texttt{beige color}, \texttt{lavender color}, \texttt{peach color}, \texttt{mint color}, \texttt{rose color}, \texttt{violet color}, \texttt{charcoal color}, \texttt{salmon color}, \texttt{chocolate color}, \texttt{emerald color}, \texttt{crimson color}, \texttt{indigo color}, \texttt{tan color}, \texttt{ruby color}, \texttt{amber color}, \texttt{sapphire color}, \texttt{fuchsia color}, \texttt{bronze color}, \texttt{copper color}, \texttt{ivory color}, \texttt{plum color}, \texttt{mustard color}, \texttt{khaki color}, \texttt{periwinkle color}, \texttt{sand color}, \texttt{slate color}, \texttt{burgundy color}, \paragraph{Textures}
\texttt{smooth texture}, \texttt{rough texture}, \texttt{bumpy texture}, \texttt{grainy texture}, \texttt{glossy texture}, \texttt{matte texture}, \texttt{silky texture}, \texttt{velvety texture}, \texttt{fuzzy texture}, \texttt{prickly texture}, \texttt{crinkled texture}, \texttt{wrinkled texture}, \texttt{ribbed texture}, \texttt{knitted texture}, \texttt{woven texture}, \texttt{quilted texture}, \texttt{embossed texture}, \texttt{pebbled texture}, \texttt{brushed texture}, \texttt{etched texture}, \texttt{polished texture}, \texttt{satin texture}, \texttt{metallic texture}, \texttt{rubberized texture}, \texttt{plastic texture}, \texttt{woodgrain texture}, \texttt{stone texture}, \texttt{leather texture}, \texttt{suede texture}, \texttt{linen texture}, \texttt{canvas texture}, \texttt{corduroy texture}, \texttt{denim texture}, \texttt{fur texture}, \texttt{feathered texture}, \texttt{wool texture}, \texttt{lace texture}, \texttt{mesh texture}, \texttt{net texture}, \texttt{sheer texture}, \texttt{transparent texture}, \texttt{opaque texture}, \texttt{speckled texture}, \texttt{striped texture}, \texttt{plaid texture}, \texttt{checked texture}, \texttt{houndstooth texture}, \texttt{chevron texture}, \texttt{paisley texture}, \texttt{floral texture}, \texttt{geometric texture}, \texttt{blotchy texture}, \texttt{dotted texture}, \texttt{banded texture}, \texttt{smeared texture}, \texttt{porous texture}, \texttt{pitted texture}, \texttt{fibrous texture}, \texttt{veined texture}, \texttt{perforated texture}, \texttt{crosshatched texture}, \texttt{sprinkled texture}, \texttt{polka-dotted texture}, \texttt{marbled texture}, \texttt{stained texture}, \texttt{grid texture}, \texttt{gauzy texture}, \texttt{interlaced texture}, \texttt{frilly texture}, \texttt{zigzagged texture}, \texttt{spiralled texture}, \texttt{swirly texture}, \texttt{cracked texture}, \texttt{studded texture}, \texttt{matted texture}, \texttt{flecked texture}, \texttt{potholed texture}, \texttt{scaly texture}, \texttt{stratified texture}, \texttt{braided texture}, \texttt{lined texture}, \texttt{waffled texture}, \texttt{freckled texture}, \texttt{honeycombed texture}, \texttt{lacelike texture}, \texttt{chequered texture}, \texttt{crystalline texture}, \texttt{bubbly texture}, \texttt{grooved texture}, \texttt{pleated texture}, \texttt{cobwebbed texture}, \texttt{waves texture}, \texttt{textile texture}, \texttt{text texture}, \texttt{font texture}, \texttt{lettering texture}, \texttt{noise texture}, \paragraph{Animals}
\texttt{animal}, \texttt{mammal}, \texttt{bird}, \texttt{reptile}, \texttt{amphibian}, \texttt{insect}, \texttt{dog}, \texttt{cat}, \texttt{cow}, \texttt{horse}, \texttt{sheep}, \texttt{pig}, \texttt{chicken}, \texttt{duck}, \texttt{goose}, \texttt{turkey}, \texttt{rabbit}, \texttt{deer}, \texttt{mouse}, \texttt{rat}, \texttt{elephant}, \texttt{tiger}, \texttt{lion}, \texttt{bear}, \texttt{wolf}, \texttt{fox}, \texttt{monkey}, \texttt{gorilla}, \texttt{chimpanzee}, \texttt{panda}, \texttt{kangaroo}, \texttt{koala}, \texttt{giraffe}, \texttt{zebra}, \texttt{hippopotamus}, \texttt{rhinoceros}, \texttt{leopard}, \texttt{cheetah}, \texttt{hyena}, \texttt{crocodile}, \texttt{alligator}, \texttt{snake}, \texttt{lizard}, \texttt{frog}, \texttt{toad}, \texttt{turtle}, \texttt{tortoise}, \texttt{fish}, \texttt{shark}, \texttt{whale}, \texttt{dolphin}, \texttt{octopus}, \texttt{crab}, \texttt{lobster}, \texttt{penguin}, \texttt{ostrich}, \texttt{human}, \texttt{man}, \texttt{woman}, \texttt{baby}, \paragraph{Objects}
\texttt{plate}, \texttt{fan}, \texttt{sponge}, \texttt{joystick}, \texttt{sunglasses}, \texttt{sandals}, \texttt{tissue}, \texttt{light}, \texttt{drum}, \texttt{iron}, \texttt{wrench}, \texttt{ironing board}, \texttt{modem}, \texttt{game console}, \texttt{projector}, \texttt{ship}, \texttt{cable}, \texttt{knife}, \texttt{perfume}, \texttt{pliers}, \texttt{tennis ball}, \texttt{chair}, \texttt{scarf}, \texttt{fork}, \texttt{keyboard}, \texttt{cup}, \texttt{ring}, \texttt{router}, \texttt{sweater}, \texttt{pillow}, \texttt{notebook}, \texttt{shorts}, \texttt{shirt}, \texttt{blouse}, \texttt{boots}, \texttt{dishwasher}, \texttt{garbage can}, \texttt{rug}, \texttt{heater}, \texttt{spoon}, \texttt{charger}, \texttt{backpack}, \texttt{cabinet}, \texttt{horn}, \texttt{car}, \texttt{pen}, \texttt{oven}, \texttt{soap}, \texttt{mouse}, \texttt{comb}, \texttt{television}, \texttt{gloves}, \texttt{hammer}, \texttt{paper}, \texttt{automobile}, \texttt{carpet}, \texttt{tablet}, \texttt{conditioner}, \texttt{fireplace}, \texttt{shampoo}, \texttt{lens}, \texttt{dress}, \texttt{table}, \texttt{sneakers}, \texttt{watch}, \texttt{truck}, \texttt{nail polish}, \texttt{belt}, \texttt{stamp}, \texttt{remote}, \texttt{clothing}, \texttt{shower}, \texttt{air conditioner}, \texttt{coat}, \texttt{sock}, \texttt{makeup}, \texttt{apparel}, \texttt{lipstick}, \texttt{cosmetics}, \texttt{chandelier}, \texttt{printer}, \texttt{bucket}, \texttt{jeans}, \texttt{razor}, \texttt{kettle}, \texttt{usb}, \texttt{stove}, \texttt{candle}, \texttt{headphones}, \texttt{sheet}, \texttt{bag}, \texttt{map}, \texttt{toaster}, \texttt{pants}, \texttt{mirror}, \texttt{scanner}, \texttt{mattress}, \texttt{sink}, \texttt{drawer}, \texttt{first aid kit}, \texttt{recycling bin}, \texttt{garbage}, \texttt{glue}, \texttt{plug}, \texttt{necklace}, \texttt{instrumentmedicine}, \texttt{pill}, \texttt{lamp}, \texttt{tape}, \texttt{toothbrush}, \texttt{bookshelf}, \texttt{hat}, \texttt{phone}, \texttt{broom}, \texttt{shoes}, \texttt{toothpaste}, \texttt{scissors}, \texttt{boat}, \texttt{curtain}, \texttt{poster}, \texttt{jacket}, \texttt{suitcase}, \texttt{hairbrush}, \texttt{footwear}, \texttt{package}, \texttt{mousepad}, \texttt{van}, \texttt{juicer}, \texttt{bus}, \texttt{towel}, \texttt{scale}, \texttt{computer}, \texttt{speaker}, \texttt{laptop}, \texttt{vacuum cleaner}, \texttt{desk}, \texttt{mixer}, \texttt{glasses}, \texttt{bottle}, \texttt{mug}, \texttt{calendar}, \texttt{globe}, \texttt{screen}, \texttt{pan}, \texttt{fridge}, \texttt{tie}, \texttt{motorcycle}, \texttt{letter}, \texttt{coffee maker}, \texttt{flag}, \texttt{door}, \texttt{lotion}, \texttt{pencil}, \texttt{earring}, \texttt{painting}, \texttt{blender}, \texttt{ruler}, \texttt{glass}, \texttt{water}, \texttt{tripod}, \texttt{sculpture}, \texttt{guitar}, \texttt{bathtub}, \texttt{bed}, \texttt{box}, \texttt{underwear}, \texttt{microwave}, \texttt{brush}, \texttt{envelope}, \texttt{piano}, \texttt{trumpet}, \texttt{monitor}, \texttt{drawing}, \texttt{stick}, \texttt{binoculars}, \texttt{bowl}, \texttt{pot}, \texttt{bicycle}, \texttt{can}, \texttt{mop}, \texttt{wallet}, \texttt{bandage}, \texttt{camera}, \texttt{microphone}, \texttt{airplane}, \texttt{memory card}, \texttt{golf ball}, \texttt{plane}, \texttt{violin}, \texttt{adapter}, \texttt{vacuum bag}, \texttt{window}, \texttt{skirt}, \texttt{teapot}, \texttt{bracelet}, \texttt{clock}, \texttt{toilet}, \texttt{statue}, \texttt{battery}, \texttt{remote control}, \texttt{tent}, \texttt{blanket}, \texttt{cream}, \texttt{train}, \texttt{laundry basket}, \texttt{purse}, \texttt{extension cord}, \texttt{book}, \texttt{tool}, \texttt{cap}, \texttt{screwdriver}, \paragraph{Animal\_parts}
\texttt{head}, \texttt{eye}, \texttt{ear}, \texttt{nose}, \texttt{mouth}, \texttt{tongue}, \texttt{teeth}, \texttt{fang}, \texttt{beak}, \texttt{hand}, \texttt{arm}, \texttt{finger}, \texttt{body}, \texttt{snout}, \texttt{horn}, \texttt{antler}, \texttt{neck}, \texttt{throat}, \texttt{shoulder}, \texttt{back}, \texttt{spine}, \texttt{tail}, \texttt{leg}, \texttt{paw}, \texttt{hoof}, \texttt{claw}, \texttt{foot}, \texttt{toe}, \texttt{wing}, \texttt{feather}, \texttt{fin}, \texttt{gill}, \texttt{scale}, \texttt{fur}, \texttt{skin}, \texttt{coat}, \texttt{whisker}, \texttt{mane}, \texttt{tusk}, \texttt{trunk}, \texttt{tentacle}, \texttt{gill}, \texttt{flipper}, \texttt{shell}, \texttt{plumage}, \texttt{beard}, \texttt{comb}, \texttt{wattle}, \texttt{crest}, \texttt{spur}, \texttt{hind leg}, \texttt{foreleg}, \texttt{udder}, \texttt{legs}, \texttt{feet}, \paragraph{Object\_parts}
\texttt{handle}, \texttt{knob}, \texttt{button}, \texttt{lever}, \texttt{switch}, \texttt{dial}, \texttt{screen}, \texttt{display}, \texttt{key}, \texttt{lock}, \texttt{lid}, \texttt{cap}, \texttt{cover}, \texttt{base}, \texttt{leg}, \texttt{foot}, \texttt{wheel}, \texttt{axle}, \texttt{blade}, \texttt{edge}, \texttt{tip}, \texttt{point}, \texttt{corner}, \texttt{side}, \texttt{face}, \texttt{panel}, \texttt{frame}, \texttt{bracket}, \texttt{hinge}, \texttt{joint}, \texttt{bolt}, \texttt{screw}, \texttt{nut}, \texttt{washer}, \texttt{spring}, \texttt{hook}, \texttt{clasp}, \texttt{clip}, \texttt{latch}, \texttt{rivet}, \texttt{pin}, \texttt{peg}, \texttt{plug}, \texttt{socket}, \texttt{cord}, \texttt{cable}, \texttt{chain}, \texttt{strap}, \texttt{belt}, \texttt{wheel}, \texttt{pulley}, \texttt{rope}, \texttt{string}, \texttt{chinese text}, \paragraph{Object\_scenes}
\texttt{kitchen}, \texttt{bathroom}, \texttt{living room}, \texttt{bedroom}, \texttt{dining room}, \texttt{office}, \texttt{garden}, \texttt{garage}, \texttt{classroom}, \texttt{library}, \texttt{restaurant}, \texttt{café}, \texttt{supermarket}, \texttt{mall}, \texttt{park}, \texttt{playground}, \texttt{beach}, \texttt{mountain}, \texttt{forest}, \texttt{desert}, \texttt{farm}, \texttt{barn}, \texttt{factory}, \texttt{warehouse}, \texttt{airport}, \texttt{train station}, \texttt{bus station}, \texttt{hospital}, \texttt{clinic}, \texttt{laboratory}, \texttt{gym}, \texttt{stadium}, \texttt{theater}, \texttt{cinema}, \texttt{museum}, \texttt{gallery}, \texttt{workshop}, \texttt{studio}, \texttt{hotel}, \texttt{church}, \texttt{temple}, \texttt{mosque}, \texttt{market}, \texttt{street}, \texttt{alley}, \texttt{highway}, \texttt{bridge}, \texttt{tunnel}, \texttt{harbor}, \texttt{dock}, \texttt{music studio}, \texttt{dirt track}, \texttt{sky}, \texttt{sea}, \texttt{snow}, \paragraph{Vegetations}
\texttt{vegetation}, \texttt{grass}, \texttt{tree}, \texttt{shrub}, \texttt{bush}, \texttt{fern}, \texttt{moss}, \texttt{vine}, \texttt{flower}, \texttt{weed}, \texttt{algae}, \texttt{cactus}, \texttt{palm}, \texttt{pine}, \texttt{oak}, \texttt{maple}, \texttt{birch}, \texttt{willow}, \texttt{cedar}, \texttt{spruce}, \texttt{fir}, \texttt{bamboo}, \texttt{ivy}, \texttt{lichen}, \texttt{herb}, \texttt{succulent}, \texttt{reed}, \texttt{sedge}, \texttt{lily}, \texttt{rose}, \texttt{tulip}, \texttt{daisy}, \texttt{sunflower}, \texttt{orchid}, \texttt{poppy}, \texttt{lavender}, \texttt{thistle}, \texttt{mint}, \texttt{basil}, \texttt{thyme}, \texttt{sage}, \texttt{parsley}, \texttt{rosemary}, \texttt{chives}, \texttt{oregano}, \texttt{cilantro}, \texttt{spinach}, \texttt{lettuce}, \texttt{cabbage}, \texttt{broccoli}, \texttt{kale}, \texttt{plant}, \paragraph{Food}
\texttt{food}, \texttt{fruit}, \texttt{apple}, \texttt{banana}, \texttt{orange}, \texttt{lemon}, \texttt{lime}, \texttt{grape}, \texttt{cherry}, \texttt{strawberry}, \texttt{blueberry}, \texttt{raspberry}, \texttt{blackberry}, \texttt{pineapple}, \texttt{watermelon}, \texttt{cantaloupe}, \texttt{honeydew}, \texttt{kiwi}, \texttt{peach}, \texttt{plum}, \texttt{pear}, \texttt{apricot}, \texttt{mango}, \texttt{papaya}, \texttt{coconut}, \texttt{avocado}, \texttt{tomato}, \texttt{cucumber}, \texttt{bell pepper}, \texttt{carrot}, \texttt{celery}, \texttt{broccoli}, \texttt{cauliflower}, \texttt{cabbage}, \texttt{lettuce}, \texttt{spinach}, \texttt{kale}, \texttt{arugula}, \texttt{chard}, \texttt{beet}, \texttt{radish}, \texttt{turnip}, \texttt{potato}, \texttt{sweet potato}, \texttt{yam}, \texttt{cassava vegetable}, \texttt{taro vegetable}, \texttt{pumpkin}, \texttt{squash}, \texttt{zucchini}, \texttt{eggplant}, \texttt{pea}, \texttt{bean}, \texttt{lentil}, \texttt{chickpea}, \texttt{peanut}, \texttt{almond}, \texttt{cashew}, \texttt{pistachio}, \texttt{walnut}, \texttt{pecan}, \texttt{hazelnut}, \texttt{macadamia}, \texttt{peanut}, \texttt{pistachio}, \texttt{almond}, \texttt{cashew}, \texttt{walnut}, \texttt{pecan}, \texttt{hazelnut}, \texttt{macadamia}, \texttt{peanut}, \texttt{pistachio}, \texttt{almond}, \texttt{cashew}, \texttt{walnut}, \texttt{pecan}, \texttt{hazelnut}, \texttt{macadamia}, \texttt{peanut}, \texttt{pistachio}, \texttt{almond}, \texttt{cashew}, \texttt{walnut}, \texttt{pecan}, \texttt{hazelnut}, \texttt{macadamia}, \texttt{peanut}, \texttt{pistachio}, \texttt{almond}, \texttt{cashew}, \texttt{walnut}, \texttt{pecan}, \texttt{hazelnut}, \texttt{macadamia}, \texttt{peanut}, \texttt{pistachio}, \texttt{almond}, \texttt{cashew}, \texttt{walnut}, \texttt{pecan}, \texttt{hazelnut}, \texttt{macadamia}, \texttt{peanut}, \texttt{pistachio}, \texttt{almond}, \texttt{cashew}, \texttt{walnut}, 
}

\subsection{UMAP Embeddings}
\label{app:exp:describe:umap}
\begin{figure}[t]
    \centering
    \includegraphics[width=0.99\textwidth]{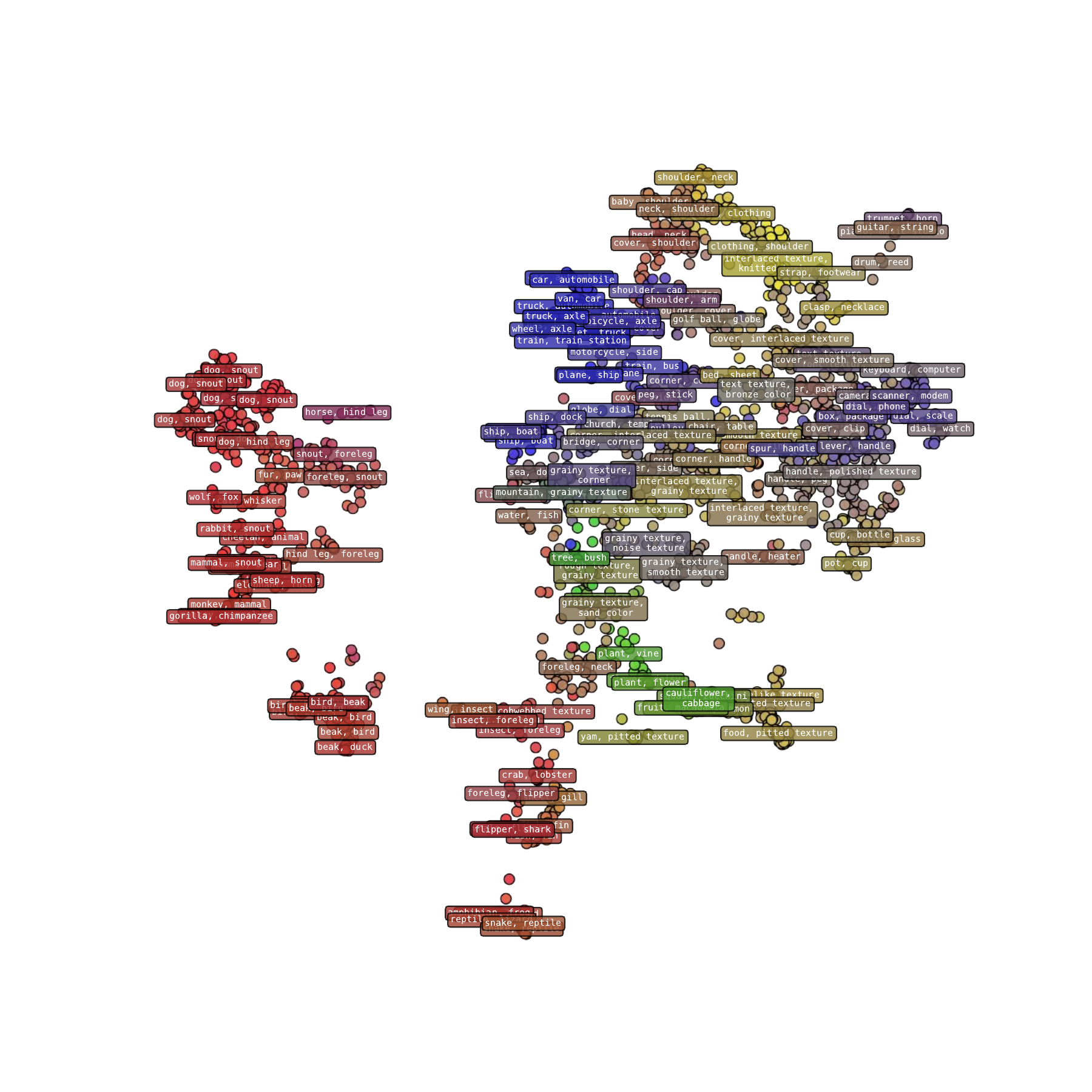}
    \caption{UMAP embedding of a ResNet50v2's semantic representation, where points are colour-coded according to the concepts similarity to \texttt{vegetation, plant, tree} (green), \texttt{animal, living} (red), \texttt{vehicle} (blue), and \texttt{texture, pattern} (yellow). 
    We further provide the top two labels for each of the 160 clusters.
    }
    \label{fig:app:explore:describe:umap_resnet50v2}
\end{figure}

\begin{figure}[t]
    \centering
    \includegraphics[width=0.99\textwidth]{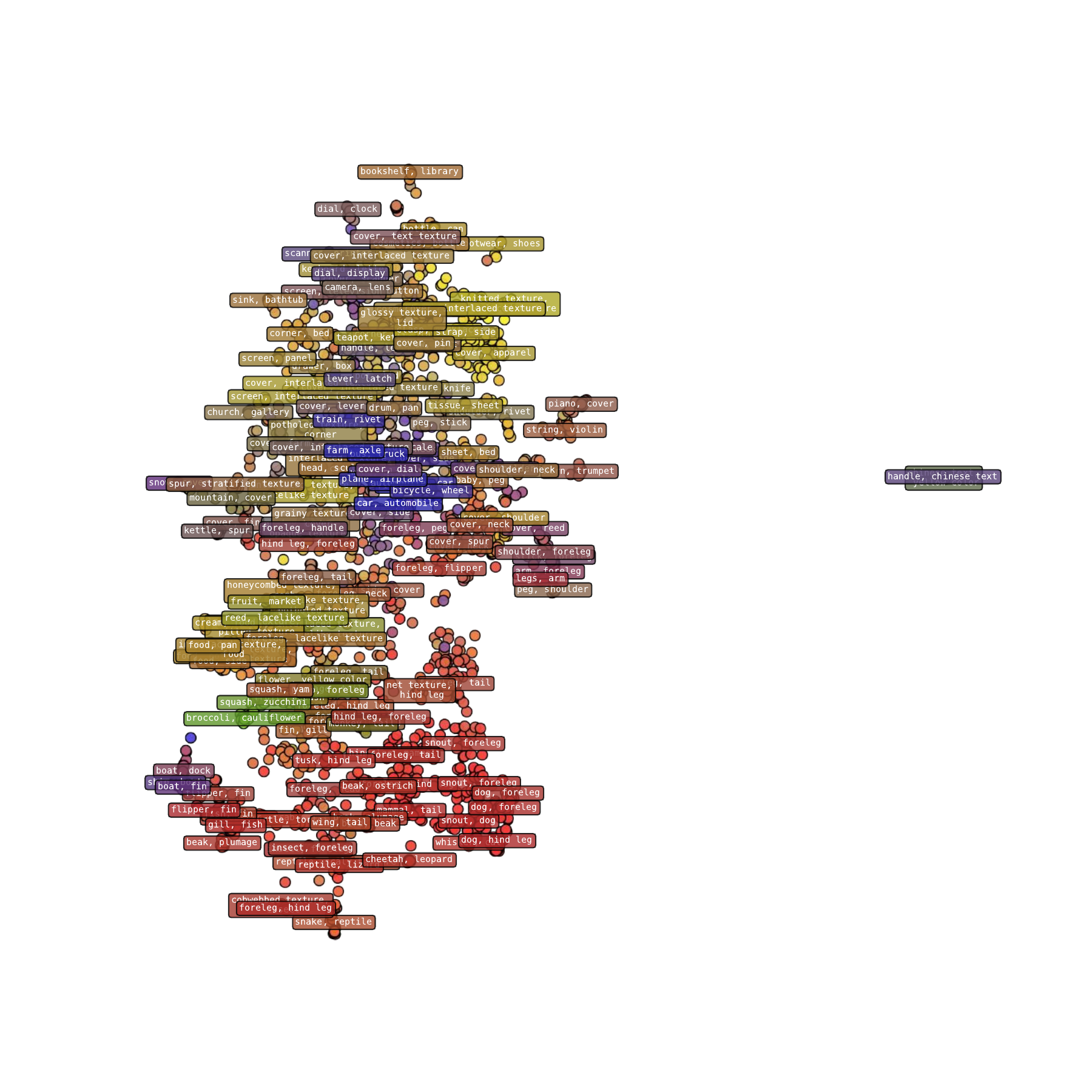}
    \caption{UMAP embedding of a ResNet50's semantic representation, where points are colour-coded according to the concepts similarity to \texttt{vegetation, plant, tree} (green), \texttt{animal, living} (red), \texttt{vehicle} (blue), and \texttt{texture, pattern} (yellow). 
    We further provide the top two labels for each of the 160 clusters.
    An outlier cluster corresponding to a well-known spurious correlation (\texttt{Chinese text}) in the ImageNet dataset is visible.
    }
    \label{fig:app:explore:describe:umap_resnet50v1}
\end{figure}

\begin{figure}[t]
    \centering
    \includegraphics[width=0.99\textwidth]{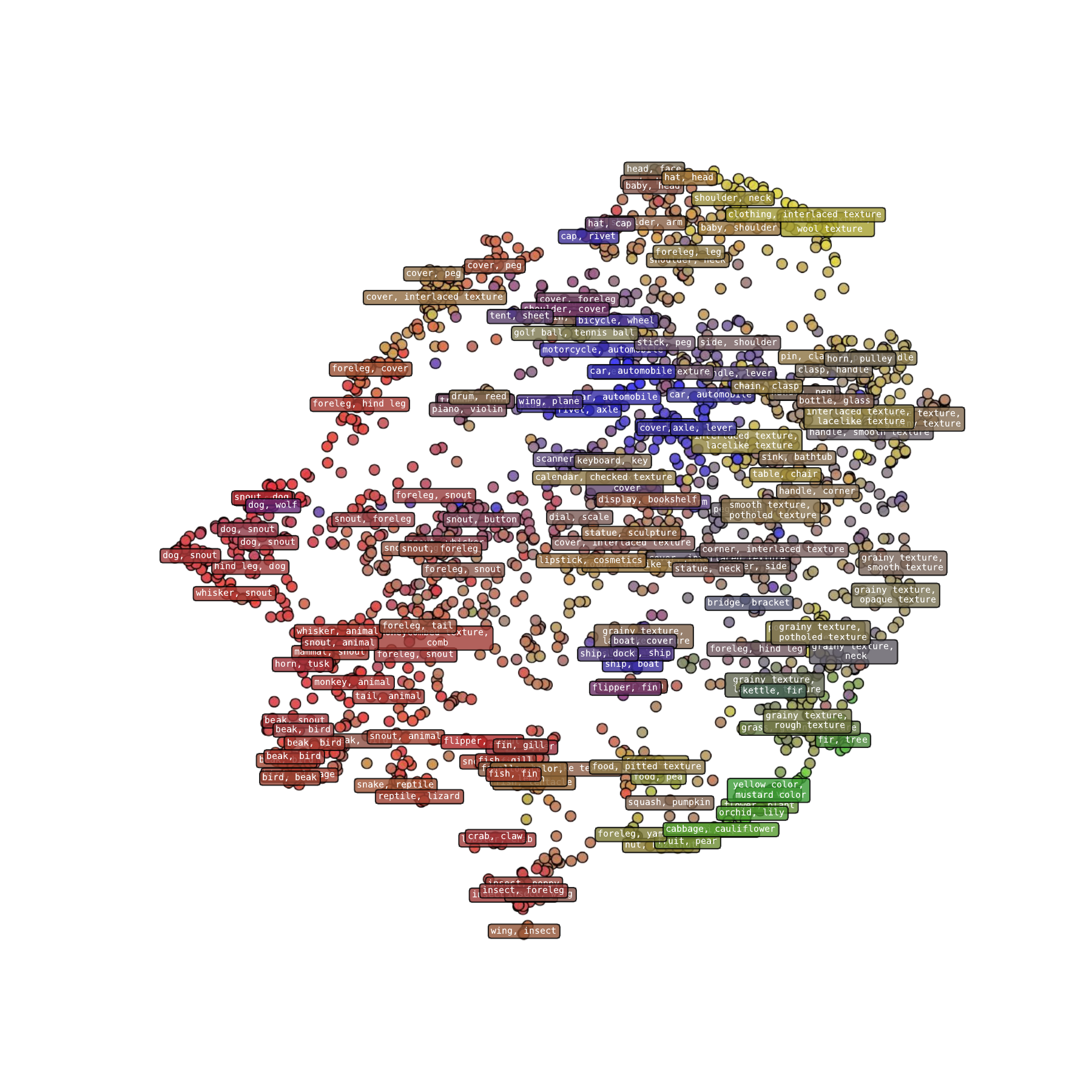}
    \caption{UMAP embedding of a ResNet50.a1's semantic representation, where points are colour-coded according to the concepts similarity to \texttt{vegetation, plant, tree} (green), \texttt{animal, living} (red), \texttt{vehicle} (blue), and \texttt{texture, pattern} (yellow). 
    We further provide the top two labels for each of the 160 clusters.
    }
    \label{fig:app:explore:describe:umap_resnet50_timm}
\end{figure}

As shown in \cref{fig:results:describe}a,
a UMAP~\cite{mcinnes2018umap} embedding can be effective to gain an overview of the concepts encoded by the model components.
For ImageNet models,
we use the 50 maximally activating images as concept examples per neuron, which are embedded by the CLIP-Mobile-S2 foundation model to receive semantic embeddings \gls{cone}.
Further,
we perform $k$-means clustering with $k{=}60$ for \cref{fig:results:describe}a, and $k{=}160$ for the additional examples shown in \cref{fig:app:explore:describe:umap_resnet50v2,fig:app:explore:describe:umap_resnet50v1,fig:app:explore:describe:umap_resnet50_timm} for the ResNet50v2, ResNet50, and ResNet50.a1, respectively.
The points in semantic space are colour-coded corresponding to their similarity to the concepts of \texttt{vegetation, plant, tree} (green), \texttt{animal, living} (red), \texttt{vehicle} (blue), and \texttt{texture, pattern} (yellow). 
In order to label the clusters,
we retrieve the two most aligned concept-labels for the average \gls{cone} of a cluster.

For all models,
concepts related to \texttt{vegetation, plant, tree}, \texttt{animal, living}, \texttt{vehicle}, and \texttt{texture, pattern} can be seen.
However, there are also differences in structure.
For example,
the ResNet50 embedding shows a clear outlier which refers to \texttt{Chinese text} that is known to be present as a spurious correlation in the ImageNet dataset.

\subsection{Network Dissection}
\label{app:explore:describe:dissection}
As described in \cref{sec:results:explore:describe},
we can group labelled neurons into parent categories (if these parent categories are available),
referred to as Network Dissection~\cite{bau2017network}.
The concept-labels and parent categories used in this work are listed in \cref{app:explore:describe:labels}.

For labelling of neurons,
we use text embedding templates of ``\texttt{<concept>}'', ``a \texttt{<concept>}'', ``\texttt{<concept>}-like'', and ``an image of a close-up of \texttt{<concept>}'' where \texttt{<concept>} is replaced by the actual concept-label.
To compute alignment scores $a$ we further subtract the alignment to using an empty template.
Finally,
all neurons below an alignment of $a<0.025$ are filtered out.

\begin{figure}[t]
    \centering
    \includegraphics[width=0.99\textwidth]{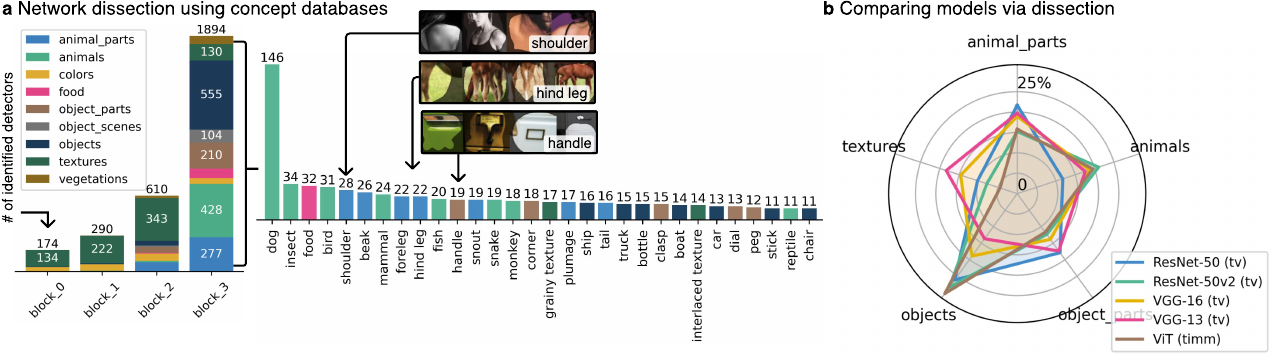}
    \caption{Dissection of neural networks to quantitatively understand and compare representations.
    \textbf{a})
    For all layers of a ResNet50v2, 
    we group components (\ie, neurons) into parent categories to gain an overview over the learned concepts.
    Overall,
    higher-level concepts are more dominant in later layers,
    whereas lower-level concepts such as textures are more dominant in earlier layers.
    The most assigned concept refers is \texttt{dog} with over 146 neurons of 2048.
    \textbf{b})
    We can also compare the relative shares of neurons in parent categories across models.
    More extensively trained models, and deeper architectures show relatively more higher-level concepts in the last layer (\eg, ResNet50 vs. ResNet50v2, and VGG-13 vs. VGG-16).
    }
    \label{fig:app:explore:describe:dissection}
\end{figure}
An example for the resulting number of neurons per parent category in a ResNet50v2 for all layers is shown in \cref{fig:app:explore:describe:dissection}a.
The example shows, that relatively more texture and colour-related concepts exist in lower-level layers,
and more higher-level concepts such as ``animals'' or ``objects'' in higher-level layers.
We further illustrate that one can inspect the most-matched concepts further in \cref{fig:app:explore:describe:dissection}a (\emph{middle}),
which correspond to \texttt{dog}, \texttt{insect}, \texttt{food}, and \texttt{bird} with 146, 34, 32, and 31 neurons in the last layer, respectively.

The relative share of neurons in the parent categories can be also compared across models,
as shown in \cref{fig:app:explore:describe:dissection}b,
where the relative shares of a ResNet50, ResNet50v2, VGG-16, and VGG-13 are depicted.
Comparing the ResNet50 and ResNet50v2,
it is apparent that the ResNet50v2 has learned more higher-level concepts (\eg, ``animal'') and less lower-level concepts (\eg, ``animal parts'').
This can be expected, 
as the ResNet50v2 is trained more extensively, leading to more class-specific (corresponding to ``object'' and ``animal'') concepts in  later layers~\cite{liu2022towards,fel2024understanding}.
A similar trend can also be seen when comparing the smaller VGG-13 and larger VGG-16 model, where the larger one shows more higher-level concepts.

\begin{figure}[t]
    \centering
    \includegraphics[width=0.99\textwidth]{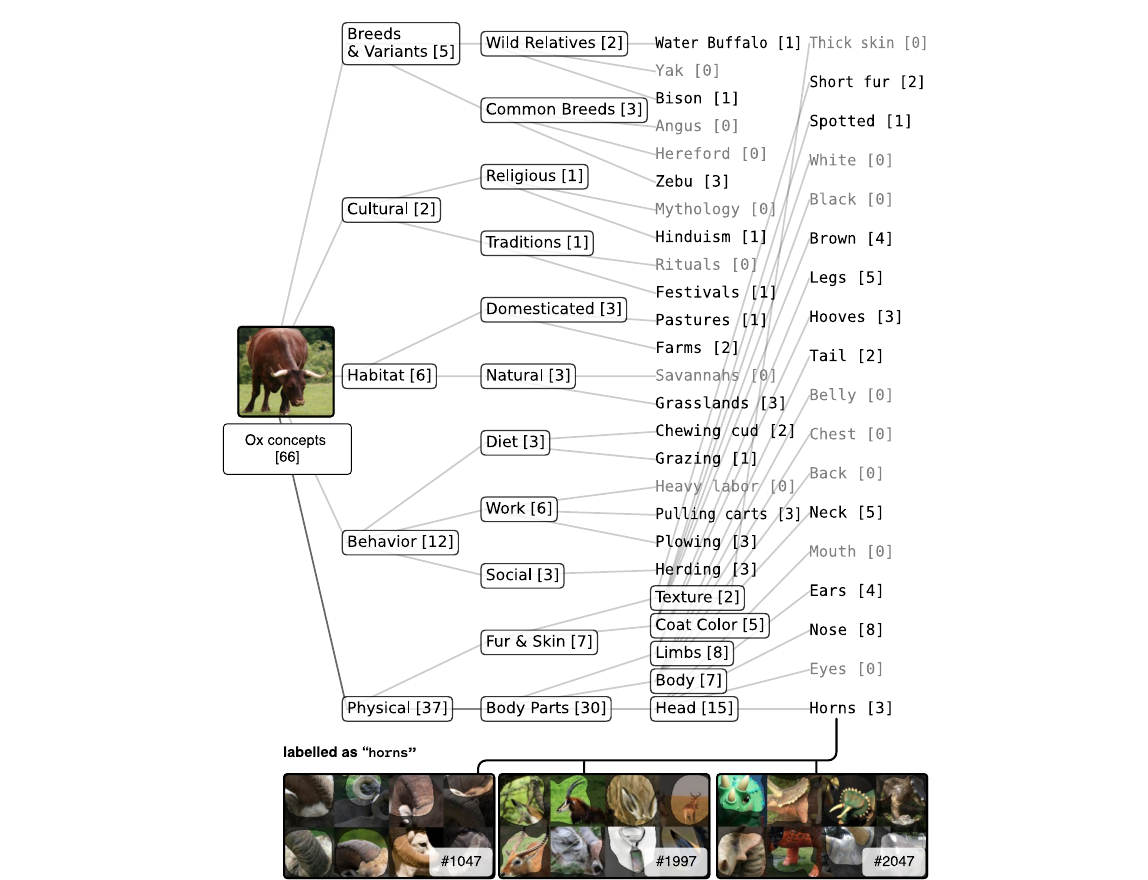}
    \caption{Grouping of concepts in parent categories for the ``Ox'' class of ImageNet using a ResNet50v2 model.
    Having multiple levels of parent categories available allows to structure and simplify the knowledge encoded by a model.
    For examples,
    we can identify three neurons (number of neurons per concept/group is given in brackets) belonging to \texttt{horns},
    which are hierarchically grouped into ``Head'', ``Body Parts' and ``Physical''.
    }
    \label{fig:app:explore:describe:dissection_ox}
\end{figure}

Another example
of network dissection is shown in \cref{fig:app:explore:describe:dissection_ox}.
Here,
we
group concepts into parent categories for the ``Ox'' class of ImageNet using a ResNet50v2 model.
Having multiple levels of parent categories available allows to structure and simplify the knowledge encoded by a model.
For examples,
we can identify three neurons (number of neurons per concept/group is given in brackets) belonging to \texttt{horns},
which are hierarchically grouped into ``Head'', ``Body Parts' and ``Physical''.

\subsection{Evaluation of Concept Labelling}
\label{sec:app:explore:describe:evaluation}

In this section, we first compare the labelling abilities of \gls{ours} with two recent neuron-labelling methods, \gls{invert}~\cite{bykov2024labeling} and \gls{clip-dissect}~\cite{oikarinen2022clip}, using an evaluation procedure adapted from \cite{kopf2024cosy}, which analyses neuron activation patterns on synthetic concept examples. 
Secondly, we conduct an ablation study on key hyperparameters, including the use of prompt templates, the number of concept examples used for semantic embedding computation, and the activation pooling strategy applied during concept example sampling.

\subsubsection{Evaluation Procedure}
\label{sec:app:explore:describe:evaluation:eval_procedure}

The lack of ground truth information makes evaluating explanations a challenging endeavour.
For our specific case of neuron labelling, we adapt the recently proposed procedure in \cite{kopf2024cosy} to assess the faithfulness of a label.
Concretely, 
the evaluation assumes that a good label for a neuron, when represented as model inputs, induces high activation in the neuron.
To test this, we utilize a text-to-image model to generate a set of synthetic concept examples for each concept-label $k\in{T}$, which we denote by $C_k=\{x_1,...,x_n\}$ where  $T$ is the set of labels used during the labelling process. 
With this synthetic test set we can define the response of a neuron $i$ (given by $\mathcal{M}_i$) to the concept $k$ as 
$$
a_i(k)=\tfrac{1}{|C_k|}\sum_{x\in C_k}\mathcal{M}_i(x)
$$
and score a neuron labelling $(i,k)$ via
\begin{equation}
\label{eq:app:describe:benchmark:score}
\phi(i,k)\coloneqq \frac{a_i(k) - \min_t a_i(t)}{\max_t a_i(t)- \min_t a_i(t)} \in[0,1]
\end{equation}
where more faithful labels will result in a score closer to 1.
In the following study, we report the mean and standard error of this score across multiple neuron-label pairs for each method under investigation.

\subsubsection{Benchmark}
\label{sec:app:explore:describe:evaluation:benchmark}

The benchmark study presented in this section aims to validate the faithfulness of the labels derived via \gls{ours} across a variety of settings. 
Specifically, we evaluate the investigated methods - \gls{ours}, \gls{invert} and \gls{clip-dissect} - on the first 200 neurons in the final feature layer of five distinct \gls{imagenet} classifiers: ResNet50v2, ResNet101v2, VGG13, VGG16 and VGG19. 
For this,
we use concept-labels and activations extracted from three different datasets, namely \gls{imagenet}, \gls{broden} \cite{broden2017bau} and \gls{paco} \cite{paco2023}.
For \gls{imagenet}, 
the 1k class names (\eg, ``coffee mug'', ``strawberry'') are used, while for \gls{broden} and \gls{paco} we extract 1.344 and 531 concept-labels, respectively, using the provided finer-grained segmentation annotations such as object parts (\eg, ``mug handle''), colours (\eg, ``red colour''), and textures (\eg, ``striped texture'').
The synthetic concept example sets $C_k$ used in the evaluation are generated using the openly available Stable Diffusion 1.5v model \cite{rombach2022high} and the prompt template ``photograph of a \texttt{<concept>}'', with each set comprising 40 images.
Prompt templates are also used for generating the concept-label embeddings
for \gls{ours} and \gls{clip-dissect} via the CLIP-Mobile-S2 model. Specifically, we used the average embedding for the templates ``\texttt{<concept>}'', ``\texttt{<concept>}-like'', ``a \texttt{<concept>}'' and ``an image of a close up of \texttt{<concept>}''. 
For sampling activation-maximizing concept examples of neurons in \gls{ours} and \gls{clip-dissect}, we follow the approach in \cite{oikarinen2022clip} and apply mean pooling across the spatial dimensions of the neuron activations.
The number of activation-maximizing samples in both methods are set to 30.
The effect of hyperparameters on the labelling quality of \gls{ours} are studied in \cref{sec:app:describe:benchmark:ablation_study}.

The benchmarking results in \cref{app:tab:descibe:benchmark} demonstrate that \gls{ours} is a competitive neuron-labelling method, consistently outperforming the random baseline across the investigated datasets and networks. It performs particularly well on ResNet architectures, achieving high scores on \gls{imagenet} and \gls{paco}, such as $0.774 \pm 0.022$ on ResNet101 for \gls{imagenet}, closely matching the performance of \gls{clip-dissect} and \gls{invert}. 
Notably, on \gls{paco}, \gls{ours} outperforms the other methods on all investigated models except VGG19. This strong performance may be attributed to the practice of cropping concept examples to the relevant image regions, which gives an advantage for highly localized concept-labels, such as object parts that occupy only a small portion of the image and may be overlooked by other methods that use the full activation-maximizing samples during labelling.
However, on VGG networks, \gls{ours} struggles with larger label spaces, such as \gls{broden}, often scoring below \gls{clip-dissect} and \gls{invert}. 
\begin{table}[h]
    \centering
\caption{Comparison of labelling Methods. This table presents the average quality scores (with standard error) as defined in \cref{eq:app:describe:benchmark:score} for the evaluated methods. The scores were averaged across 200 neurons from the final layers of the studied classifiers. For \gls{clip-dissect} and \gls{ours}, the results also account for the employed foundation models. Both mean and max pooling strategies are reported for activation-maximizing sample selection. Higher scores indicate better performance. \gls{ours} achieves results comparable to the other studied methods.}
    \begin{tabular}{lllll}
\toprule
 &  & \multicolumn{3}{c}{Score $\phi$ ($\uparrow$)} \\
 \cmidrule{3-5}
Model & Method & Paco & Broden & ImageNet  \\
\midrule
\multirow[t]{4}{*}{ResNet101v2} & \textit{Random}       & \textit{0.141$\pm$0.004}  & \textit{0.092$\pm$0.003}  & \textit{0.056$\pm$0.003}  \\
                                & \gls{invert}          & 0.280$\pm$0.017           & 0.438$\pm$0.021           & \textbf{0.809$\pm$0.020}  \\
                                & \gls{clip-dissect}    & 0.485$\pm$0.022           & \textbf{0.588$\pm$0.023}  & 0.736$\pm$0.024           \\
                                & \gls{ours}            & \textbf{0.508$\pm$0.022}  & 0.551$\pm$0.023           & 0.774$\pm$0.022           \\
\cmidrule{1-5}
\multirow[t]{4}{*}{ResNet50v2}  & \textit{Random}       & \textit{0.129$\pm$0.004}  & \textit{0.086$\pm$0.003}  & \textit{0.055$\pm$0.002}  \\
                                & \gls{invert}          & 0.279$\pm$0.018           & 0.343$\pm$0.022           & \textbf{0.797$\pm$0.019}  \\
                                & \gls{clip-dissect}    & 0.448$\pm$0.023           & \textbf{0.490$\pm$0.026}  & 0.713$\pm$0.024           \\
                                & \gls{ours}            & \textbf{0.488$\pm$0.023}  & 0.448$\pm$0.026           & 0.754$\pm$0.022           \\
\cmidrule{1-5}
\multirow[t]{4}{*}{VGG13}       & \textit{Random}       & \textit{0.251$\pm$0.007}  & \textit{0.209$\pm$0.006}  & \textit{0.200$\pm$0.006}  \\
                                & \gls{invert}          & 0.437$\pm$0.019           & 0.474$\pm$0.020           & \textbf{0.682$\pm$0.017}  \\
                                & \gls{clip-dissect}    & 0.487$\pm$0.019           & \textbf{0.541$\pm$0.021}  & 0.679$\pm$0.017           \\
                                & \gls{ours}            & \textbf{0.534$\pm$0.019}  & 0.393$\pm$0.022           & 0.583$\pm$0.018           \\
\cmidrule{1-5}
\multirow[t]{4}{*}{VGG16}       & \textit{Random}       & \textit{0.240$\pm$0.006}  & \textit{0.196$\pm$0.005}  & \textit{0.179$\pm$0.005}  \\
                                & \gls{invert}          & 0.382$\pm$0.018           & 0.465$\pm$0.018           & \textbf{0.700$\pm$0.016}  \\
                                & \gls{clip-dissect}    & 0.476$\pm$0.020           & \textbf{0.493$\pm$0.022}  & 0.694$\pm$0.017           \\
                                & \gls{ours}            & \textbf{0.508$\pm$0.019}  & 0.394$\pm$0.021           & 0.635$\pm$0.019           \\
\cmidrule{1-5}
\multirow[t]{4}{*}{VGG19}       & \textit{Random}       & \textit{0.216$\pm$0.006}  & \textit{0.175$\pm$0.005}  & \textit{0.157$\pm$0.004}  \\
                                & \gls{invert}          & 0.399$\pm$0.018           & 0.526$\pm$0.018           & 0.701$\pm$0.017           \\
                                & \gls{clip-dissect}    & \textbf{0.511$\pm$0.020}  & \textbf{0.582$\pm$0.019}  & \textbf{0.718$\pm$0.017}  \\
                                & \gls{ours}            & 0.496$\pm$0.020           & 0.456$\pm$0.021           & 0.609$\pm$0.020           \\
\bottomrule
\end{tabular}

        \label{app:tab:descibe:benchmark}
\end{table}
\cref{app:tab:descibe:benchmark:layers} shows the results across the investigated layers in the ResNet50v2 model, where neurons in the final layer of each of the four ResNet blocks were labelled. The results are consistent with those presented in \cref{app:tab:descibe:benchmark}.
We observe that the random baseline score for Block 0 (first feature layer) and Block 1 (second feature layer) is close to 0.5, suggesting that the set of potential concept labels may not be well-suited for these layers. This leads to a small gap between the highest and lowest possible neuron responses across the concept-labels, 
$(\max_t a_i(t)-\min_t a_i(t))$, resulting in an artificially high score for the random label assignment.
This aligns with previous results suggestion that neurons in earlier layers are more difficult to label \cite{kopf2024cosy}.
\begin{table}[t!]
    \centering
\caption{
Comparison of labelling methods applied to four layers of varying depth. 
This table presents the average quality scores (with standard error), as defined in \cref{eq:app:describe:benchmark:score}, for the investigated methods applied to the first 200 neurons across the four ResNet blocks of the ResNet50v2 model. For methods requiring embeddings, we use the CLIP-Mobile-S2 model with prompt templates. 
We observe, higher scores of the random baseline in earlier layers (Block 0 and Block 1), likely because the concept labels are not well suited to neurons in these layers. This reduces the difference between the maximal and minimal neuron responses, inflating the random assignment scores (see \cref{eq:app:describe:benchmark:score}).
\gls{invert} dominates the lower layers on \gls{broden} and \gls{imagenet}, whereas on \gls{paco}, \gls{ours} achieves the best performance in Block 3 and performs competitively with \gls{clip-dissect}, which leads in Block 1 and Block 0.
}
    
\begin{tabular}{llllll}
\toprule
 &  & \multicolumn{4}{c}{Score $\phi$ ($\uparrow$)} \\
 \cmidrule{3-6}
 & Method & Block 0 & Block 1 & Block 2 & Block 3  \\
\midrule
\parbox[t]{2mm}{\multirow{4}{*}{\rotatebox[origin=c]{90}{Broden}}} 
 & \textit{Random} & \textit{0.505$\pm$0.009} & \textit{0.509$\pm$0.008} & \textit{0.389$\pm$0.007} & \textit{0.086$\pm$0.003} \\
 & \gls{invert} & \textbf{0.595$\pm$0.014} & \textbf{0.614$\pm$0.016} & \textbf{0.554$\pm$0.016} & 0.343$\pm$0.022 \\
 & \gls{clip-dissect} & 0.565$\pm$0.012 & 0.508$\pm$0.014 & 0.524$\pm$0.016 & \textbf{0.490$\pm$0.026} \\
 & \gls{ours} & 0.534$\pm$0.012 & 0.536$\pm$0.012 & 0.457$\pm$0.015 & 0.448$\pm$0.026 \\
\cmidrule{1-6}
\parbox[t]{2mm}{\multirow{4}{*}{\rotatebox[origin=c]{90}{ImageNet}}} 
 & \textit{Random} & \textit{0.482$\pm$0.009} & \textit{0.506$\pm$0.01} & \textit{0.378$\pm$0.008} & \textit{0.055$\pm$0.002} \\
 & \gls{invert} & \textbf{0.648$\pm$0.014} & \textbf{0.680$\pm$0.013} & \textbf{0.661$\pm$0.01}4 & \textbf{0.797$\pm$0.019} \\
 & \gls{clip-dissect} & 0.583$\pm$0.014 & 0.624$\pm$0.015 & 0.609$\pm$0.015 & 0.713$\pm$0.024 \\
 & \gls{ours} & 0.577$\pm$0.011 & 0.583$\pm$0.014 & 0.516$\pm$0.015 & 0.754$\pm$0.022 \\
\cmidrule{1-6}

\parbox[t]{2mm}{\multirow{4}{*}{\rotatebox[origin=c]{90}{Paco}}} 
 & \textit{Random} & \textit{0.484$\pm$0.007} & \textit{0.494$\pm$0.008} & \textit{0.381$\pm$0.007} & \textit{0.129$\pm$0.004} \\
 & \gls{invert} & 0.497$\pm$0.012 & 0.534$\pm$0.014 & 0.435$\pm$0.015 & 0.279$\pm$0.018 \\
 & \gls{clip-dissect} & \textbf{0.534$\pm$0.013} & \textbf{0.579$\pm$0.014} & 0.518$\pm$0.017 & 0.448$\pm$0.023 \\
 & \gls{ours} & 0.509$\pm$0.013 & 0.556$\pm$0.014 & \textbf{0.542$\pm$0.015} & \textbf{0.488$\pm$0.023} \\
\bottomrule
\end{tabular}

    \label{app:tab:descibe:benchmark:layers}
\end{table}

\subsubsection{Ablation Study}
\label{sec:app:describe:benchmark:ablation_study}

In this section, we investigate how various hyperparameters influence the labelling performance of \gls{ours} and \gls{clip-dissect}. Specifically, we analyse the choice of foundation models, the use of prompt templates, the pooling strategy applied during activation maximization (ActMax), and the number of concept examples used within \gls{ours} to derive the semantic embeddings of neurons. 
Our goal is to identify configurations that yield robust and accurate neuron labels.

\begin{figure}[t]
    \centering
    \includegraphics[width=\textwidth]{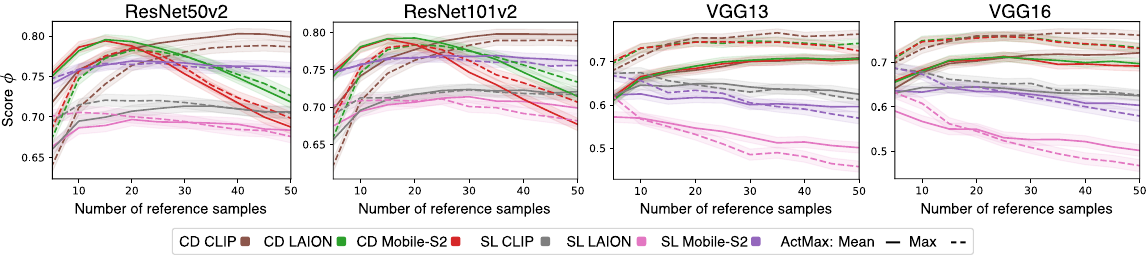}
    \caption{
    Evaluation of the influence of the count of ActMax samples on the labelling abilities of \gls{ours}.
    We evaluate the labels derived via \gls{ours} and \gls{clip-dissect} for neurons in the final feature layers of four different neural networks, as defined in \cref{eq:app:describe:benchmark:score}, using varying number of reference samples. The average score over all neurons in the inspected layer is plotted on the $y$-axis.
    Concept labels and ActMax samples were derived using \gls{imagenet} class names and test data. 
    We see distinct effects of the number of used samples on the performance of \gls{ours} between ResNet and VGG models. In case of ResNet we see a steep accent in performance up to 15 samples, stabilizing thereafter with a slight decline near the maximum of 50 samples.
    In contrast, for both VGG models we observe a monotone decrease of the overall performance, peaking with 5 reference samples and gradually declining as more samples are used.
    }
    \label{fig:app:explore:describe:eval:ablation}
\end{figure}

\paragraph{Setup}
For all ablation experiments, we use the same evaluation metric $\phi$ described in \cref{eq:app:describe:benchmark:score}. 
The results are averaged across the five neural network architectures considered in the main benchmark and are reported along the standard error. 
We begin by examining the effects of the foundation model, prompt template, and pooling strategies, as summarized in \cref{app:tab:describe:eval:ablation}. After that, we present a detailed analysis of how the number of concept examples samples affects labelling performance, shown in \cref{fig:app:explore:describe:eval:ablation}.

\paragraph{Effect of the Foundation Models}
\cref{app:tab:describe:eval:ablation} compares the performance when using different foundation models (CLIP-OpenAI, CLIP-LAION, and CLIP-Mobile-S2). 
We find that \gls{clip-dissect} achieves its highest overall score ($0.588$) with CLIP-OpenAI, also used in the original paper \cite{oikarinen2022clip}. 
In contrast, \gls{ours} attains its best performance ($0.542$) with CLIP-Mobile-S2, indicating that the lighter-weight, mobile-tailored model provides embeddings that better complement \gls{ours}’s labelling pipeline. 
LAION embeddings consistently underperform compared to the other two models for both methods.

\paragraph{Effect of the Prompt Template}
The use of a descriptive prompt template, ``a photograph of a \texttt{<concept label>}'', consistently improves labelling faithfulness. 
Both \gls{clip-dissect} and \gls{ours} benefit from templates across all foundation models and pooling strategies. 
However, the improvement for \gls{clip-dissect} is modest — in its best-performing configuration with the CLIP-OpenAI model, the score increases from $0.578$ to $0.588$ — whereas \gls{ours} shows more substantial gains, improving from $0.497$ to $0.542$. 
Further optimization of the prompt template could yield additional improvements in labelling accuracy.

\paragraph{Effect of the Pooling Strategy during Activation Maximization}
We evaluate both \emph{max} and \emph{mean} pooling strategies during the activation maximization for obtaining the neurons concept examples. \gls{clip-dissect}’s performance does not significantly change regardless of the pooling strategy. \gls{ours}, on the other hand, exhibits a slight but consistent improvement when using mean pooling (\eg, from $0.537$ to $0.542$ with CLIP-Mobile-S2 and template). The mean strategy may provide a more representative snapshot of neuron activation patterns, enhancing \gls{ours}’ ability to infer accurate labels.
\begin{table}[ht]
    \centering
    \caption{ 
    Influence of hyperparameters. 
    This table presents the labelling scores of \gls{ours} and \gls{clip-dissect} average across the investigated three datasets and five neural networks 
    under various hyperparameter selections. 
    Higher values are better. 
    Higher scores indicate better performance. Both methods perform best with the prompt template ``photograph of a \texttt{<concept label>}'' across all foundation models. While the optimal activation aggregation strategy (ActMax in the table) varies by model, its overall impact is limited. \gls{ours} achieves the highest performance with CLIP-Mobile-S2, while \gls{clip-dissect} performs best with CLIP-OpenAI.
    }
    \begin{tabular}{llllll}
\toprule
 &  &  & \multicolumn{3}{c}{Score $\phi$ ($\uparrow$)} \\
 \cmidrule{4-6}
Method & ActMax & Template & CLIP & LAION & Mobile-S2  \\
\midrule
\multirow[t]{4}{*}{\gls{clip-dissect}} & \multirow[t]{2}{*}{Max} & False  & 0.578$\pm$0.038 & 0.573$\pm$0.039 & 0.576$\pm$0.034 \\
 &  & True  & \textbf{0.588$\pm$0.039} & 0.582$\pm$0.036 & 0.581$\pm$0.034 \\
\cmidrule{2-6}
 & \multirow[t]{2}{*}{Mean} & False  & 0.577$\pm$0.030 & 0.569$\pm$0.030 & 0.576$\pm$0.027 \\
 &  & True   & \textbf{0.586$\pm$0.029} & 0.580$\pm$0.029 & 0.576$\pm$0.027 \\
\cmidrule{1-6} 
\multirow[t]{4}{*}{\gls{ours}} & \multirow[t]{2}{*}{Max} & False  & 0.464$\pm$0.034 & 0.423$\pm$0.023 & 0.500$\pm$0.037 \\
 &  & True  & 0.523$\pm$0.031 & 0.480$\pm$0.025 & \textbf{0.537$\pm$0.028} \\
\cmidrule{2-6}
 & \multirow[t]{2}{*}{Mean} & False & 0.454$\pm$0.036 & 0.435$\pm$0.024 & 0.497$\pm$0.039 \\
 &  & True  & 0.518$\pm$0.031 & 0.491$\pm$0.024 & \textbf{0.542$\pm$0.029} \\
\bottomrule
\end{tabular}

    \label{app:tab:describe:eval:ablation}
\end{table}

\paragraph{Effect of the Number of ActMax Samples}
In addition to embedding choice, templates, and pooling strategies, the number of ActMax samples plays a critical role for the labelling performance.
\cref{fig:app:explore:describe:eval:ablation} shows how 
the labelling score $\phi$ changes as we vary the number of ActMax samples from five to 50 across ResNet (ResNet50v2 and ResNet101v2) and VGG (VGG13 and VGG16) architectures.
Here again, distinct trends emerge between these two architecture families, with method-specific variations observed for \gls{ours} and \gls{clip-dissect}.

For ResNet models, performance generally improves as the number of samples increases. 
For \gls{ours}, performance stabilizes around 20 samples, suggesting that sufficient activation coverage is achieved, with additional samples providing no further gains. 
In contrast, \gls{clip-dissect} exhibits varying behaviour depending on the foundation model.
With CLIP-LAION and CLIP-Mobile-S2, performance peaks at approximately 15 samples before declining linearly, eventually falling below the initial scores observed at 5 samples. 
Interestingly, \gls{clip-dissect} with the CLIP-OpenAI model consistently improves as the number of samples increases, reaching its highest performance at 50 samples.

For VGG models, 
\gls{ours} shows performance degradation when using the max pooling strategy, as the number of ActMax samples increases. 
However, 
with mean pooling and the CLIP-Mobile-S2 or CLIP-OpenAI models, 
performance remains fairly stable across varying sample counts. 
One possible hypothesis for this behaviour is that VGG neurons are inherently less interpretable, leading to concept examples of lower clarity, as observed in \cref{app:exp:audit,fig:results:eval}. 
As a consequence, the semantic embeddings may become increasingly noisy or diffuse with the number of samples, making it harder for \gls{ours} to assign faithful concept labels. The mean pooling strategy seems to mitigate this effect to some extent.
In contrast, \gls{clip-dissect} performance initially improves with increased number of samples and then stabilizes. 
A plausible explanation is that, unlike \gls{ours}, \gls{clip-dissect} determines concept assignments by maximizing the pointwise mutual information between the neurons activations on the concept samples and candidate concept labels. This process inherently assigns probabilistic weights to individual samples, rather than treating their contributions as binary, as is the case in \gls{ours}. This probabilistic weighting likely acts as a form of denoising, reducing the impact of ambiguous or noisy activations.

Based on these observations, a potential improvement for \gls{ours} could involve incorporating weighted averaging strategies when computing semantic embeddings for neurons. By adopting a mechanism akin to the SoftWPMI approach used in \gls{clip-dissect}, where ActMax samples influence the embedding process non-uniformly, \gls{ours} could achieve greater robustness, particularly in architectures like VGG, where neuron interpretability is lower.

Overall, the ablation study highlights the importance of hyperparameter selection for \gls{ours}, which achieves its best performance with CLIP-Mobile-S2 embeddings, prompt templates, and mean pooling. Notably, the effect of the number of used concept examples on labelling faithfulness of varies by architecture.
However, for both investigated model families, performance stabilizes at around the maximal value above 20 samples. 
As stated in \cref{sec:methods:transformation}, 
we use 30 concept samples within \gls{ours} unless stated otherwise.
A promising future direction is to introduce weighted averaging for \gls{ours}, which could further improve its performance, particularly for architectures with less interpretable neurons.

\subsection{Attribution Graph}
\label{app:explore:describe:attribution_graph}
In \cref{sec:results:explore:describe},
we show an attribution graph (inspired by \cite{achtibat2023attribution}) of a ResNet50v2,
which illustrates where concepts are located in the architecture (which layer),
and how concepts are dependent on one another.
An attribution graph is computed through backpropagation of relevance scores via \gls{crp}.
The computational steps are outlined in the following:

\paragraph{1) Relevance filtering of components}
We compute the highest relevance of a component on the test set for a specific output target (here ``Ox'').
This relevance score is used to filter out irrelevant neurons.
Concretely,
we filter out neurons with relevance that is below 1\,\% or $\frac{5}{\text{\#neurons}}\%$.

\paragraph{2) Labelling of components}
To label neurons,
we first embed the following textual descriptions via the text model of CLIP:
\texttt{Horns}, \texttt{Eyes}, \texttt{Nose}, \texttt{Ears}, \texttt{Mouth}, \texttt{Neck}, \texttt{Back}, \texttt{Chest}, \texttt{Belly}, \texttt{Tail}, \texttt{Hooves}, \texttt{Legs}, \texttt{Brown}, \texttt{Black}, \texttt{Lined texture}, \texttt{Striped texture}, \texttt{Spotted texture}, \texttt{Brown colour},
\texttt{Short fur}, \texttt{Long fur}, \texttt{Thick skin}, \texttt{White}, \texttt{Spotted},
\texttt{Herding}, \texttt{Plowing}, \texttt{Pulling carts}, \texttt{Heavy labor}, \texttt{Grazing}, \texttt{Chewing cud},
\texttt{Grasslands}, \texttt{Savannahs}, \texttt{Farms}, \texttt{Pastures},
\texttt{Strength}, \texttt{Fertility}, \texttt{Festivals}, \texttt{Rituals}, \texttt{Hinduism}, \texttt{Mythology},
\texttt{Zebu}, \texttt{Hereford}, \texttt{Angus}, \texttt{Bison}, \texttt{Yak}, \texttt{Water Buffalo},
\texttt{Face}, \texttt{Grass}, \texttt{Arching}, \texttt{Angular}, \texttt{Branches}, \texttt{Mountains}, \texttt{Texture}, \texttt{Water}, \texttt{Sky}, \texttt{White color}, \texttt{Textures}.

We therefore use the templates of ``\texttt{<concept>}'', ``\texttt{<concept>}-like'', and ``an image of \texttt{<concept>}''.
In order to compute an alignment score $a$,
we subtract cosine similarity resulting from an empty template and threshold to $a \geq 0.01$.
All remaining components are grouped according to their most aligned label.

\paragraph{3) Relevance propagation}
We use \gls{crp} to attribute groups of components.
Concretely,
we start with \emph{layer 3} and attribute component groups using the \gls{lrp} $\varepsilon z^+ \flat$-rule \wrt to the ``Ox'' class on the test set.
Subsequently,
all lower-level (\emph{layer} 2) component groups are attributed \wrt an upper layer component group.

\begin{figure}[t]
    \centering
    \includegraphics[width=0.99\textwidth]{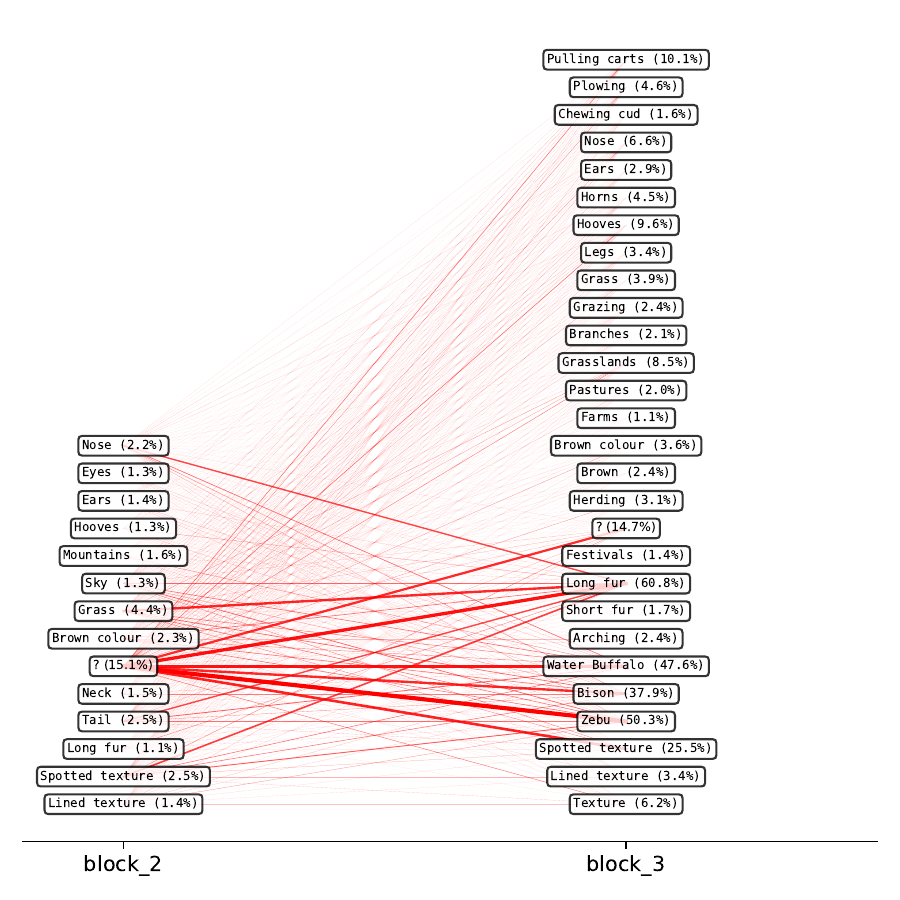}
    \caption{An attribution graph of a ResNet50v2 for the ``Ox'' class with concepts in \emph{layer 2} (``block\_2'') and \emph{layer 3} (``block\_3'').
    Unaligned components are marked with ``?'',
    indicating concepts that have not been though of, \ie, included in the list of expected concepts.
    Highest relevance scores on the test set for a group of components (with the same label) are shown in parentheses.
    The size of edges between concepts indicate their importance.
    }
    \label{fig:app:explore:describe:attribution_graph_resnet50v2_ox}
\end{figure}

\paragraph{4) Investigate the attribution graph}
The full attribution graph is shown in \cref{fig:app:explore:describe:attribution_graph_resnet50v2_ox}
for \emph{layer 2} (``block\_2'') and \emph{layer 3} (``block\_3'').
Notably,
unaligned components are marked with ``?'',
indicating concepts that have not been though of, \ie, included in the list of expected concepts.
As discussed in \cref{sec:results:explore:describe},
some concepts refer to copyright signs, or background vegetation.
Here, we additionally show the highest relevance scores on the test set for a group of components (with the same label) in parentheses.

\setcounter{figure}{0}
\setcounter{table}{0}\setcounter{equation}{0}

\section{Compare in Semantic Space}
\label{app:exp:compare}
This section provides more details on \cref{sec:results:explore:compare},
where we describe how \gls{ours} can be used to compare the knowledge of different models.
\begin{figure}[t]
    \centering
    \includegraphics[width=0.99\textwidth]{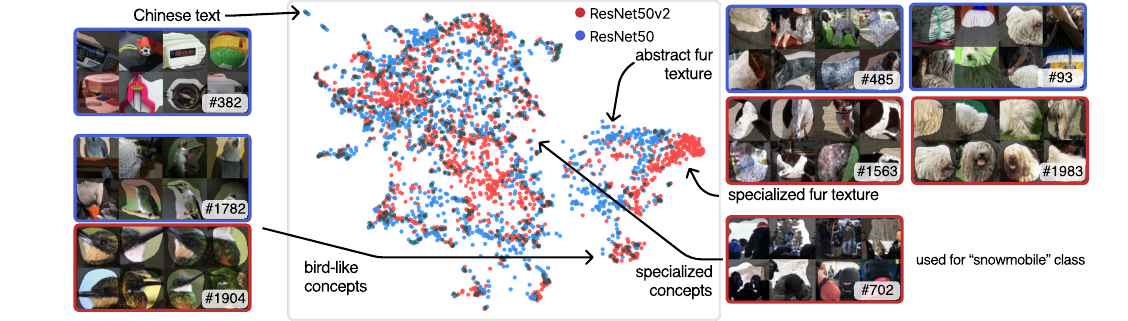}
    \caption{Comparing concept alignment between models (ResNet50 and ResNet50v2) based on their semantic embeddings.
    Both models share common knowledge, \eg, bird-related concepts.
    However, whereas the better trained ResNet50v2 has learned more specific concepts, \eg, specific fur textures of dogs, the other has learned more abstract concepts that are shared throughout classes.
    }
    \label{fig:app:explore:compare:compare_qualitatively}
\end{figure}
\gls{ours} allows to \emph{qualitatively} compare two models, as is shown in \cref{fig:app:explore:compare:compare_qualitatively} with two ResNet50 models trained on ImageNet, where one (ResNet50v2) is trained more extensively and results in higher test accuracy.
Both models share common knowledge, \eg, bird-related concepts.
However, whereas the better trained ResNet50v2 has learned more specific concepts, \eg, specific fur textures of dogs, the other has learned more abstract concepts that are shared throughout classes.
For the dog breed ``Komondor'' which has a white mop-like coat, for example,
the ResNet50 has learned a mop-like concept (neuron \texttt{\#93}) that is used to detect ``Komondor'' as well as ``mop'',
whereas the ResNet50v2 learned a class-specific concept.
Further,
the ResNet has learned a spotted burrito texture (neuron \texttt{\#485}) that is used to detect burritos or German Shorthaired Pointer dogs.
This is in line with works that study generalization of neural networks for long training regimes,
observing that latent model components become more structured and class-specific~\cite{liu2022towards}.
We further provide quantitative comparisons via network dissection in \cref{app:explore:describe:dissection}.

Concretely, 
two models $\mathcal{N}$ and $\mathcal{M}$ may be quantitatively compared via the number of neurons that were assigned to concept labels as introduced by NetDissect~\cite{bau2017network} and detailed in \cref{app:explore:describe:dissection},
or by measuring set similarity $S_{{\mathcal{V}_\mathcal{M}} \rightarrow \mathcal{V}_\mathcal{N}}$ based on average maximal pairwise similarity:
\begin{equation}\label{eq:app:explore:compare:concept_comparison}
    S_{{\mathcal{V}_\mathcal{M}} \rightarrow \mathcal{V}_\mathcal{N}} = \frac{1}{|\mathcal{V}_\mathcal{M}|}
    \sum_{\boldsymbol{\vartheta} \in\mathcal{V}_\mathcal{M}}\max_{\boldsymbol{\vartheta}' \in\mathcal{V}_\mathcal{N}}
    s(\boldsymbol{\vartheta},\boldsymbol{\vartheta}'),
\end{equation}
that quantifies the degree to which the knowledge (semantics) encoded in model $\mathcal{M}$ is also encoded in model $\mathcal{N}$.

\begin{figure}[t]
    \centering
    \includegraphics[width=0.99\textwidth]{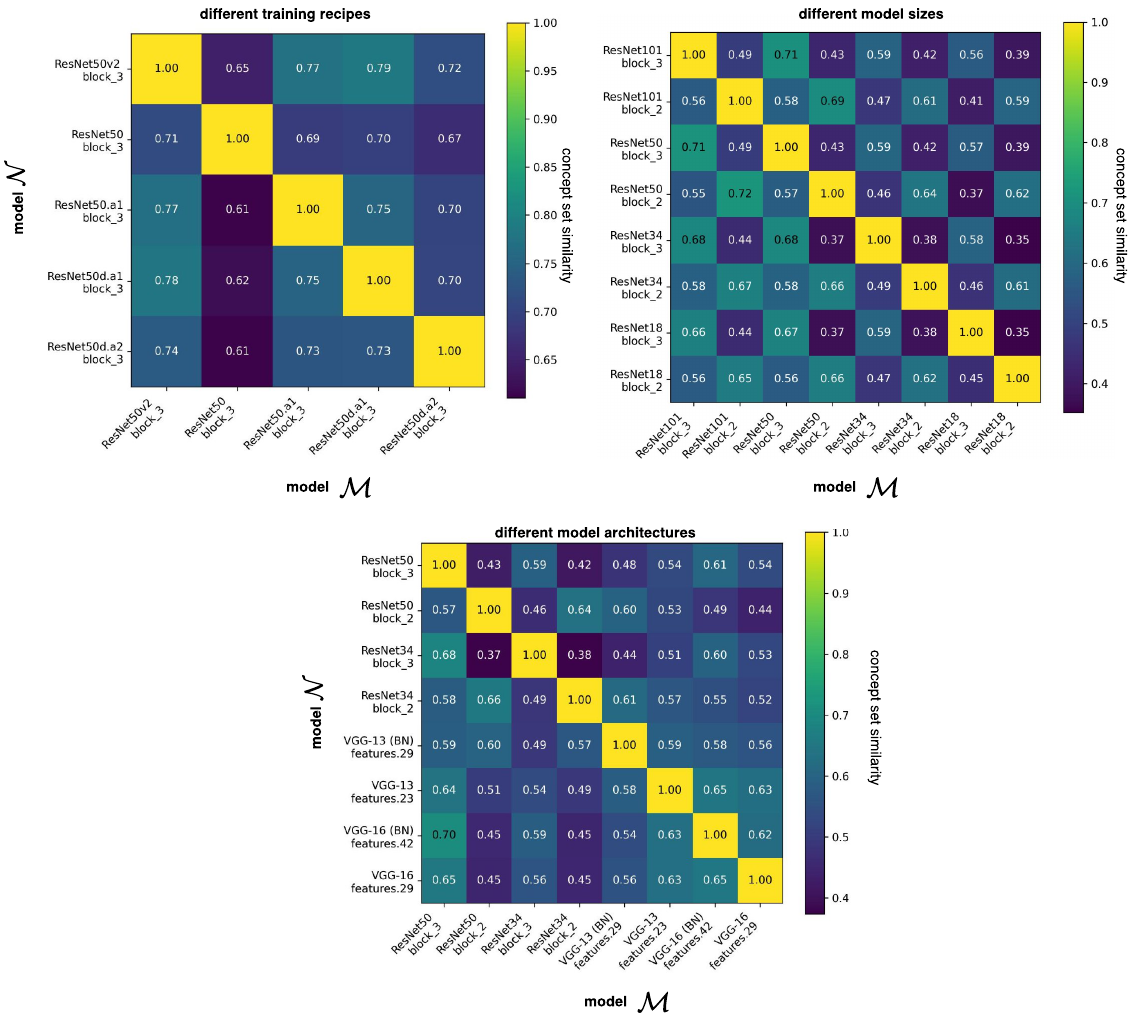}
    \caption{Comparing concept alignment between models based on their semantic embeddings.
    The set of semantic embeddings is hereby compared by computing the average maximal pairwise similarity of semantic embeddings between models as given by \cref{eq:app:explore:compare:concept_comparison}.
    \emph{Top left:}
    Comparing ResNet50 models (last feature layer) with different training recipes.
    \emph{Top right:}
    Comparing ResNet models (penultimate and last feature layer).
    \emph{Bottom middle:}
    Comparing ResNet and VGG models (last feature layer).
    }
    \label{fig:app:explore:compare:quantitatively}
\end{figure}

\cref{eq:app:explore:compare:concept_comparison} allows us, \eg, to compare the same architecture with different training recipes, different model sizes and layers, or different model architectures as shown in \cref{fig:app:explore:compare:quantitatively},
where the DINOv2 is used as the foundation model to generate semantic embeddings (using the top 30 most activating image patches).

\textbf{Training recipes matter}:
When comparing ResNet50 models trained with different training recipes in 
\cref{fig:app:explore:compare:quantitatively} (\emph{top left}),
it is apparent that differences in alignment exist.
Interestingly,
the ResNet50v2 aligns overall the best,
whereas the ResNet50 aligns generally low with the other models.
Notably,
there exists a correlation with test accuracy/training efficiency.
The ResNet50v2 and a1-version are trained for 600 epochs,
whereas the a2 version and ResNet50 are only trained for 300 and 90 epochs, respectively.

\textbf{Similar model sizes are more aligned}:
In \cref{fig:app:explore:compare:quantitatively} (\emph{top right}),
we compare ResNet models of different size (ResNet18/34/50/101).
It can be observed,
that ResNets of similar size also align more (\eg, ResNet101 and ResNet50 compared to ResNet101 and ResNet 34).

\textbf{Similar layer depths are more aligned}:
Further,
for all ResNet models,
representations of layer ``block\_3'' and ``block\_2'' are more aligned with one-another.

\textbf{Intra family alignment can be higher than inter family alignment}:
In \cref{fig:app:explore:compare:quantitatively} (\emph{top right}),
we compare ResNets and VGG models \wrt their semantic embedding alignment.
Interestingly,
the ResNet50's and VGG-16's (with batch normalization) last feature layers are more aligned than the ResNet50's and ResNet34's,
 or the VGG-16's and VGG-13's.

\setcounter{figure}{0}
\setcounter{table}{0}\setcounter{equation}{0}
\section{Audit in Semantic Space}
\label{app:exp:audit}
This section provides more details on \cref{sec:results:audit,sec:results:medical},
where we demonstrate how to audit neural network representations using \gls{ours} on ImageNet and ISIC, respectively.

\subsection{ImageNet Visual Object Recognition}
This section provides more details on \cref{sec:results:audit},
where we investigate neural network representations using \gls{ours} on ImageNet for a concrete examples (Ox predictions), and, more generally, for 26 classes.
We present further details in the following.

\subsubsection{Ox Example}
In \cref{sec:results:audit} we demonstrate an alignment audit for a ResNet50v2 and predictions for the class ``Ox''.
In order to receive alignment scores,
we measure the alignment with textual embeddings of valid and spurious concepts using the templates of ``\texttt{<concept>}'' and ``an image of \texttt{<concept>}''. 
Specifically, the following concepts are used to measure alignment:

\textbf{Valid:}
{
    \texttt{large muscular body}, 
    \texttt{curved horns}, 
    \texttt{hooves}, 
    \texttt{thick neck}, 
    \texttt{short, rough fur}, 
    \texttt{soft fur}, 
    \texttt{long fur}, 
    \texttt{gray skin}, 
    \texttt{fur texture}, 
    \texttt{gritted texture}, 
    \texttt{brown coat}, 
    \texttt{black coat}, 
    \texttt{white coat}, 
    \texttt{legs}, 
    \texttt{long tail}, 
    \texttt{wide muzzle}
}

\textbf{Spurious:}{
    \texttt{grassland, savanna}, 
    \texttt{sky}, 
    \texttt{tree}, 
    \texttt{water}, 
    \texttt{grain, straw}, 
    \texttt{cart, carriage}, 
    \texttt{carriage}, 
    \texttt{wheel}, 
    \texttt{indian person}, 
    \texttt{copyright watermark}, 
    \texttt{mud, dirt}, 
    \texttt{person}, 
    \texttt{wooden}, 
    \texttt{palm tree}, 
    \texttt{people}, 
    \texttt{grayish earth texture}
}

To compute final alignment scores,
we subtract the alignment to an empty textual embedding, as also detailed in \cref{sec:methods}.
We further filter out all neurons of the last feature layer (\emph{layer 3} for the ResNet) that have a maximal relevance of below 2.8\,\% on the test set for the ``Ox'' class using CRP for attribution.
\begin{figure}[t]
    \centering
    \includegraphics[width=0.99\textwidth]{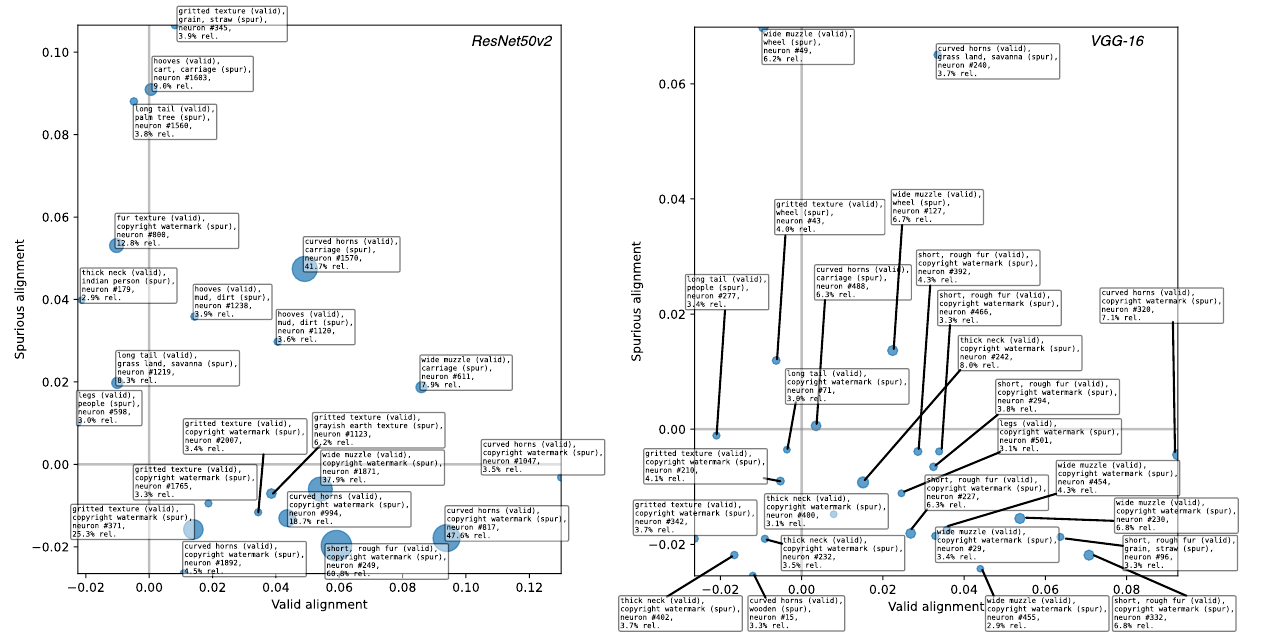}
    \caption{Valid and spurious alignment scores of all relevant neurons for detecting the ``Ox'' class on the ImageNet test set for a ResNet50v2 (\emph{left}) and VGG-16 model (\emph{right}). We include labels for all representations (highest spurious and valid aligned concept), with the highest relevance score on the test set as well as their number/id.
    }
    \label{fig:app:explore:audit:audit}
\end{figure}
We show the valid and spurious alignment of all relevant components/neurons and their concepts in form of a scatter plot in \cref{fig:results:audit}a,
and in more detail in \cref{fig:app:explore:audit:audit}.
In \cref{fig:app:explore:audit:audit}, 
the same analysis with a VGG-16 is also included,
with labels for all representations, the highest relevance score on the test set as well as neuron number/id.

It is apparent,
that the ResNet often relies on fewer neurons as neurons have up to 60.8\,\% of relevance compared to 8.0\,\% for the VGG.
Whereas both models seem to rely on \texttt{cart} and \texttt{savanna}-related concepts,
the ResNet further utilizes concepts corresponding to \texttt{copyright watermark}, \texttt{Indian person} and \texttt{palm tree}.

\begin{figure}[t]
    \centering
    \includegraphics[width=0.99\textwidth]{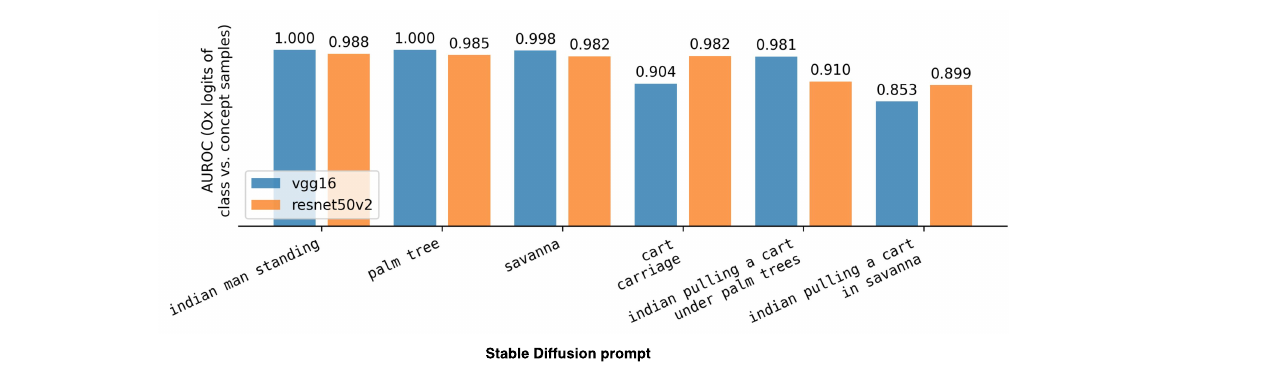}
    \caption{AUROC when separating ``Ox'' logits computed on the ``Ox'' test set and Stable Diffusion-generated samples of spurious concepts for a ResNet50v2 and VGG-16.
    Especially when spurious concepts are combined is the AUC reduced.
    }
    \label{fig:app:explore:audit:audit_test}
\end{figure}
In a subsequent experiment,
the models are evaluated on actual data.
Concretely,
we measure how well the ``Ox''-logits can be separated on samples with ``Ox'' (test set) and with spurious features only via an AUC score~\cite{neuhaus2023spurious}.
The spurious feature set is hereby generated using Stable Diffusion~\cite{rombach2022high} and text prompts.
The resulting AUC scores (low AUC is problematic) are displayed in \cref{fig:app:explore:audit:audit_test} for both models and different diffusion prompts.
As expected,
the ResNet shows slightly lower AUCs for \texttt{Indian person} and \texttt{palm tree},
whereas both show AUCs below 1.0 for \texttt{savanna} and \texttt{cart carriage}.
Especially for the combination of spurious features (\texttt{Indian pulling a cart in savanna}),
both models show AUCs below 0.9.
Interestingly,
when \texttt{palm trees} are shown instead of \texttt{savanna} (\texttt{Indian pulling a cart under palm trees}),
the ResNet reacts much more than the VGG, validating the finding in \cref{fig:app:explore:audit:audit},
that only the ResNet relies on palm tree features.

\subsubsection{Spurious Concept Reliance Everywhere}
This section provides details for
\cref{sec:results:audit},
where we measure the alignment of ResNet50v2 neurons in the last feature layer with valid expected concepts.
In
\cref{fig:app:explore:audit:audit_background_0,fig:app:explore:audit:audit_background_1,fig:app:explore:audit:audit_background_2,fig:app:explore:audit:audit_background_3,fig:app:explore:audit:audit_background_4,fig:app:explore:audit:audit_background_5,fig:app:explore:audit:audit_background_6,fig:app:explore:audit:audit_background_7,fig:app:explore:audit:audit_background_8,},
we provide for all 26 ImageNet classes the top-10 most relevant neurons (according to the highest relevance score on the test set of a class),
the best matching label and corresponding alignment score.
\begin{figure}[t]
    \centering
    \includegraphics[width=0.32\textwidth]{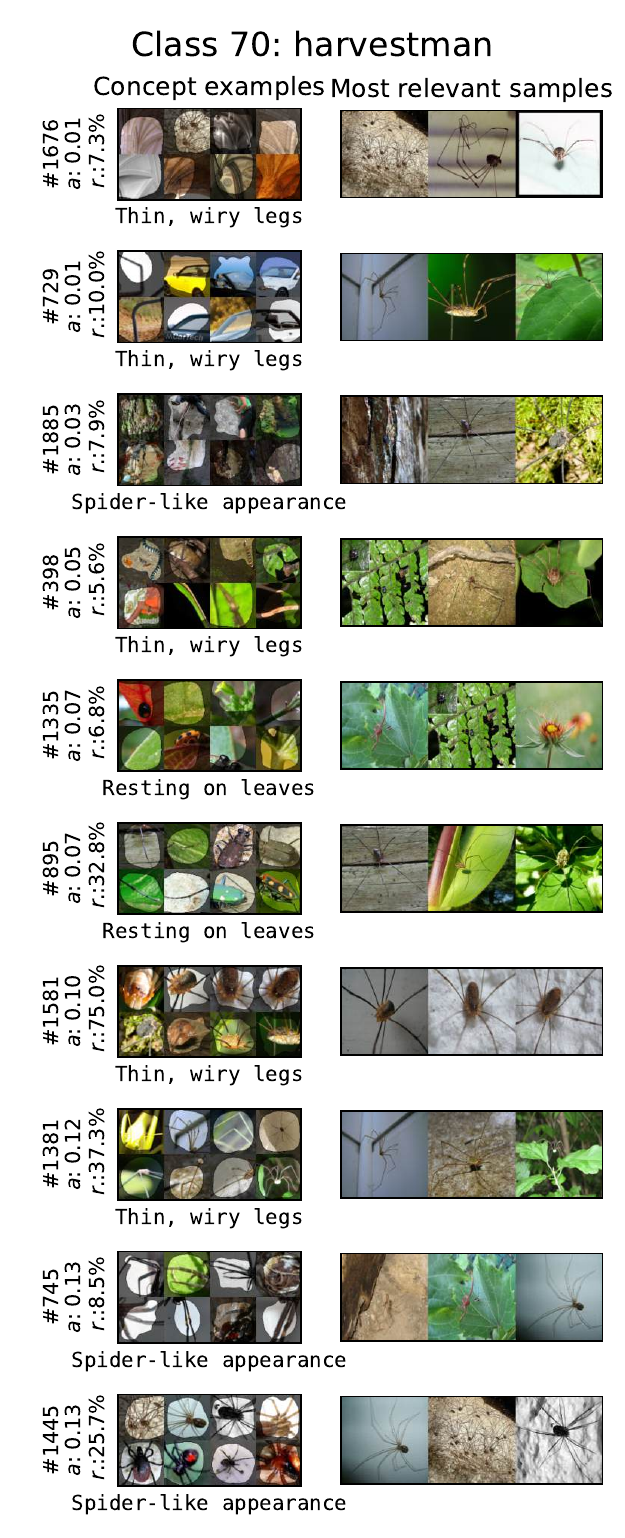}
        \includegraphics[width=0.32\textwidth]{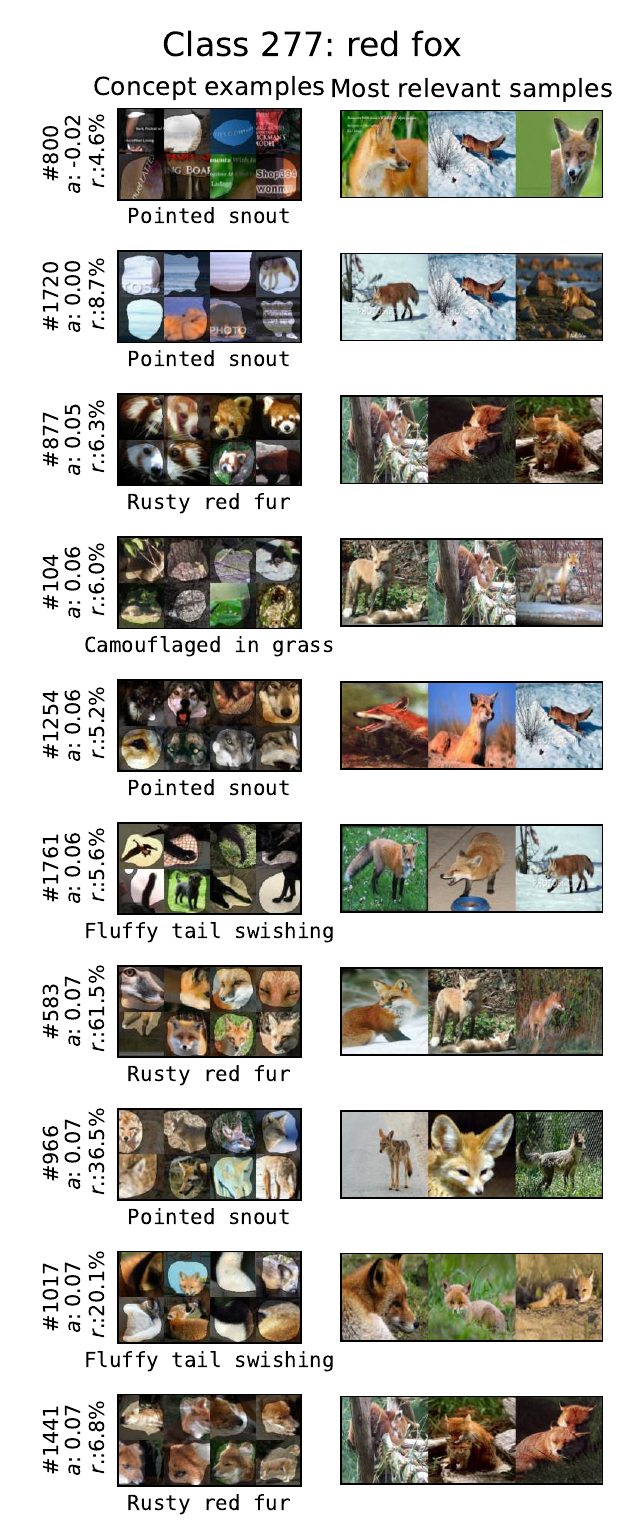}
            \includegraphics[width=0.32\textwidth]{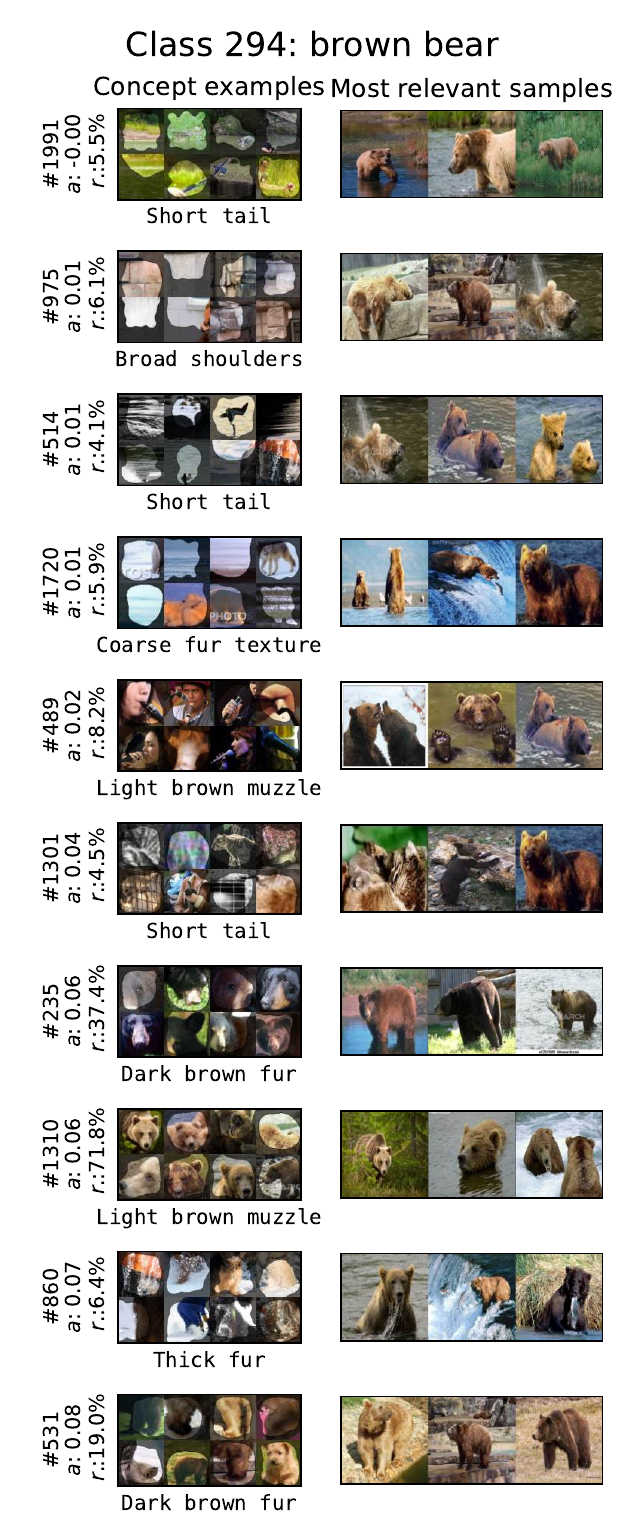}
    \caption{    
    Alignment of top-10 most relevant neurons of a ResNet50v2's last feature layer for the classes of ``harvestman'', ``red fox'' and ``brown bear''. Besides highest alignment $a$ to an expected concept (label below concept examples), we provide the highest relevance scores $r$ on the test set of the class. Further depicted are three examples where a neuron is most relevant for a class.
    }
    \label{fig:app:explore:audit:audit_background_0}
\end{figure}
\begin{figure}[t]
    \centering
    \includegraphics[width=0.32\textwidth]{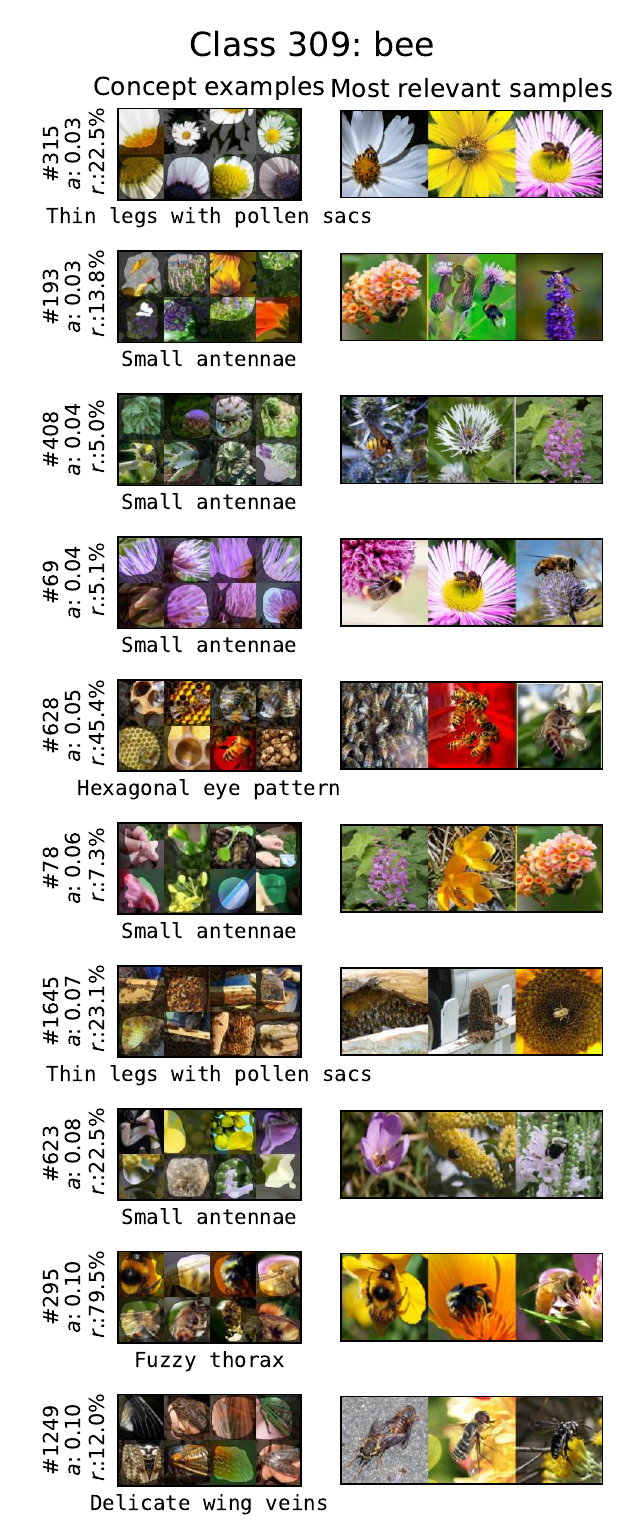}
        \includegraphics[width=0.32\textwidth]{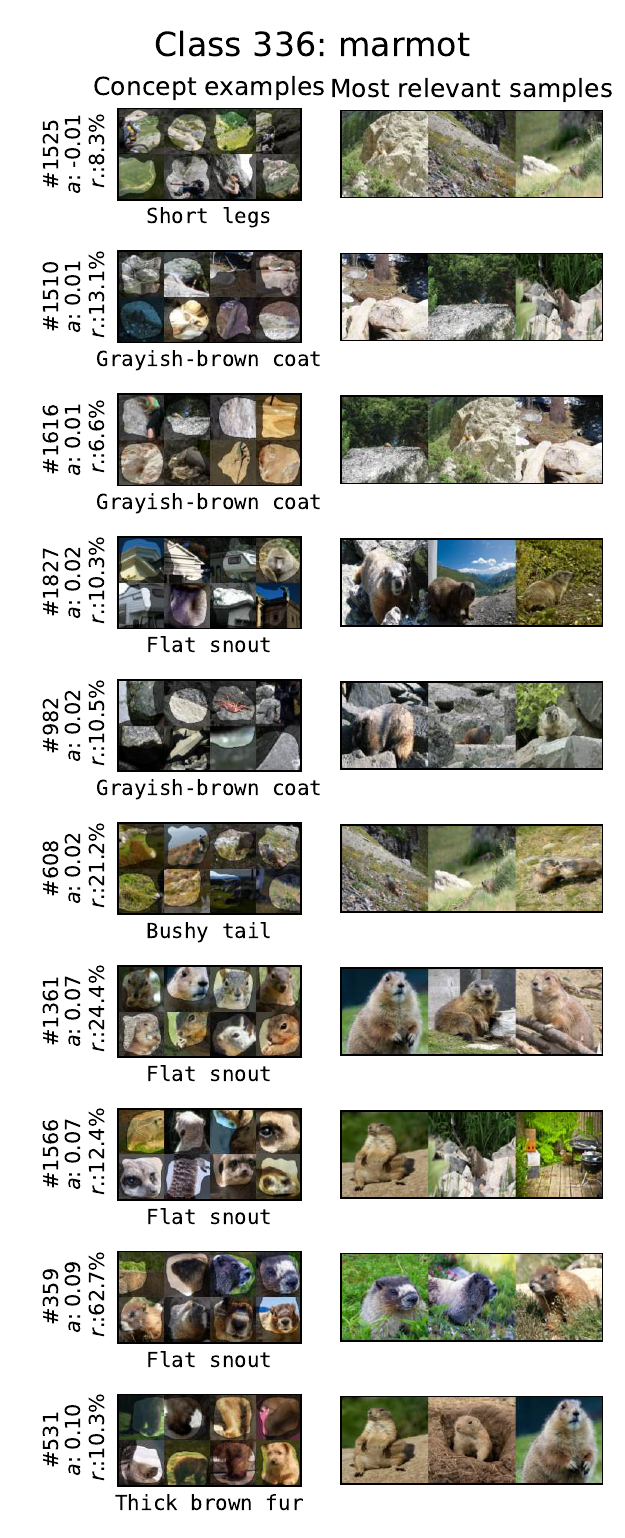}
            \includegraphics[width=0.32\textwidth]{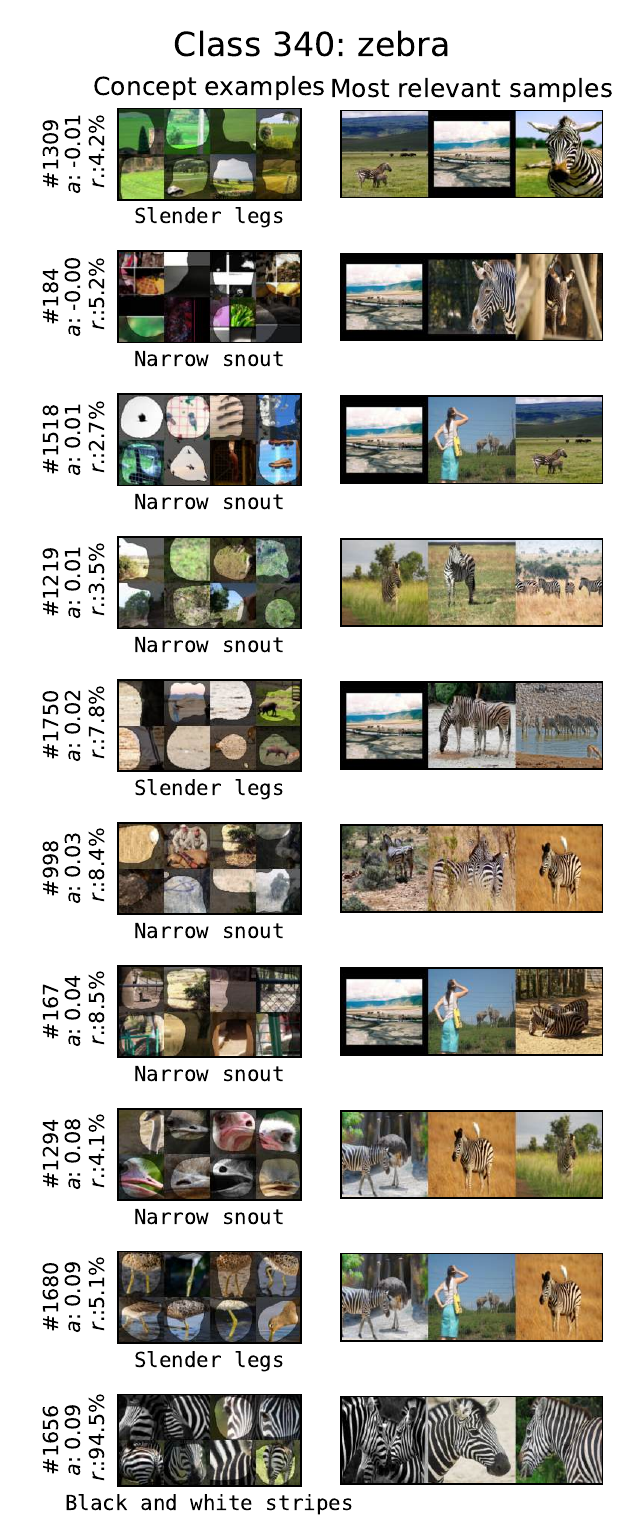}
    \caption{
    Alignment of top-10 most relevant neurons of a ResNet50v2's last feature layer for the classes of ``bee'', ``marmot'' and ``zebra''. Besides highest alignment $a$ to an expected concept (label below concept examples), we provide the highest relevance scores $r$ on the test set of the class. Further depicted are three examples where a neuron is most relevant for a class.
    }
    \label{fig:app:explore:audit:audit_background_1}
\end{figure}
\begin{figure}[t]
    \centering
    \includegraphics[width=0.32\textwidth]{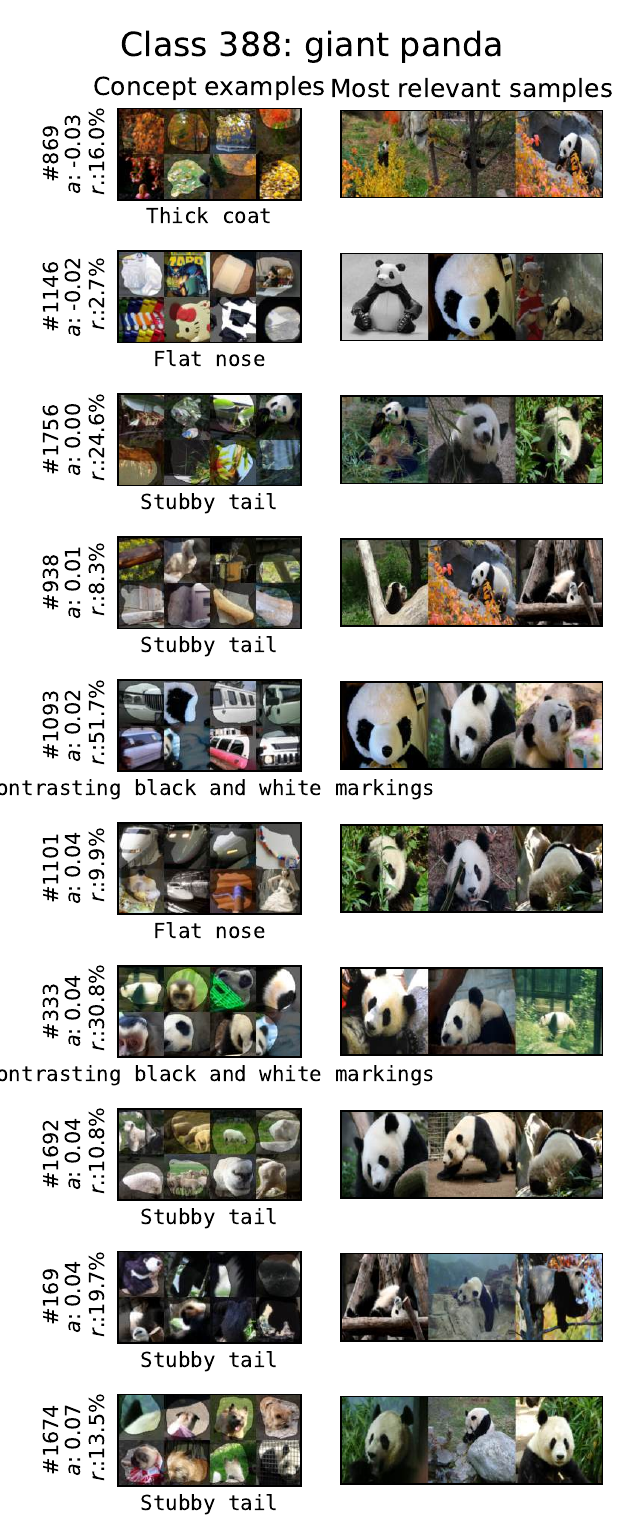}
        \includegraphics[width=0.32\textwidth]{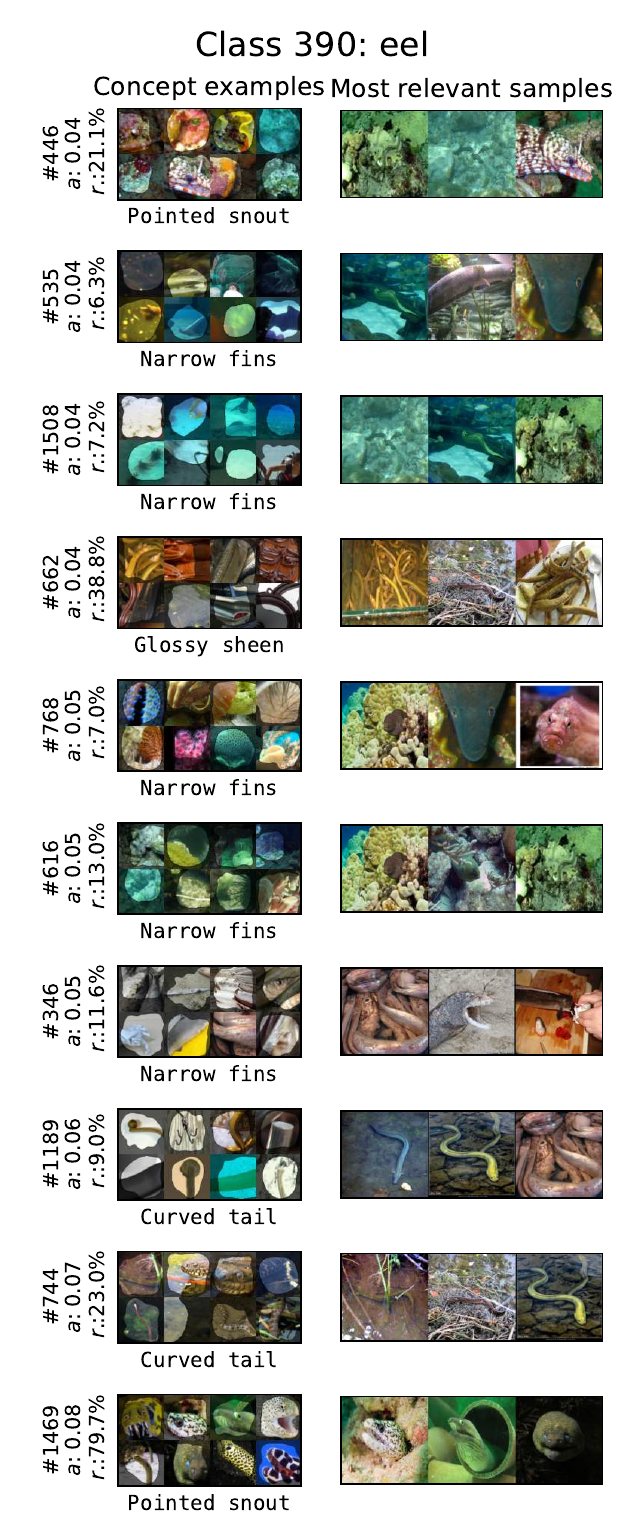}
            \includegraphics[width=0.32\textwidth]{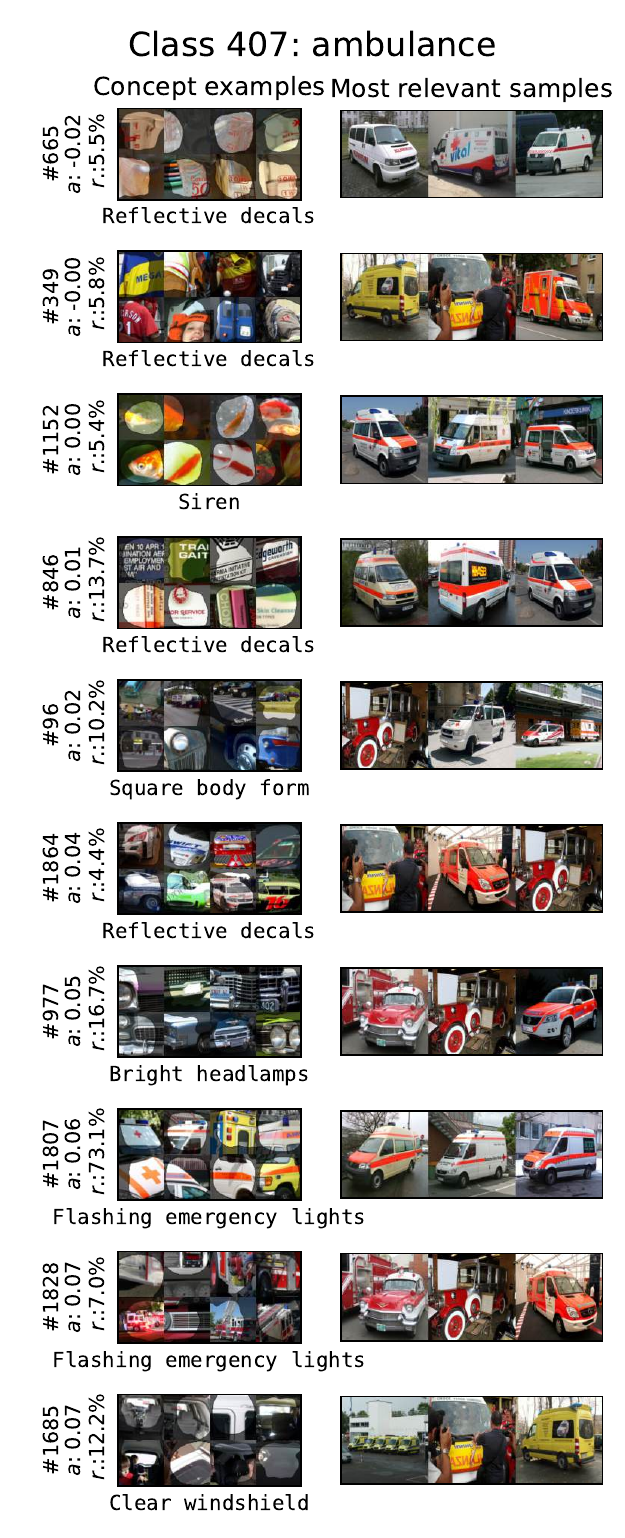}
    \caption{
    Alignment of top-10 most relevant neurons of a ResNet50v2's last feature layer for the classes of ``giant panda'', ``eel'' and ``ambulance''. Besides highest alignment $a$ to an expected concept (label below concept examples), we provide the highest relevance scores $r$ on the test set of the class. Further depicted are three examples where a neuron is most relevant for a class.
    }
    \label{fig:app:explore:audit:audit_background_2}
\end{figure}
\begin{figure}[t]
    \centering
    \includegraphics[width=0.32\textwidth]{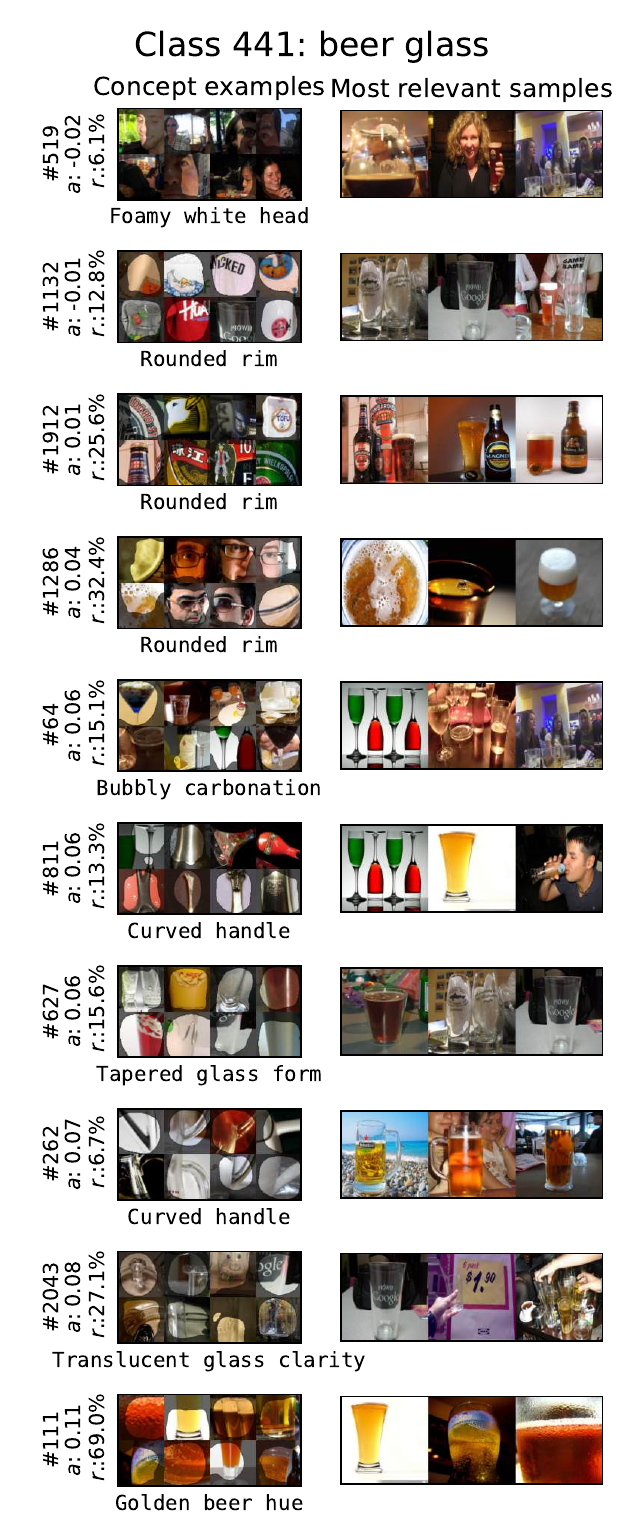}
        \includegraphics[width=0.32\textwidth]{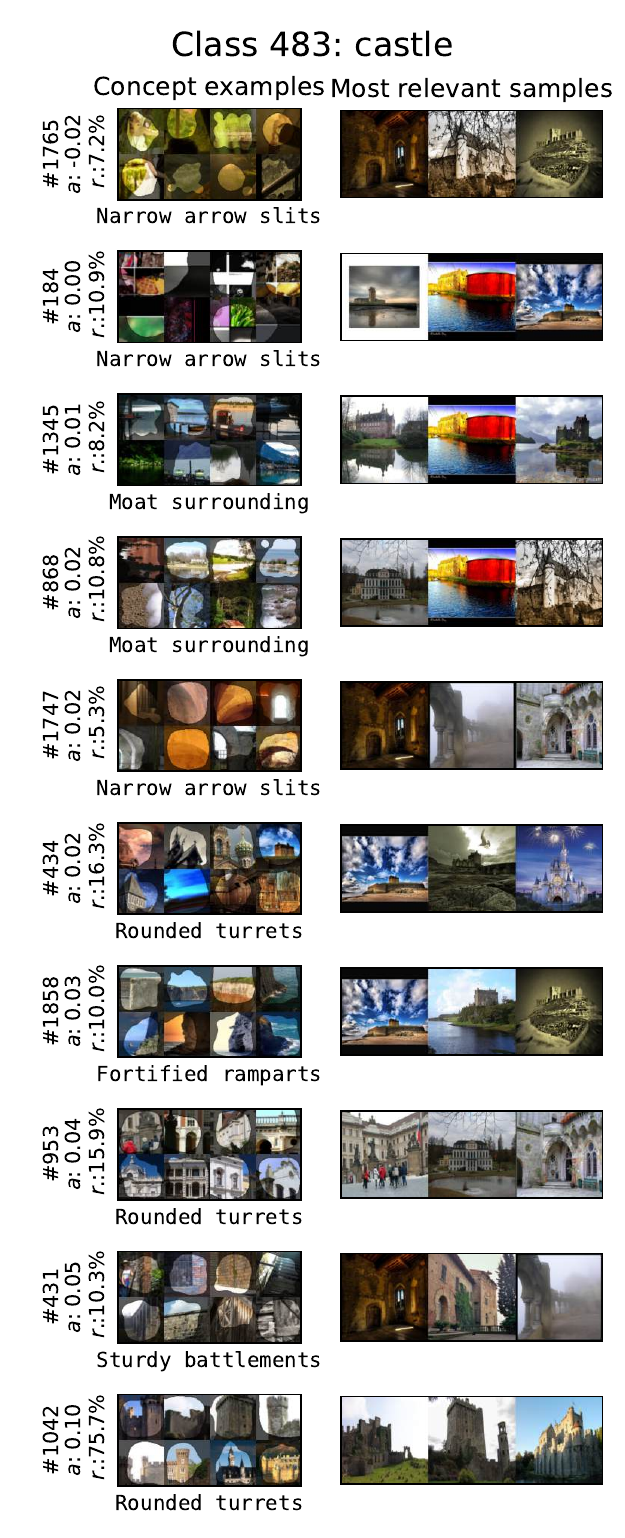}
            \includegraphics[width=0.32\textwidth]{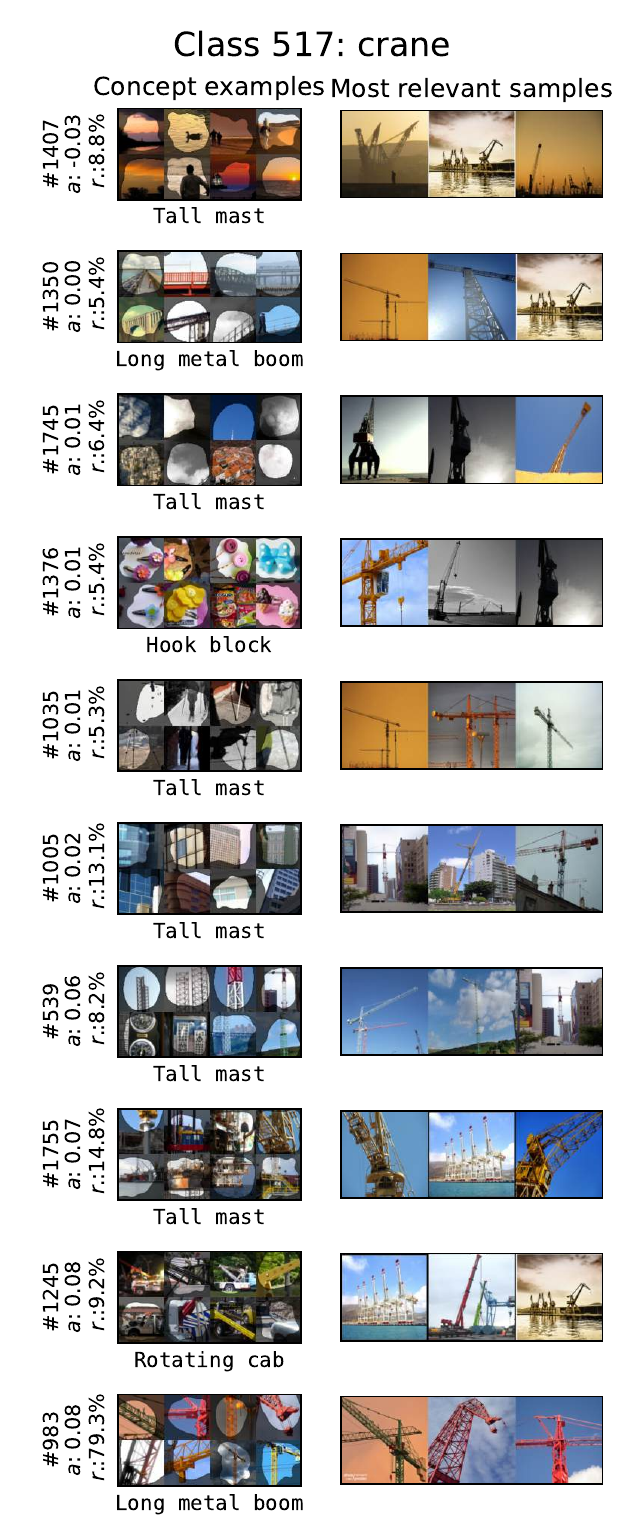}
    \caption{
    Alignment of top-10 most relevant neurons of a ResNet50v2's last feature layer for the classes of ``beer glass'', ``castle'' and ``crane''. Besides highest alignment $a$ to an expected concept (label below concept examples), we provide the highest relevance scores $r$ on the test set of the class. Further depicted are three examples where a neuron is most relevant for a class.
    }
    \label{fig:app:explore:audit:audit_background_3}
\end{figure}
\begin{figure}[t]
    \centering
    \includegraphics[width=0.32\textwidth]{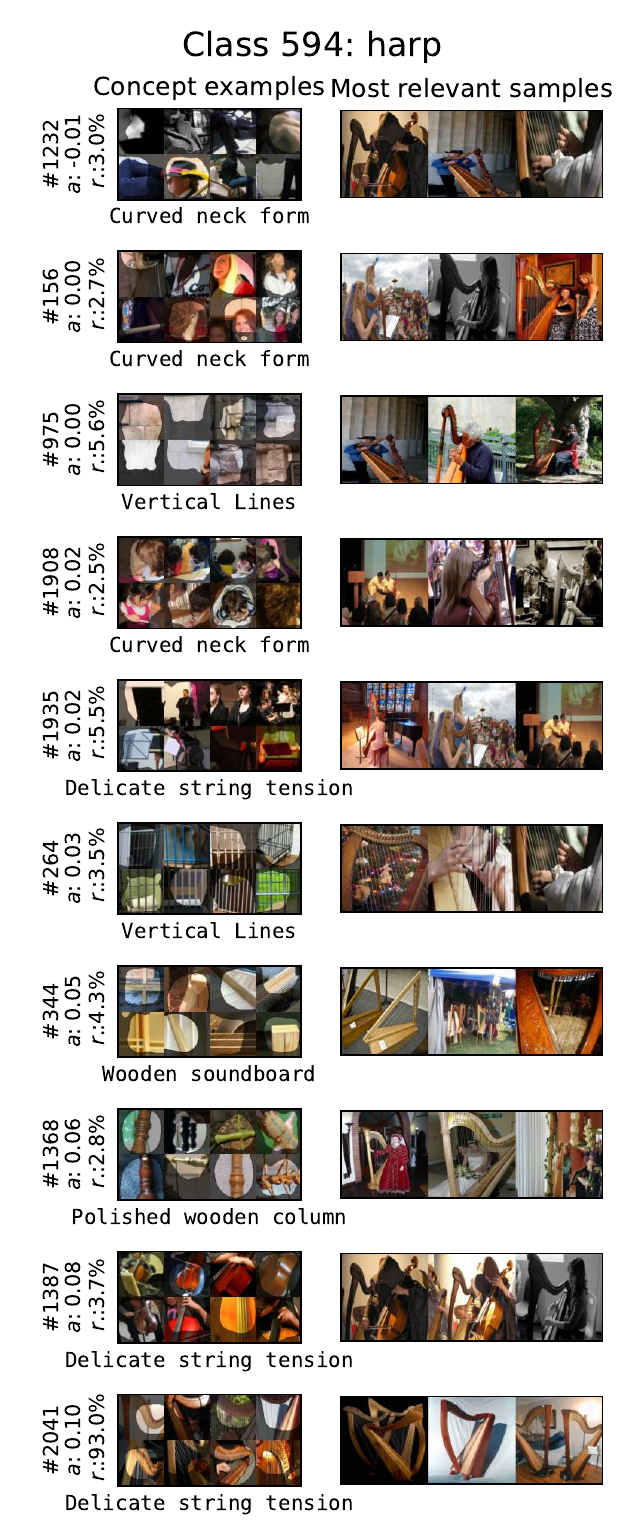}
        \includegraphics[width=0.32\textwidth]{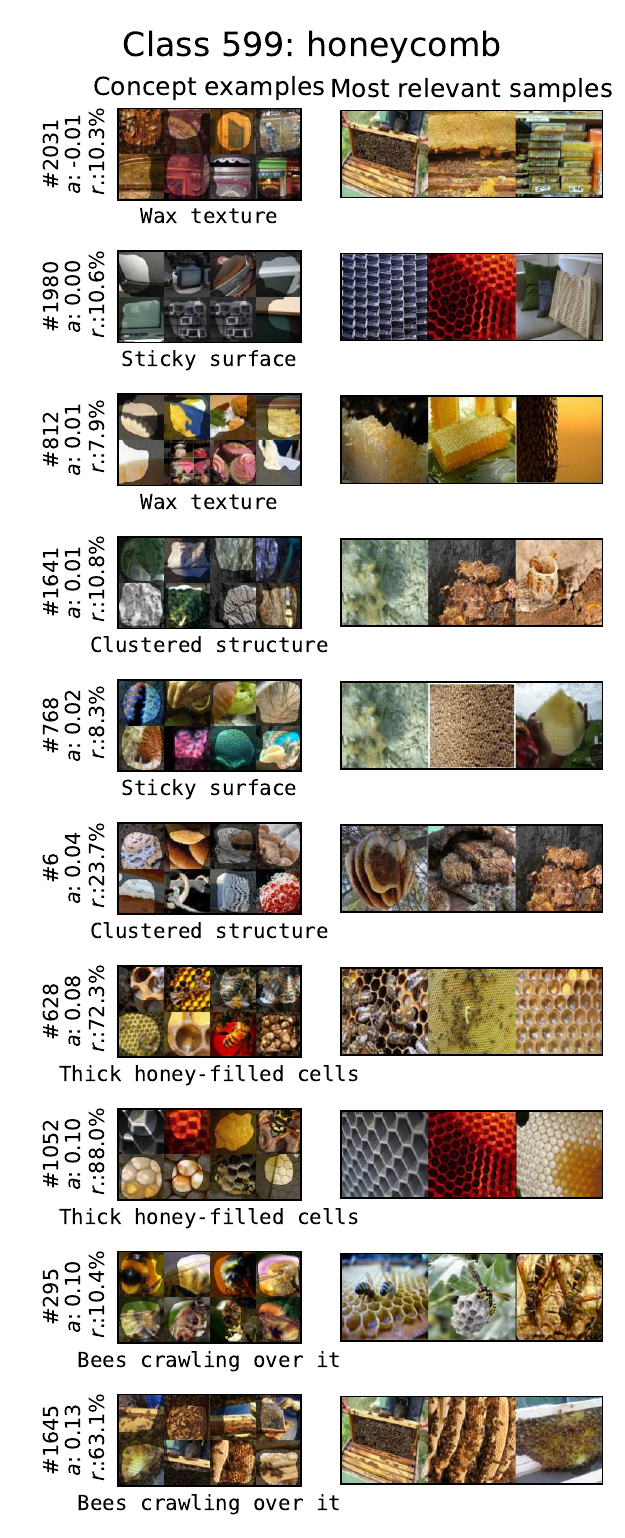}
            \includegraphics[width=0.32\textwidth]{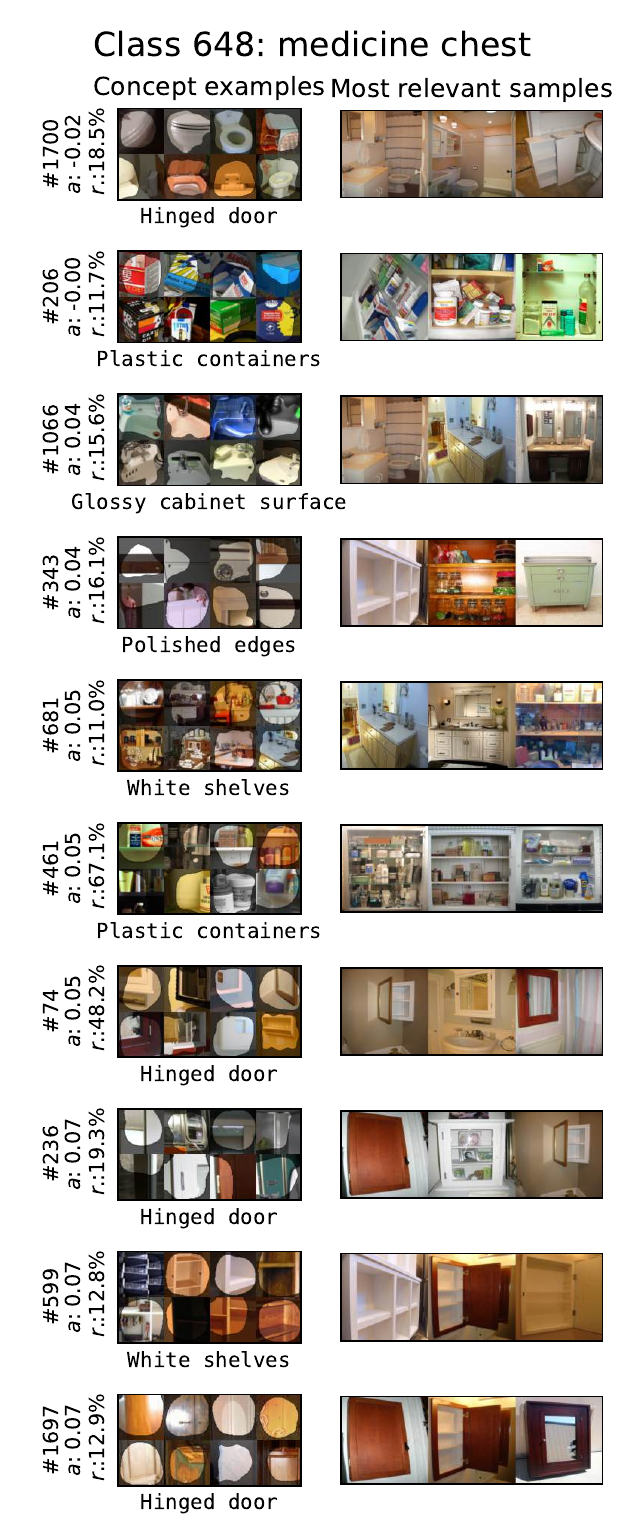}
    \caption{
    Alignment of top-10 most relevant neurons of a ResNet50v2's last feature layer for the classes of ``harp'', ``honeycomb'' and ``medicine chest''. Besides highest alignment $a$ to an expected concept (label below concept examples), we provide the highest relevance scores $r$ on the test set of the class. Further depicted are three examples where a neuron is most relevant for a class.
    }
    \label{fig:app:explore:audit:audit_background_4}
\end{figure}
\begin{figure}[t]
    \centering
    \includegraphics[width=0.32\textwidth]{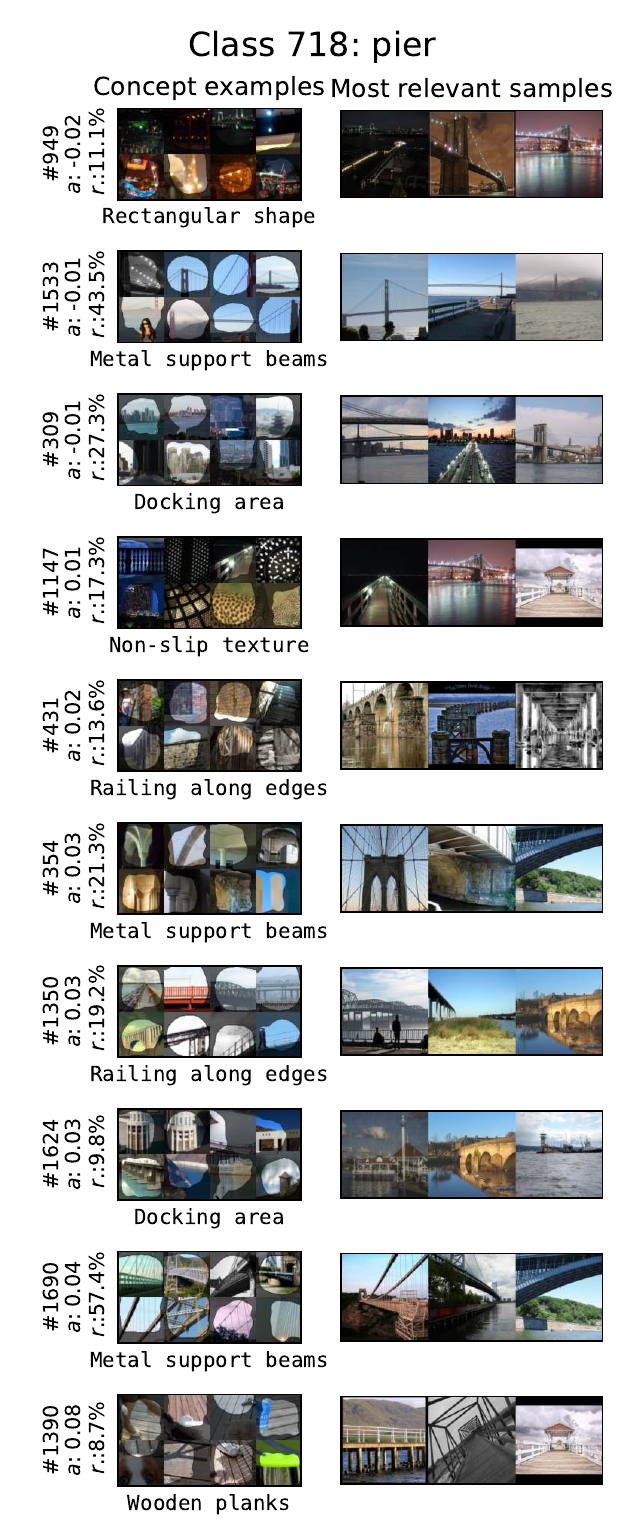}
        \includegraphics[width=0.32\textwidth]{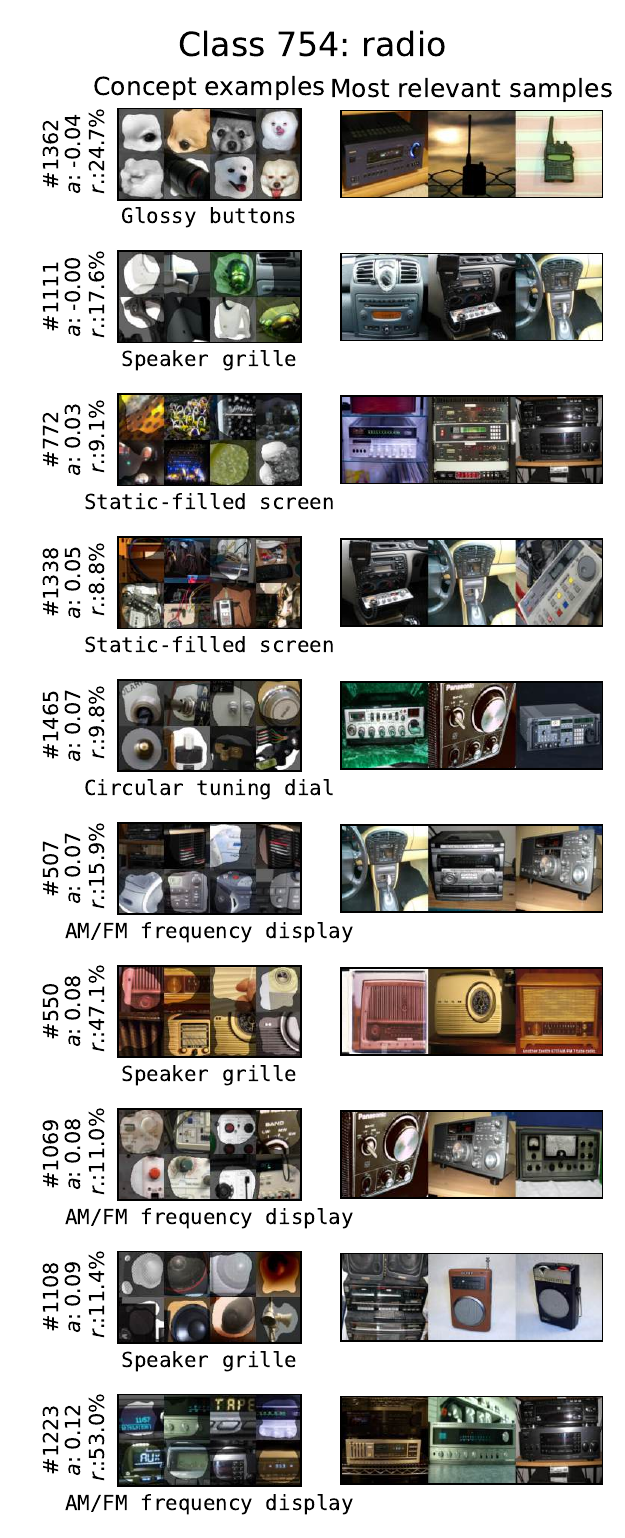}
            \includegraphics[width=0.32\textwidth]{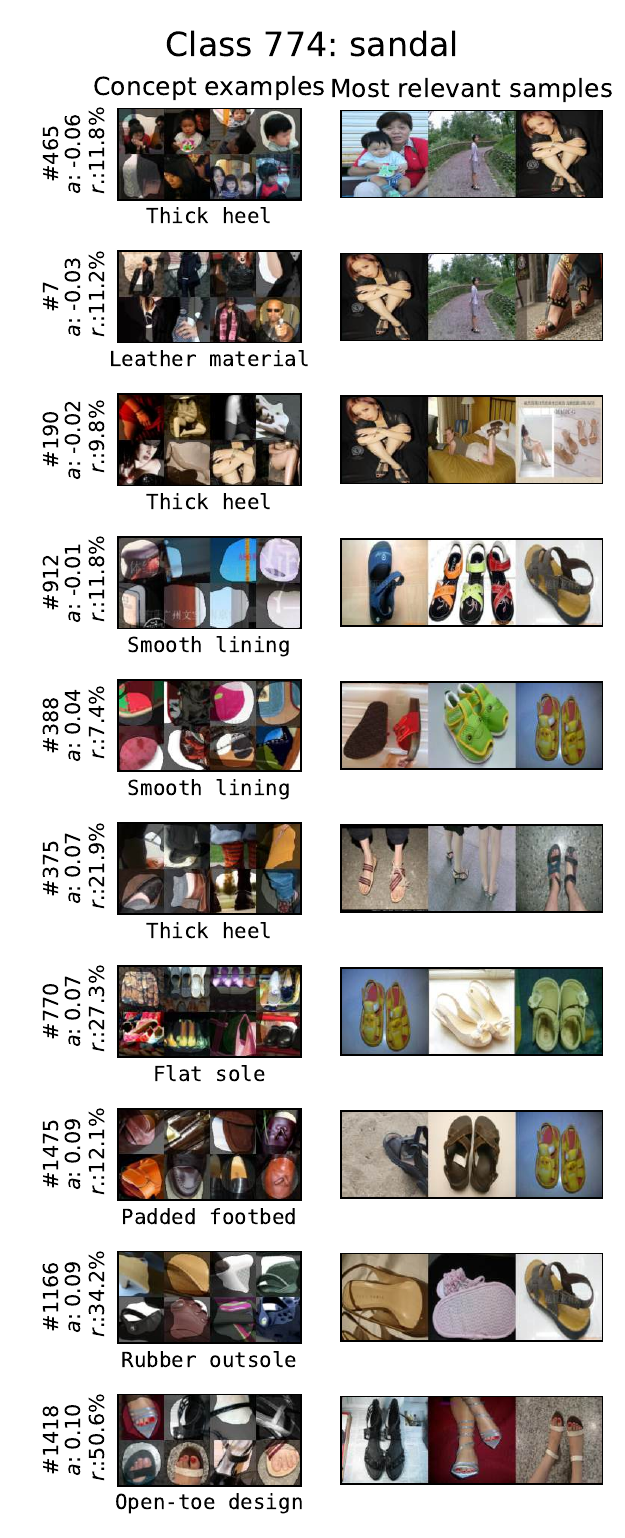}
    \caption{
    Alignment of top-10 most relevant neurons of a ResNet50v2's last feature layer for the classes of ``pier'', ``radio'' and ``sandal''. Besides highest alignment $a$ to an expected concept (label below concept examples), we provide the highest relevance scores $r$ on the test set of the class. Further depicted are three examples where a neuron is most relevant for a class.
    }
    \label{fig:app:explore:audit:audit_background_5}
\end{figure}
\begin{figure}[t]
    \centering
    \includegraphics[width=0.32\textwidth]{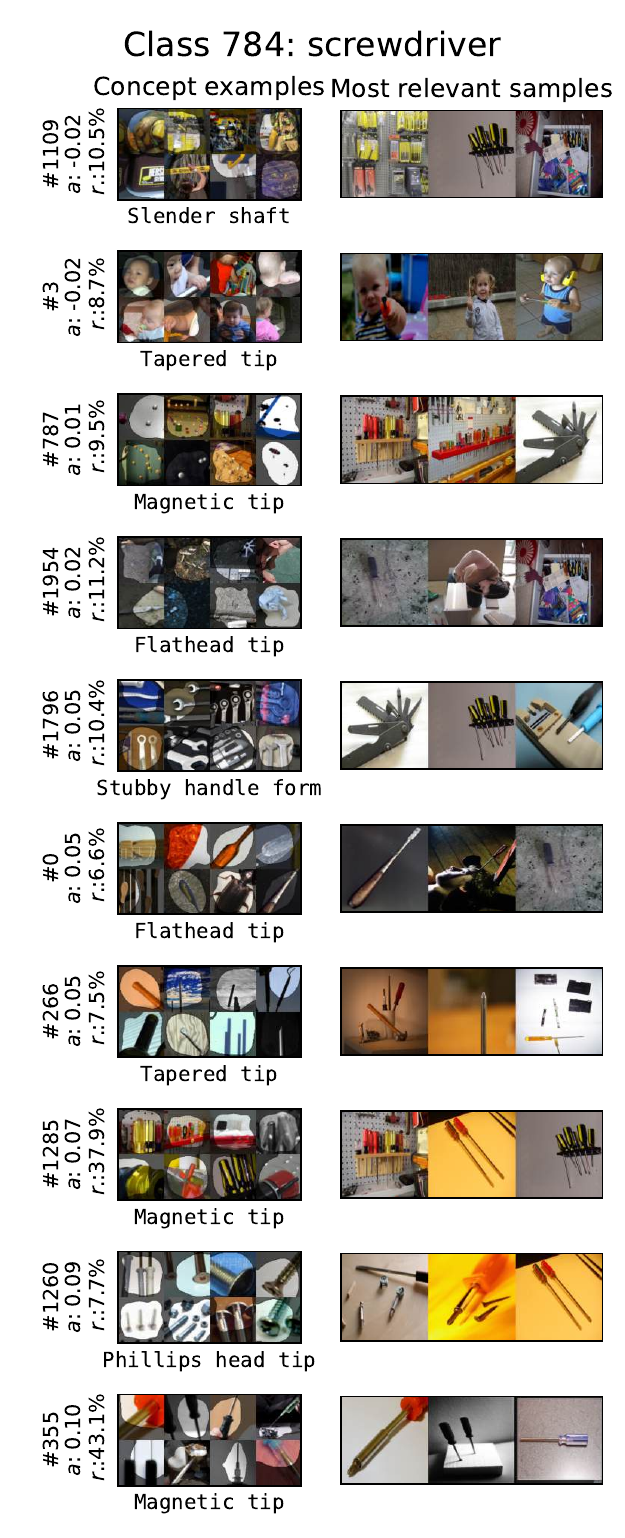}
        \includegraphics[width=0.32\textwidth]{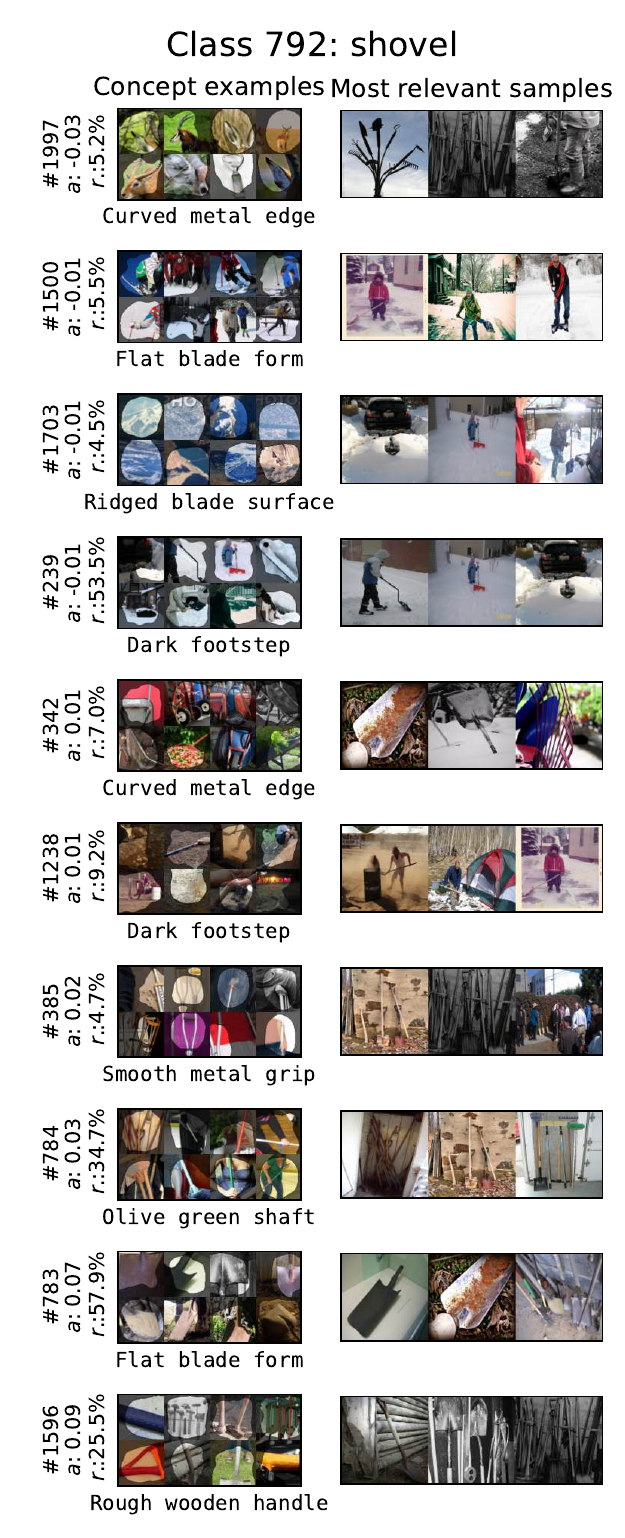}
            \includegraphics[width=0.32\textwidth]{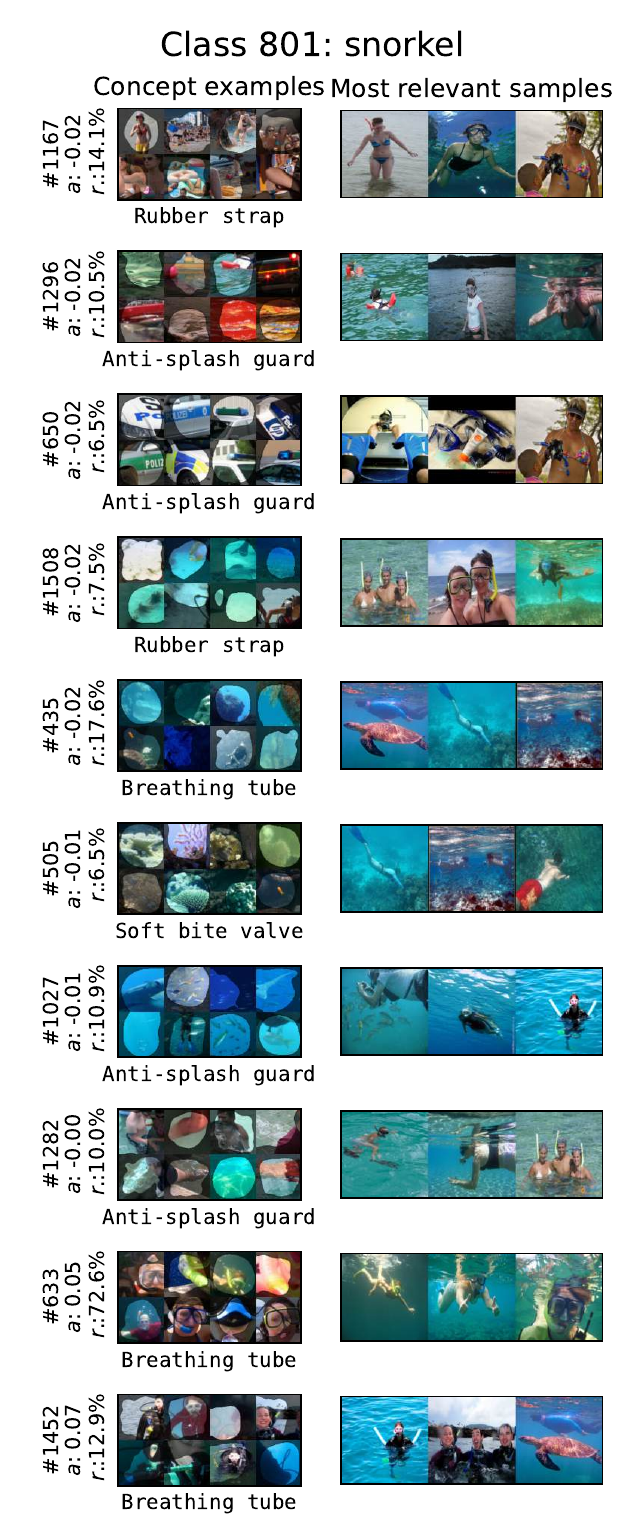}
    \caption{
    Alignment of top-10 most relevant neurons of a ResNet50v2's last feature layer for the classes of ``screwdriver'', ``shovel'' and ``snorkel''. Besides highest alignment $a$ to an expected concept (label below concept examples), we provide the highest relevance scores $r$ on the test set of the class. Further depicted are three examples where a neuron is most relevant for a class.
    }
    \label{fig:app:explore:audit:audit_background_6}
\end{figure}
\begin{figure}[t]
    \centering
    \includegraphics[width=0.32\textwidth]{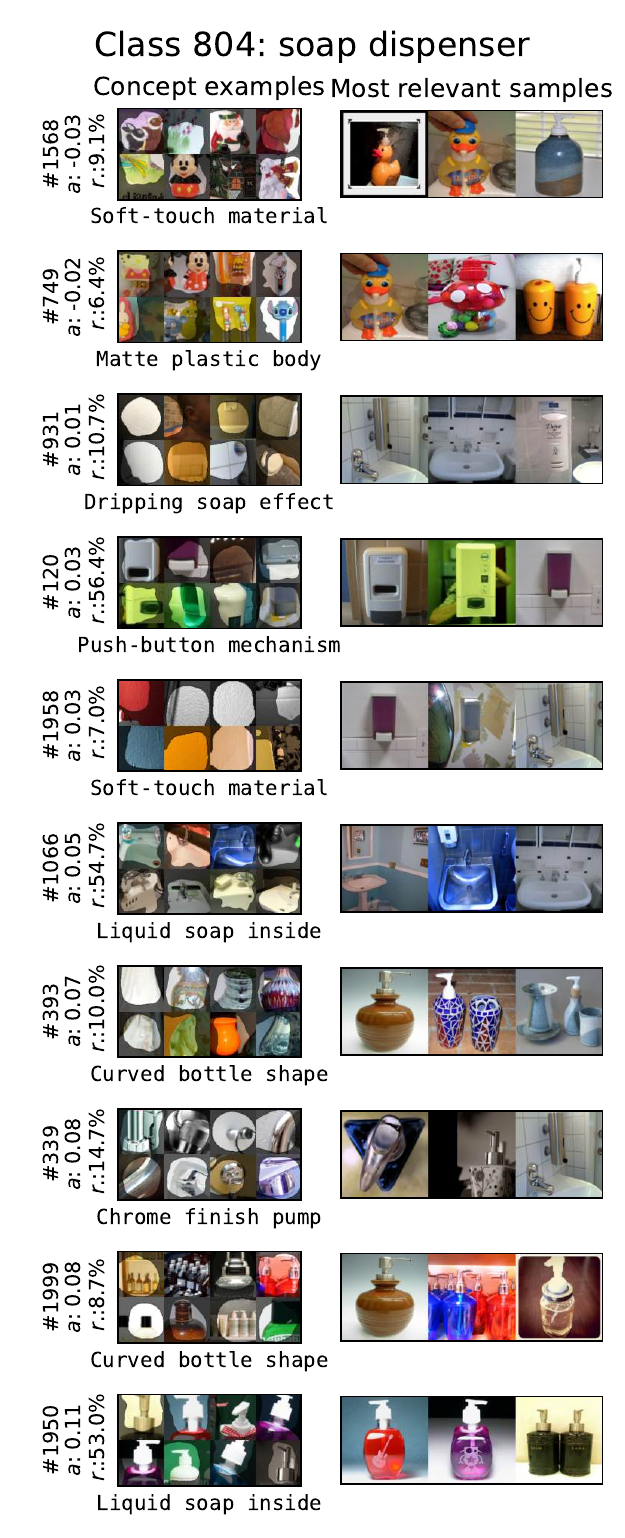}
        \includegraphics[width=0.32\textwidth]{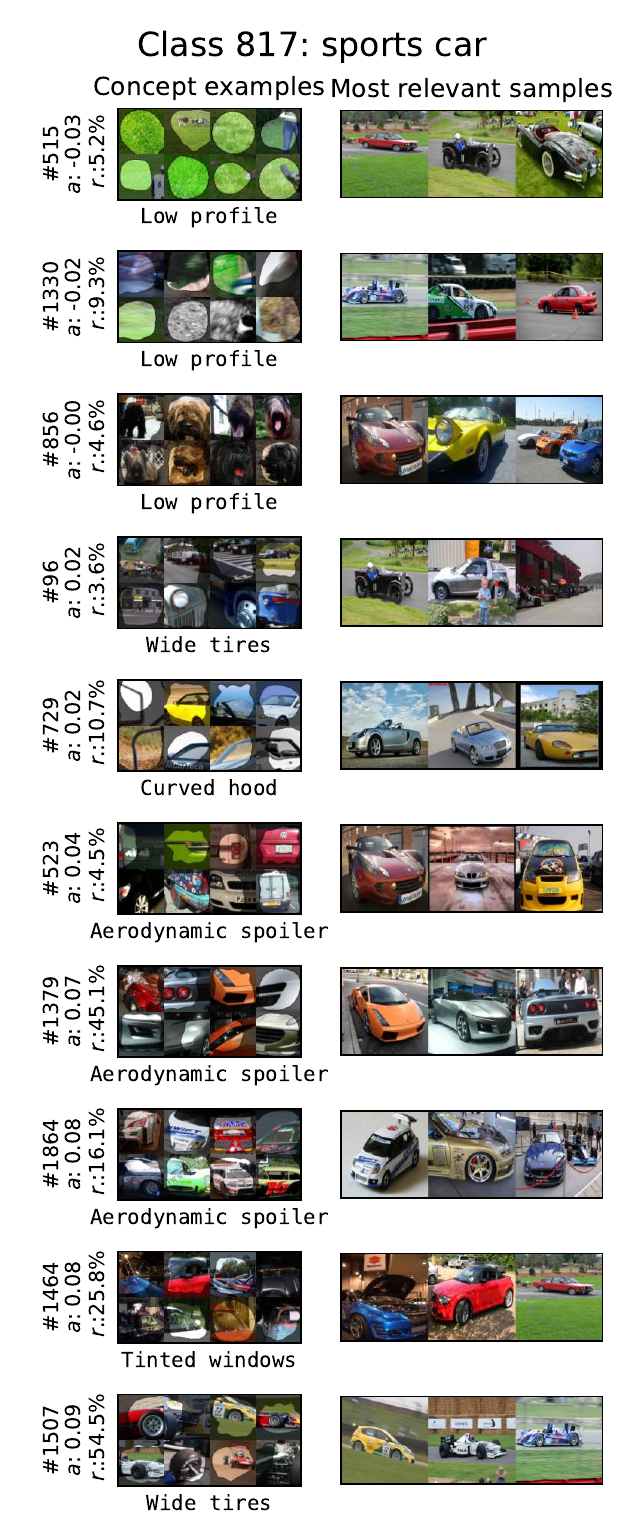}
            \includegraphics[width=0.32\textwidth]{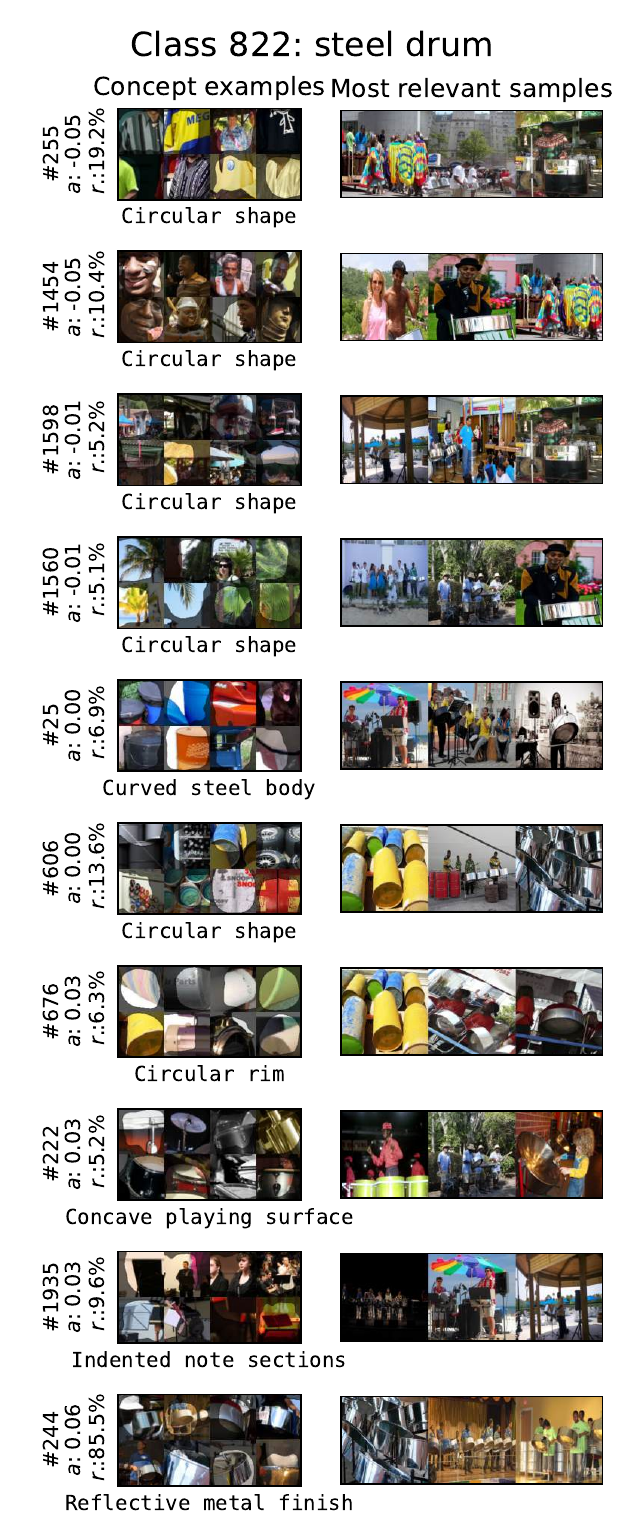}
    \caption{
    Alignment of top-10 most relevant neurons of a ResNet50v2's last feature layer for the classes of ``soap dispenser'', ``sports car'' and ``steel drum''. Besides highest alignment $a$ to an expected concept (label below concept examples), we provide the highest relevance scores $r$ on the test set of the class. Further depicted are three examples where a neuron is most relevant for a class.
    }
    \label{fig:app:explore:audit:audit_background_7}
\end{figure}
\begin{figure}[t]
    \centering
    \includegraphics[width=0.32\textwidth]{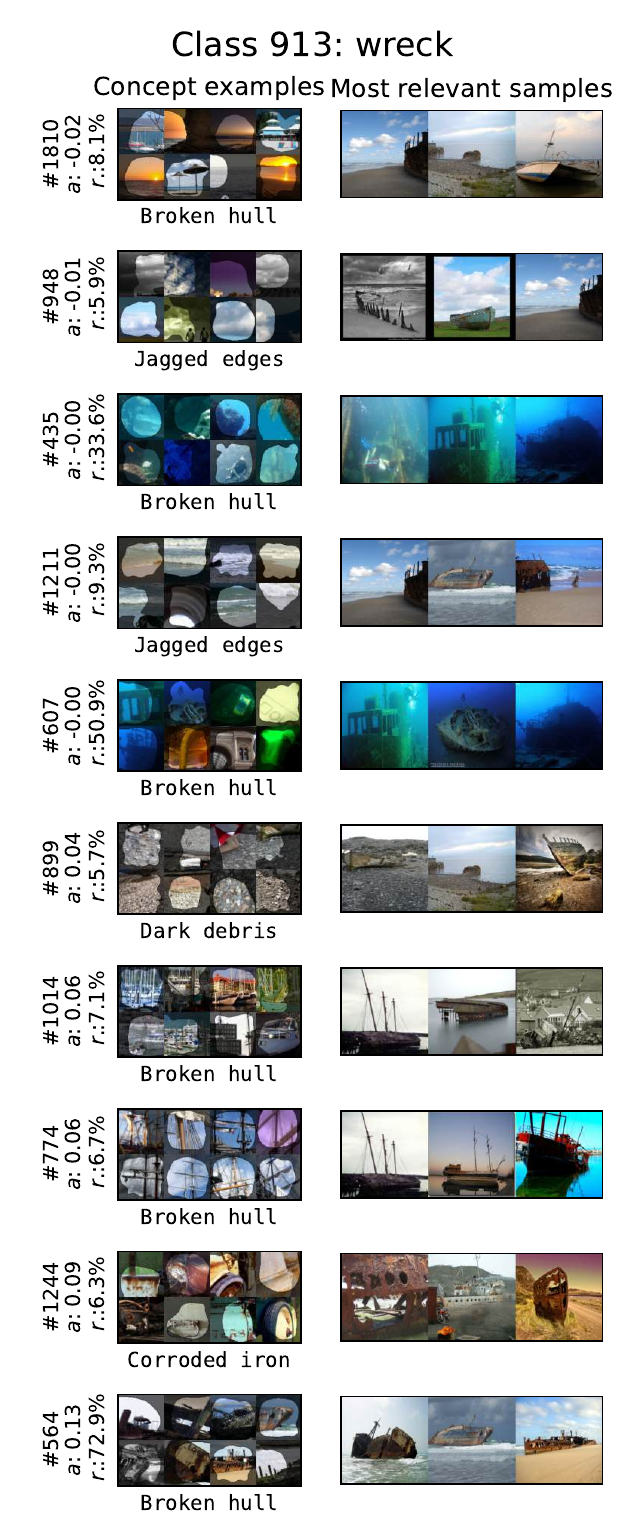}
        \includegraphics[width=0.32\textwidth]{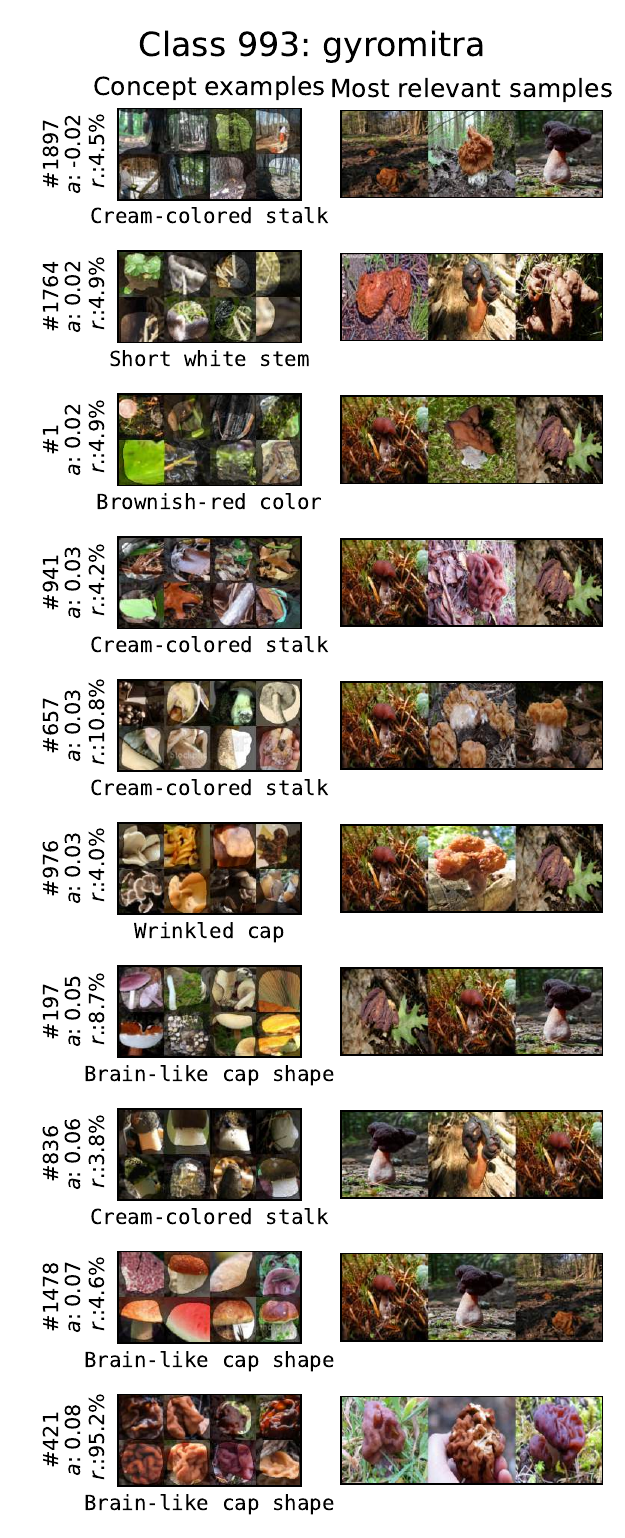}
    \caption{
    Alignment of top-10 most relevant neurons of a ResNet50v2's last feature layer for the classes of ``wreck'' and ``gyromitra''. Besides highest alignment $a$ to an expected concept (label below concept examples), we provide the highest relevance scores $r$ on the test set of the class. Further depicted are three examples where a neuron is most relevant for a class.
    }
    \label{fig:app:explore:audit:audit_background_8}
\end{figure}

The following expected concepts were defined for measuring alignment:

\paragraph*{Class 70} 
\texttt{Long, thin legs}, \texttt{Small, round body}, \texttt{Brownish-gray color}, \texttt{Legs outstretched}, \texttt{Crawling motion}, \texttt{Smooth exoskeleton}, \texttt{Tiny eyes}, \texttt{Jointed leg segments}, \texttt{Spider-like appearance}, \texttt{Delicate legs crossing}, \texttt{Resting on leaves}, \texttt{Thin, wiry legs}, \texttt{Clinging to surfaces}, \texttt{Contrasting body and leg colors}, \texttt{Huddled posture}, 
\paragraph*{Class 277} 
\texttt{Rusty red fur}, \texttt{Pointed snout}, \texttt{Bushy tail with white tip}, \texttt{Black paws}, \texttt{Alert ears}, \texttt{Sharp eyes}, \texttt{Slim body}, \texttt{White underbelly}, \texttt{Dark nose}, \texttt{Leaping pose}, \texttt{Camouflaged in grass}, \texttt{Running through snow}, \texttt{Curled up sleeping}, \texttt{Sneaky posture}, \texttt{Fluffy tail swishing}, 
\paragraph*{Class 294} 
\texttt{Dark brown fur}, \texttt{Thick fur}, \texttt{Powerful paws}, \texttt{Brown coat}, \texttt{Muscular build}, \texttt{Sharp claws}, \texttt{Round ears}, \texttt{Dark brown eyes}, \texttt{Broad shoulders}, \texttt{Coarse fur texture}, \texttt{Short tail}, \texttt{Strong jaw}, \texttt{Robust body form}, \texttt{Light brown muzzle}, \texttt{Heavy limbs}, \texttt{Curved claws}, 
\paragraph*{Class 309} 
\texttt{Transparent wings}, \texttt{Striped yellow and black body}, \texttt{Small antennae}, \texttt{Segmented abdomen}, \texttt{Hexagonal eye pattern}, \texttt{Fuzzy thorax}, \texttt{Pointed stinger}, \texttt{Tiny legs}, \texttt{Glossy wing texture}, \texttt{Round compound eyes}, \texttt{Oval-shaped body}, \texttt{Vibrant yellow stripes}, \texttt{Delicate wing veins}, \texttt{Thin legs with pollen sacs}, \texttt{Smooth exoskeleton}, 
\paragraph*{Class 336} 
\texttt{Thick brown fur}, \texttt{Short legs}, \texttt{Stocky body}, \texttt{Round ears}, \texttt{Curved claws}, \texttt{Flat snout}, \texttt{Bushy tail}, \texttt{Light brown underbelly}, \texttt{Coarse fur texture}, \texttt{Dark eyes}, \texttt{Short, sturdy limbs}, \texttt{Rounded body form}, \texttt{Sharp teeth}, \texttt{Padded paws}, \texttt{Grayish-brown coat}, 
\paragraph*{Class 340} 
\texttt{Striped coat}, \texttt{Mane of hair}, \texttt{Hoofed feet}, \texttt{Black and white stripes}, \texttt{Long tail}, \texttt{Rounded ears}, \texttt{Smooth fur}, \texttt{Slender legs}, \texttt{Muzzle with whiskers}, \texttt{Muscular body}, \texttt{Shiny coat}, \texttt{Narrow snout}, \texttt{Short mane}, \texttt{Contrasting color pattern}, \texttt{Coarse hair texture}, 
\paragraph*{Class 388} 
\texttt{Black and white fur}, \texttt{Round ears}, \texttt{Large paws}, \texttt{Plump body}, \texttt{Circular black eye patches}, \texttt{Soft fur texture}, \texttt{Flat nose}, \texttt{Stubby tail}, \texttt{Strong jaws}, \texttt{Muscular limbs}, \texttt{Thick coat}, \texttt{Wide face}, \texttt{Contrasting black and white markings}, \texttt{Rounded belly}, \texttt{Coarse paw pads}, 
\paragraph*{Class 390} 
\texttt{Slender body}, \texttt{Smooth skin}, \texttt{Elongated form}, \texttt{Dark green color}, \texttt{Pointed snout}, \texttt{Silky texture}, \texttt{Undulating movement}, \texttt{Pale underbelly}, \texttt{Slippery surface}, \texttt{Streamlined shape}, \texttt{Glossy sheen}, \texttt{Narrow fins}, \texttt{Curved tail}, \texttt{Sinous body form}, \texttt{Mottled brown pattern}, 
\paragraph*{Class 407} 
\texttt{Flashing emergency lights}, \texttt{Red and white exterior}, \texttt{Siren}, \texttt{Large rear doors}, \texttt{Wheeled stretcher}, \texttt{Bold cross symbol}, \texttt{Metallic chassis}, \texttt{Square body form}, \texttt{Reflective decals}, \texttt{Clear windshield}, \texttt{Rubber tires}, \texttt{Grippy floor texture}, \texttt{Bright headlamps}, \texttt{Compact interior}, \texttt{Smooth metal panels}, 
\paragraph*{Class 441} 
\texttt{Clear glass body}, \texttt{Rounded rim}, \texttt{Sturdy base}, \texttt{Smooth glass surface}, \texttt{Amber liquid color}, \texttt{Curved handle}, \texttt{Frosted glass texture}, \texttt{Tall cylindrical shape}, \texttt{Golden beer hue}, \texttt{Tapered glass form}, \texttt{Bubbly carbonation}, \texttt{Thick glass walls}, \texttt{Foamy white head}, \texttt{Translucent glass clarity}, \texttt{Cold glass touch}, 
\paragraph*{Class 483} 
\texttt{Stone walls}, \texttt{Tall towers}, \texttt{Wooden drawbridge}, \texttt{Gray bricks}, \texttt{Iron gates}, \texttt{Rounded turrets}, \texttt{Sturdy battlements}, \texttt{Dark dungeons}, \texttt{Narrow arrow slits}, \texttt{Moat surrounding}, \texttt{Flag flying high}, \texttt{Cobblestone courtyard}, \texttt{Heavy wooden doors}, \texttt{Fortified ramparts}, \texttt{Cold stone texture}, 
\paragraph*{Class 517} 
\texttt{Long metal boom}, \texttt{Yellow frame}, \texttt{Steel cables}, \texttt{Rotating cab}, \texttt{Hook block}, \texttt{Tall mast}, \texttt{Boom extension}, \texttt{Black counterweights}, \texttt{Lattice structure}, \texttt{Rubber tracks}, \texttt{Wire rope}, \texttt{Sleek hydraulic arms}, \texttt{Grippy control levers}, \texttt{Heavy-duty joints}, \texttt{Glossy metal surface}, 
\paragraph*{Class 594} 
\texttt{Vertical Lines}, \texttt{Golden strings}, \texttt{Wooden soundboard}, \texttt{Curved neck form}, \texttt{Polished wooden column}, \texttt{Smooth brass pedals}, \texttt{Elegant frame}, \texttt{Carved wooden body}, \texttt{Shiny metallic tuning pins}, \texttt{Glossy black finish}, \texttt{Ornate pillar design}, \texttt{Textured grip on strings}, \texttt{Deep mahogany color}, \texttt{Slender body form}, \texttt{Delicate string tension}, \texttt{Velvety string texture}, 
\paragraph*{Class 599} 
\texttt{Hexagonal cells}, \texttt{Golden amber color}, \texttt{Dripping honey}, \texttt{Wax texture}, \texttt{Symmetrical pattern}, \texttt{Sticky surface}, \texttt{Clustered structure}, \texttt{Thick honey-filled cells}, \texttt{Translucent wax}, \texttt{Crystalline form}, \texttt{Natural golden hue}, \texttt{Layered honeycomb slabs}, \texttt{Smooth honey flow}, \texttt{Bees crawling over it}, \texttt{Cut honeycomb with oozing honey}, 
\paragraph*{Class 648} 
\texttt{Glass mirror door}, \texttt{Metallic handle}, \texttt{White shelves}, \texttt{Smooth interior surface}, \texttt{Compact compartments}, \texttt{Small drawers}, \texttt{Reflective mirror finish}, \texttt{Hinged door}, \texttt{Plastic containers}, \texttt{Frosted glass}, \texttt{Rectangular frame}, \texttt{Transparent shelves}, \texttt{Polished edges}, \texttt{Glossy cabinet surface}, \texttt{Flat wooden panel}, 
\paragraph*{Class 718} 
\texttt{Wooden planks}, \texttt{Metal support beams}, \texttt{Sturdy pilings}, \texttt{Weathered surface}, \texttt{Rusty nails}, \texttt{Flat deck}, \texttt{Mooring cleats}, \texttt{Open water view}, \texttt{Railing along edges}, \texttt{Docking area}, \texttt{Rectangular shape}, \texttt{Salt-stained wood}, \texttt{Piling posts}, \texttt{Non-slip texture}, \texttt{Tide-worn edges}, 
\paragraph*{Class 754} 
\texttt{Wooden exterior}, \texttt{Circular tuning dial}, \texttt{Metal antenna}, \texttt{Speaker grille}, \texttt{Vintage knobs}, \texttt{AM/FM frequency display}, \texttt{Rectangular form}, \texttt{Portable handle}, \texttt{Static-filled screen}, \texttt{Red tuning light}, \texttt{Compact transistor shape}, \texttt{Glossy buttons}, \texttt{Retro design}, \texttt{LCD frequency screen}, \texttt{Digital display}, 
\paragraph*{Class 774} 
\texttt{Flat sole}, \texttt{Adjustable straps}, \texttt{Open-toe design}, \texttt{Leather material}, \texttt{Rubber outsole}, \texttt{Padded footbed}, \texttt{Colorful straps}, \texttt{Woven texture}, \texttt{Buckles or Velcro}, \texttt{Ankle support}, \texttt{Curved arch}, \texttt{Lightweight form}, \texttt{Breathable fabric}, \texttt{Thick heel}, \texttt{Smooth lining}, 
\paragraph*{Class 784} 
\texttt{Metallic shaft}, \texttt{Flathead tip}, \texttt{Phillips head tip}, \texttt{Ergonomic handle}, \texttt{Rubber grip}, \texttt{Chrome finish}, \texttt{Magnetic tip}, \texttt{Slender shaft}, \texttt{Red and black handle}, \texttt{Textured grip}, \texttt{Tapered tip}, \texttt{Polished steel}, \texttt{Stubby handle form}, \texttt{Ridged handle surface}, \texttt{Hexagonal shaft}, 
\paragraph*{Class 792} 
\texttt{Metallic gray blade}, \texttt{Wooden tan handle}, \texttt{Curved metal edge}, \texttt{Smooth metal grip}, \texttt{Rust brown surface}, \texttt{Coarse wooden texture}, \texttt{Matte black finish}, \texttt{Polished silver shaft}, \texttt{Flat blade form}, \texttt{Rough wooden handle}, \texttt{Tapered metal point}, \texttt{Dark footstep}, \texttt{Ridged blade surface}, \texttt{Olive green shaft}, \texttt{Grainy wooden texture}, 
\paragraph*{Class 801} 
\texttt{Breathing tube}, \texttt{Mouthpiece}, \texttt{Curved top}, \texttt{Flexible silicone}, \texttt{Clear plastic body}, \texttt{Rubber strap}, \texttt{Soft bite valve}, \texttt{Transparent mask attachment}, \texttt{Curved tube form}, \texttt{Anti-splash guard}, \texttt{Adjustable strap}, \texttt{Compact design}, 
\paragraph*{Class 804} 
\texttt{Transparent bottle}, \texttt{Pump top}, \texttt{Liquid soap inside}, \texttt{Foaming nozzle}, \texttt{Matte plastic body}, \texttt{Chrome finish pump}, \texttt{Wall-mounted version}, \texttt{Curved bottle shape}, \texttt{Sleek, minimalist design}, \texttt{Refillable container}, \texttt{Dripping soap effect}, \texttt{Frosted glass body}, \texttt{Colored liquid soap}, \texttt{Soft-touch material}, \texttt{Push-button mechanism}, 
\paragraph*{Class 817} 
\texttt{Sleek body}, \texttt{Low profile}, \texttt{Glossy red paint}, \texttt{Aerodynamic spoiler}, \texttt{Leather seats}, \texttt{Shiny chrome rims}, \texttt{Wide tires}, \texttt{Curved hood}, \texttt{Dual exhaust pipes}, \texttt{Tinted windows}, \texttt{Carbon fiber accents}, \texttt{LED headlights}, \texttt{Smooth steering wheel}, \texttt{Compact form}, \texttt{Metallic finish}, 
\paragraph*{Class 822} 
\texttt{Shiny metallic surface}, \texttt{Circular shape}, \texttt{Concave playing surface}, \texttt{Silvery gray color}, \texttt{Indented note sections}, \texttt{Polished steel texture}, \texttt{Raised center dome}, \texttt{Circular rim}, \texttt{Reflective metal finish}, \texttt{Suspended on a stand}, \texttt{Hammered metal texture}, \texttt{Brightly polished edges}, \texttt{Multiple circular note areas}, \texttt{Matte silver base}, \texttt{Curved steel body}, 
\paragraph*{Class 913} 
\texttt{Twisted metal}, \texttt{Broken hull}, \texttt{Rusty surface}, \texttt{Shattered glass}, \texttt{Dark debris}, \texttt{Scattered fragments}, \texttt{Bent framework}, \texttt{Corroded iron}, \texttt{Cracked structure}, \texttt{Faded paint}, \texttt{Exposed wires}, \texttt{Jagged edges}, \texttt{Dilapidated form}, \texttt{Burnt-out interior}, \texttt{Rough, damaged texture}, 
\paragraph*{Class 993} 
\texttt{Wrinkled cap}, \texttt{Brownish-red color}, \texttt{Curled edges}, \texttt{Short white stem}, \texttt{Brain-like cap shape}, \texttt{Irregular folds}, \texttt{Earthy brown tones}, \texttt{Hollow interior}, \texttt{Bulbous form}, \texttt{Textured surface}, \texttt{Wavy contours}, \texttt{Soft, spongy texture}, \texttt{Twisted cap structure}, \texttt{Cream-colored stalk},

\subsubsection{Aligned Models Are More Interpretable and Performant}
\begin{figure}[t]
    \centering
    \includegraphics[width=0.99\textwidth]{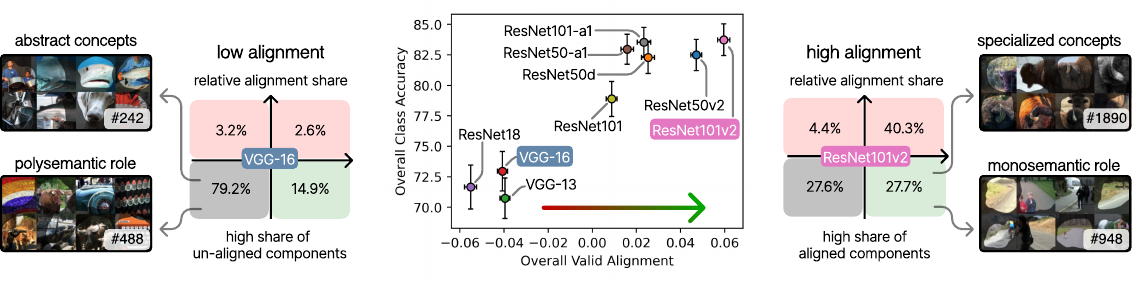}
    \caption{    
    Models with a high alignment to valid concepts are often more performant and interpretable.
    \emph{Left}: A VGG-16 shows many polysemantic units, or units that encode for abstract concepts, ultimately resulting lower alignment scores.
    \emph{Right}: A ResNet101v2 on the other hand, has more specialized and monosemantic units,
    which results in overall higher alignment scores.
    }
    \label{fig:app:explore:audit:audit_interpretability}
\end{figure}

This section provides details for \cref{sec:results:audit},
where we analyse popular pre-trained models on the ImageNet dataset,
and observe strong variations \wrt their  valid alignment.
The reason often lies in the share of knowledge that is neither aligned to valid nor spurious concepts, as demonstrated for the VGG-16 in \cref{fig:app:explore:audit:audit_interpretability}.
For instance, the VGG-16 contains several polysemantic components that perform multiple roles in decision-making,
which generally reduces alignment.
On the other hand,
we also find more abstract concepts,
\eg, encoding for gray/silver-like texture.
On the other hand,
more performant and wider models tend to have more specialized (\eg, class-specific) and monosemantic model components such as the ResNet101v2,
later quantified in \cref{sec:results:evaluate}.
Overall,
higher-performing models with larger feature spaces (more neurons per layer, as, \eg, ResNet50 and ResNet101) show thus greater alignment scores throughout experiments detailed in \cref{fig:app:explore:audit:audit_interpretability} (\emph{middle}).

\subsection{Medical Use Case}
\label{app:medical}
This section provides further details for \cref{sec:results:medical},
where we demonstrate \gls{ours} in a medical use case and debug a VGG-16 with Batch Normalization layers on the ISIC 2019 challenge dataset, as detailed in \cref{app:experimental-settings}.

For all experiments,
we leverage the CLIP model provided by \cite{yang2024textbook},
collecting concept examples $\mathcal{E}$ on the test set, and using $|\mathcal{E}| = 20$ for generating semantic embeddings \gls{cone}.
Notably,
we reduce the set size to 20 compared to 30 for ImageNet,
as the test set of ISIC is much smaller (2,533 compared to 50,000 samples). For ISIC, the activation values for the 20-st most activating sample are on average still corresponding to more than 40\,\% of the highest activation value for a neuron.

\subsubsection{Concept Labels}
\label{sec:app:medical:labels}
We use the following labels to annotate concepts, where we mostly adhere to the ABCDE-rule~\cite{duarte2021-md}:
\small{
\paragraph{Asymmetry}
\texttt{asymmetric lesion}, \texttt{asymmetrical lesion}, \texttt{uneven form}, \texttt{irregular form}, \texttt{unsymmetrical lesion}, \texttt{irregular dark streaks}, \texttt{irregular pigment network}, \texttt{negative pigment network}, \texttt{irregular pseudonetwork}, \texttt{off-centered blotch}, \texttt{atypical dots or globules}, \texttt{regression structures}, \texttt{atypical pigment network}, \texttt{ulcer}, \texttt{irregular sws}, \texttt{unsymmetric lesion}

\paragraph{Border}
\texttt{irregular border}, \texttt{irregular fuzzy border}, \texttt{uneven border}, \texttt{ragged border}, \texttt{jagged border}, \texttt{fuzzy and slightly jagged border}, \texttt{fuzzy and strongly jagged border}, \texttt{notched border}, \texttt{indistinct border}, \texttt{obscure border}, \texttt{poorly defined border}, \texttt{ill-defined border}, \texttt{irregular blurred border}, \texttt{scalloped border}

\paragraph{Colour}
\texttt{variegated color}, \texttt{multicolored}, \texttt{multicoloured}, \texttt{polychromatic}, \texttt{polychrome}, \texttt{blue-white veil}, \texttt{crystalline structures}, \texttt{varying color}, \texttt{changing color}, \texttt{different color}, \texttt{multiple color}, \texttt{many color}, \texttt{several color}, \texttt{various color}, \texttt{shades of tan}, \texttt{shades of brown}, \texttt{shades of black}, \texttt{shades of black and brown}, \texttt{shades of blue}, \texttt{red and blue color}, \texttt{red and black color}, \texttt{blue color}, \texttt{black and brown color},
\texttt{milky-red globules}~\cite{braun2009dermoscopy}

\paragraph{Diameter}
\texttt{Large diameter}, \texttt{Large size}, \texttt{larger than 6mm}, \texttt{large mole}, \texttt{large growth}, \texttt{large patch}, \texttt{large lesion}, \texttt{large spot}

\paragraph{Evolving}
\texttt{bleeding}, \texttt{changing}, \texttt{growing}, \texttt{crusty}, \texttt{crusty surface}, \texttt{warty surface}, \texttt{warty}, \texttt{warty lesion}, \texttt{scaly}, \texttt{scaly lesion}

\paragraph{Other}
\begin{itemize}
    \item \texttt{vascular lesion} is a sub-class of the ISIC dataset for ``other'',
    \item \texttt{actinic keratosis} is a sub-class of the ISIC dataset for ``other'',
    \item \texttt{dermatofibroma} is a sub-class of the ISIC dataset for ``other'',
    \item \texttt{basal cell carcinoma} is a sub-class of the ISIC dataset for ``other'',
    \item \texttt{dermatofibroma} is a sub-class of the ISIC dataset for ``other'',
    \item \texttt{white and yellowish structures}
is an indication of the Squamous cell carcinoma sub-class of the ISIC dataset \cite{salafranca2024dermoscopy},
    \item \texttt{white streaks}, \texttt{crystalline structures}, \texttt{white scar-like areas}, \texttt{scab, scabbed}, \texttt{white scales} and \texttt{white dots}
    (can be indication of melanoma, but also the Basal cell carcinoma sub-class of the ISIC dataset~\cite{navarrete2016association,verzi2018diagnostic,thomson2020interventions,balagula2012significance}),
    \item \texttt{large blue-gray ovoid nests} are highly suggestive of basal cell carcinoma~\cite{lallas2015dermoscopy},
    \item \texttt{milia-like cysts}, \texttt{fingerprint-like structures} and \texttt{moth-eaten borders} are an indication for sub-class of ISIC (benign keratosis)~\cite{minagawa2017dermoscopy},
    \item \texttt{red, maroon, or black lacunae}, and \texttt{widespread red-blue lacunes}~\cite{kim2012dermoscopy},
    \item \texttt{clear cell acanthoma} is a benign epidermal tumor, 
    \item \texttt{reactive haemangioma} is a common benign vascular tumor,
    \item \texttt{vascular blush}~\cite{braun2009dermoscopy},
    \item \texttt{cherry angioma} common asymptomatic vascular skin lesions~\cite{nazer2020cherry},
    \item regular ABCDE rule variants: \texttt{smooth border}, \texttt{uniform color}, \texttt{regular border}, \texttt{regular fuzzy border}, \texttt{regular hazy border}, \texttt{regular blurred border}, \texttt{uniform color}, \texttt{regular pigment network}, \texttt{small size}, \texttt{smaller than 6mm}, \texttt{smaller than 4mm}, \texttt{small diameter}, \texttt{small mole}, \texttt{small growth}, \texttt{small patch}, \texttt{small lesion}, \texttt{pink growths}, \texttt{regular dark streaks}, \texttt{red-bluish-black homogeneous areas},
    \item \texttt{talon noir, black heel} is a harmless coloration~\cite{choudhury2023talon},
\end{itemize}

\paragraph{Spurious}
\texttt{redness of the skin}, \texttt{vascular skin}, \texttt{red skin}, \texttt{dark corner visible}, \texttt{ink marker}, \texttt{plaster}, \texttt{ruler visible}, \texttt{white spots}, \texttt{fingernail}, \texttt{white reflections}, \texttt{measuring scale}, \texttt{measurement scale bar}, \texttt{size marker}, \texttt{purple skin marker}, \texttt{strongly haired}, \texttt{hairs}, \texttt{hairy}, \texttt{ink marker}, \texttt{band-aid}, \texttt{colorful patch}, \texttt{patch or band-aid}, \texttt{ruler}, \texttt{blue coloured band-aid or patch}, \texttt{orange coloured band-aid or patch}
}

\subsubsection{Training}
We train models using a sparsity loss (see \cref{tab:app:interptretability:optimizing_sparsity} for loss details)
with a regularization strength of $\lambda=1.0$ for 300 epochs with an initial learning rate of $0.01$ which is decreased by a factor of 10 after 240 and 290 epochs, respectively.
The optimization algorithm used is stochastic gradient descent.

\subsubsection{Evaluation}
\label{sec:app:medical:evaluation}

Instead of Stable Diffusion (which is not available for skin lesion data),
we manipulate the available test data to evaluate whether pruning or retraining had a positive effect on model reliability and robustness.

``Red coloured skin'' is artificially added to clean samples by adding a red hue over images.
Concretely,
we transform the RGB channels $r$, $g$ and $b$ of an image by
\begin{align}
    r \rightarrow 0.7r  + 0.3 \times 231 \\
    g \rightarrow 0.7g  + 0.3 \times 128 \\
    b \rightarrow 0.7b  + 0.3 \times 151
\end{align}
which corresponds to overlaying a uniform colour of [231, 128, 151] (mean RGB values of a test samples with red skin) with a transparency of 30\,\%.

Further,
\texttt{ruler} and \texttt{band-aid} are modelled according to the Reveal2Revise~\cite{pahde2023reveal} framework.
Specifically,
we first generate artefact localizations on samples from the test set.
Here,
we use support vector machines to generate a \gls{cav} on activations of the last feature layer.
Using this \gls{cav},
we generate the artefact localizations by explaining, \ie, computing input heatmaps \wrt, the dot product between latent activations and \gls{cav}.
Subsequently,
the resulting ``soft'' masks are blurred using a Gaussian blur (kernel size of 41 pixels),
and low values amplified by taking values to the power of $0.6$.
After a final normalization to the maximum value, \eg, scores are between 0 and 1, we result in a localization $M\in\mathbb{R}^{224\times224}$ with $M_{ij}\in[0, 1]$.
Another image $D\in\mathbb{R}^{3\times224\times224}$ is then poisoned by overlaying the artifact of image $A\in\mathbb{R}^{3\times224\times224}$ via
\begin{equation}
    D_\text{poisoned} = D\circ(1-M) + A\circ M  ~,
\end{equation}
where $\circ$ corresponds to the Hadamard product, and $M$ is applied to each colour channel.

\begin{figure}[t]
    \centering
    \includegraphics[width=0.99\textwidth]{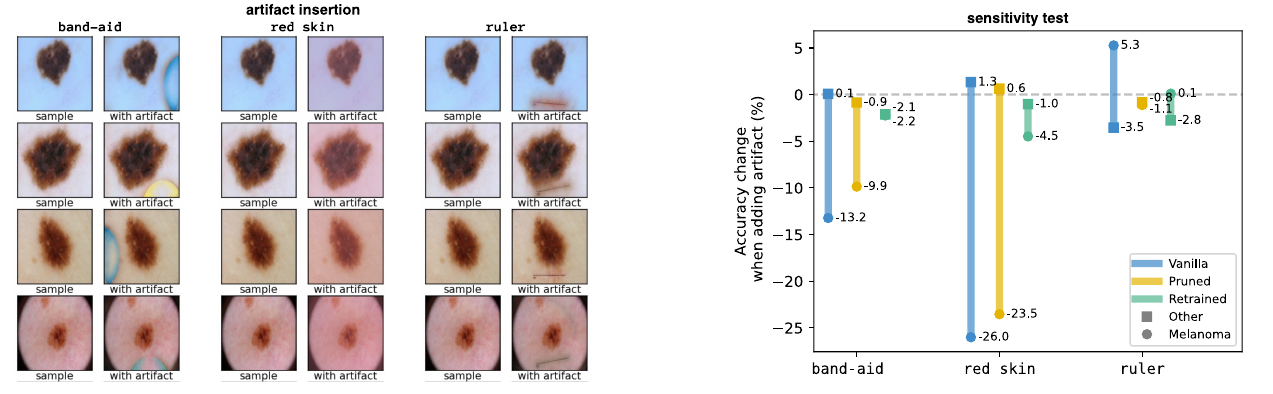}
    \caption{Testing model artefact sensitivity by adding artefacts to clean input samples, with examples given for \texttt{band-aid}, \texttt{red skin}, \texttt{ruler} (\emph{left}). 
    In the bar chart (\emph{right}),
    the change in test set accuracy for the vanilla, pruned and retrained model, when artefacts (\texttt{band-aid}, \texttt{red skin}, \texttt{ruler}) are added to clean input samples for the medical use case is given.
    Compared to pruning,
    only retraining results in strongly reduced sensitivities for all artefacts.
    }
    \label{fig:app:medical:medical_details}
\end{figure}

Exemplary data manipulations and the effect on test accuracy when data is poisoned is shown in \cref{fig:app:medical:medical_details},
where we provide concrete values for the accuracy change that is also illustrated in \cref{fig:results:medical} (\emph{bottom}).

\begin{figure}[t]
    \centering
    \includegraphics[width=0.99\textwidth]{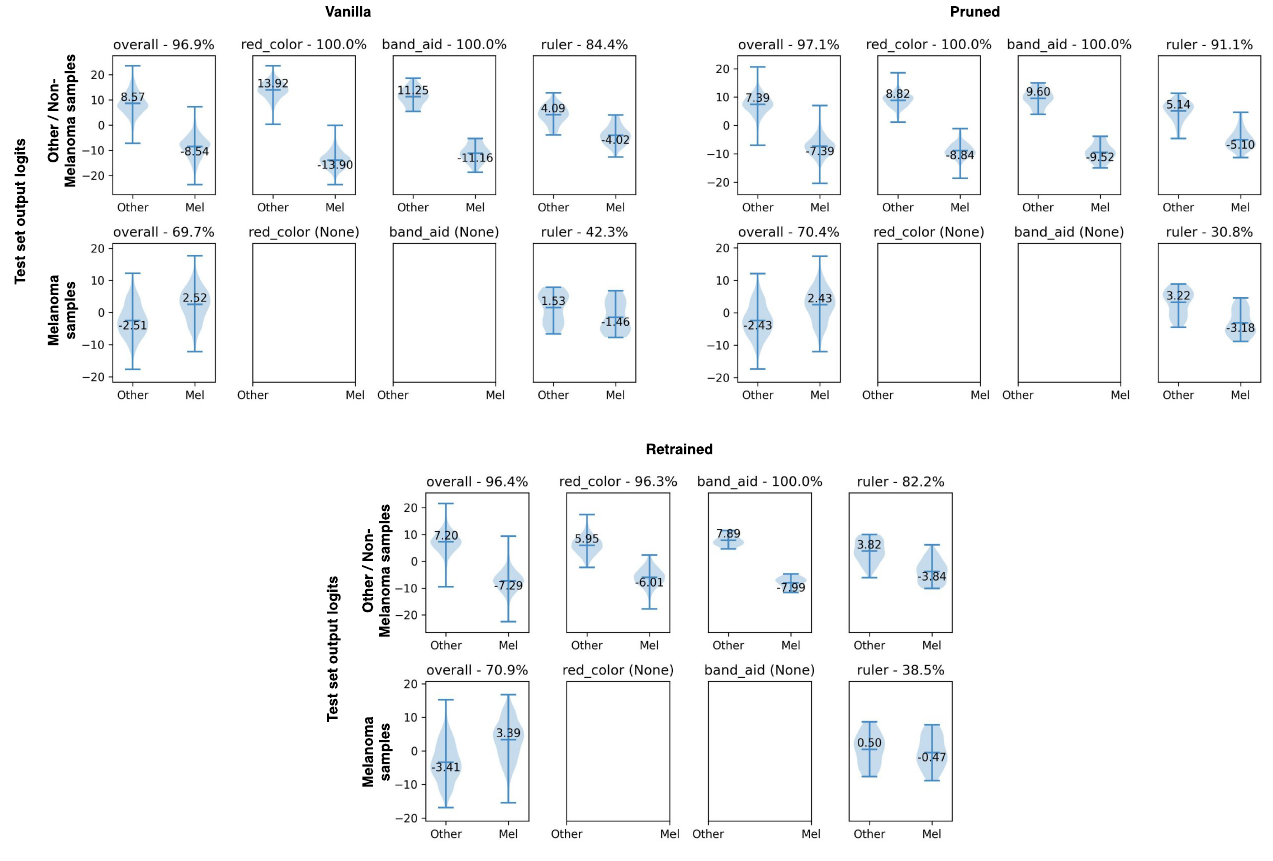}
    \caption{Output logits (vertical axis) and accuracy (sub-plot title) on subsets of the test set for the vanilla model (\emph{top left}), pruned vanilla model (\emph{top right}) and retrained model (\emph{bottom centre}).
    The output logits (mean depicted in the centre of the distribution) are given for the ``Melanoma'' and ``Other'' class.
    For the artefacts \texttt{red colour} and \texttt{band-aid}, not samples with melanoma are available.
    Especially for \texttt{red colour} and \texttt{band-aid} the logits magnitudes are reduced when pruning or retraining, indicating a reduced over-fitting on these spurious artefacts.
    }
    \label{fig:app:medical:medical_subset_accuracy}
\end{figure}

Alternatively,
we present the output logits and test set accuracies for specific subsets and all models in \cref{fig:app:medical:medical_subset_accuracy}.
\setcounter{figure}{0}
\setcounter{table}{0}\setcounter{equation}{0}
\section{Human-Interpretability Measures}
\label{app:interpretability}

\begin{figure}[t]
    \centering
    \includegraphics[width=0.99\textwidth]{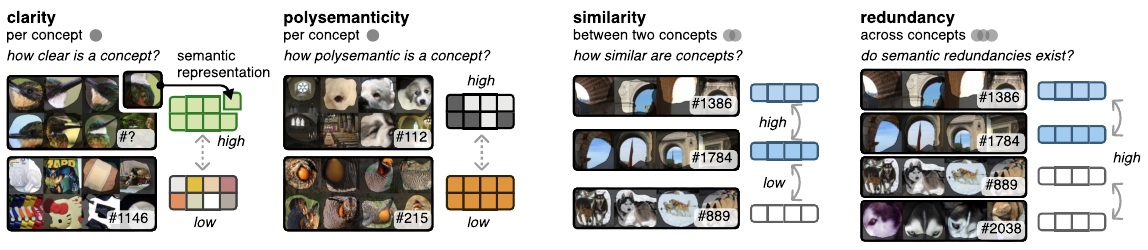}
    \caption{We introduce easy to compute human-interpretability measure that are useful to rate and improve model interpretability.
    Overall,
    we propose four measures:
    ``clarity'' for how clear and easy it is to understand the common theme of concept examples,
    ``polysemanticity'' describes if multiple distinct semantics are present in the concept examples,
    ``similarity'' for the similarity of concepts,
    and ``redundancy'' which describes the degree of redundancies in a set of concepts.
    The measures are based on transforming concept examples into their semantic representation.
    }
    \label{fig:app:interpretability:overview}
\end{figure}

In \cref{sec:methods:interpretability} and illustrated in \cref{fig:app:interpretability:overview},
we propose four measures used to quantify the human-interpretability of concepts.
All measures are based on transforming the concept examples $\mathcal{E}_k$ for a neuron $k$ into the latent feature space of a foundation model $\mathcal{F}$.
Concretely, we firstly introduced
\textbf{``clarity''} for how clear and easy it is to understand the common theme of concept examples by
evaluating the similarity between each pair of concept examples from a set $\mathcal{E}_k$.
The clarity measure is highly related to works of~\cite{kalibhat2023identifying,li2024evaluating}.

Further,
\textbf{``polysemanticity''} describes if multiple distinct semantics are present in the concept examples, computed by clustering concept examples of set $\mathcal{E}_k$ and evaluating the similarity across the cluster's semantic embeddings.

Thirdly,
\textbf{``similarity''} measures the similarity of two concepts described by set $\mathcal{E}_k$ and $\mathcal{E}_l$ by measuring cosine similarity between the pooled feature vector of each concept example set (the semantic embedding of each neuron).

Lastly, \textbf{``redundancy''} describes the degree of redundancies in a set of concepts by measuring the average maximal similarity across concepts.

\paragraph{Experimental setting}
Throughout experiments for human-interpretability,
the default foundation model used is DINOv2.

\subsection{User Study}
\label{app:interpretability:study}
The following sections provide details for the user studies performed to evaluate alignment of interpretability measures (similarity, clarity and polysemanticity) with human perception.

\paragraph{Ethics Approval}
The Ethics Commission of the Fraunhofer Heinrich Hertz Institute provided guidelines for the study procedure and determined that no protocol approval is required. Informed consent has been obtained from all participants. 

\subsubsection{Similarity Measure}
We in the following provide details \wrt study design, data filtering and results.

\textbf{Study design:}
\begin{figure}[t]
    \centering
    \includegraphics[width=0.99\textwidth]{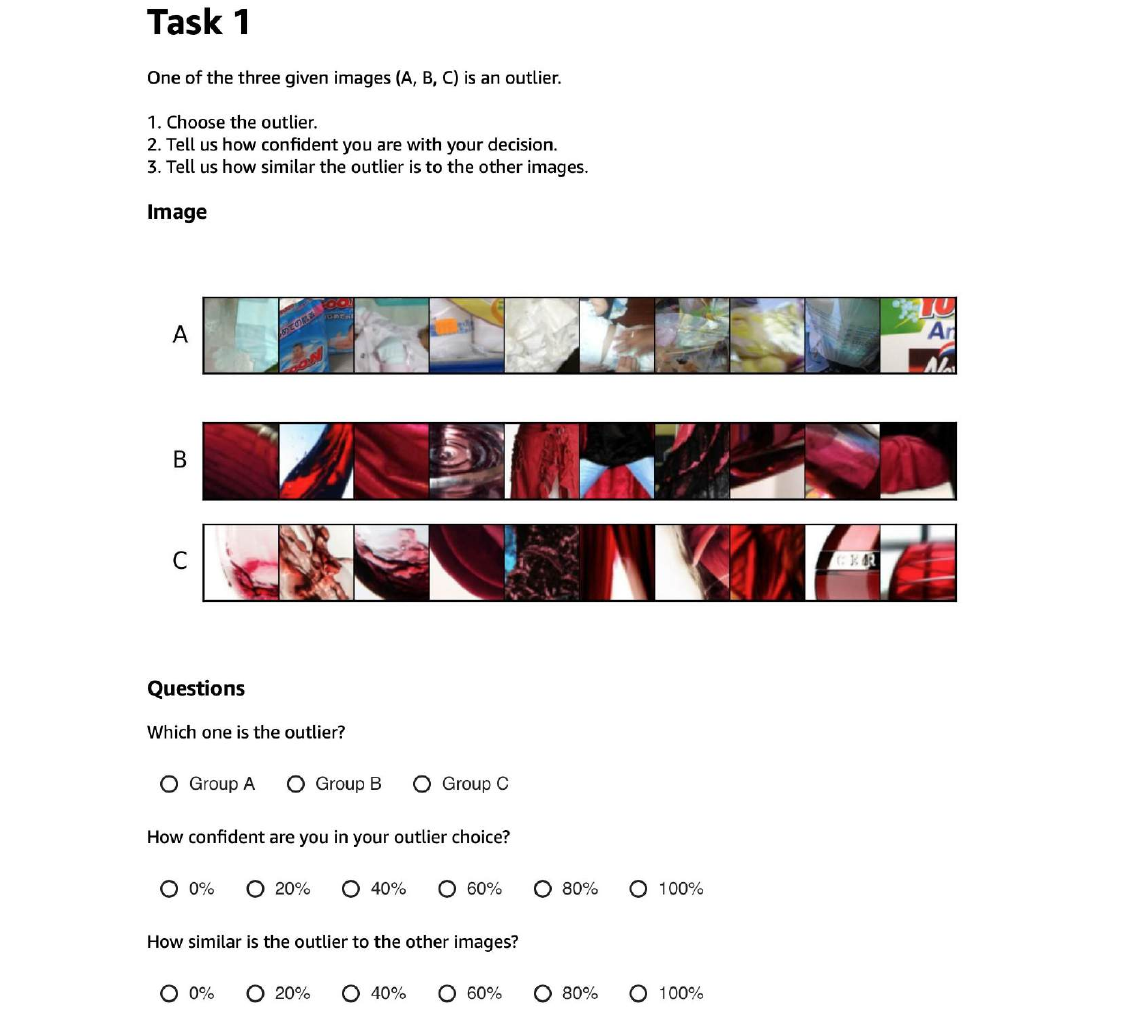}
    \caption{Study design to evaluate alignment of our similarity measure with human perception.
    }
    \label{fig:app:interpretability:study:similarity_study}
\end{figure}
We evaluate concept similarity (as perceived by humans) in an odd-one-out experiment,
where participants are asked to detect the outlier out of three sets of concept examples (``A'', ``B'', ``C''),
as shown in \cref{fig:app:interpretability:study:similarity_study}.
Concretely,
we choose two neurons randomly,
and create the three sets as follows:

``A'' corresponds to 10 concept examples out of the top 20 most activating concept examples of the first neuron (every second example is taken).

``B'' corresponds to 10 concept examples out of the top 20 most activating concept examples of the second neuron (every second example is taken).

``C'' corresponds to 10 concept examples out of the top 20 most activating concept examples of the second neuron (every second example is taken, but beginning with the second).

Subsequently,
the labels ``A'', ``B'', ``C'' are shuffled again.
For each odd-one-out task,
we further question participants for their confidence and the similarity of images.
All in all,
we create six experiments with 50 of such sets for the ResNet50v2 (four times), ResNet50 and ResNet34, using neurons of the penultimate layer.

\textbf{Compensation:}
The compensation for participation is divided into two parts: a base payment, and a bonus. 
Firstly, participation will receive a base payment of $2.0$ US dollars (estimated time per study are 15 minutes) if their performance was above 41\,\% (represents statistically significantly better than random performance according to a binomial test with 50 draws and success probability of 33\,\%). 
Performance is measured by the number of correctly recognized outliers in the odd-one-out task. 
The participants are further warned that responses
corresponding to random guessing are rejected, leading to no payout. 
In addition, participants are motivated to perform well by giving a bonus of 1.0 US dollar when performance was above 75\,\%,
or 1.5 US dollar when performance was above 85\,\%.

\textbf{Data filtering:}
We remove study results where the overall detection accuracy is below 55\%,
which corresponds to a statistically significant result above random performance ($p<0.002$) according to a binomial test.
We further expect a negative correlation between confidence and similarity values perceived by humans.
Thus,
we additionally filter out results where the correlation is above 0.15.
Overall,
we receive 59 of 108 experimental results after filtering.

\textbf{Results:}
\begin{figure}[t]
    \centering
    \includegraphics[width=0.99\textwidth]{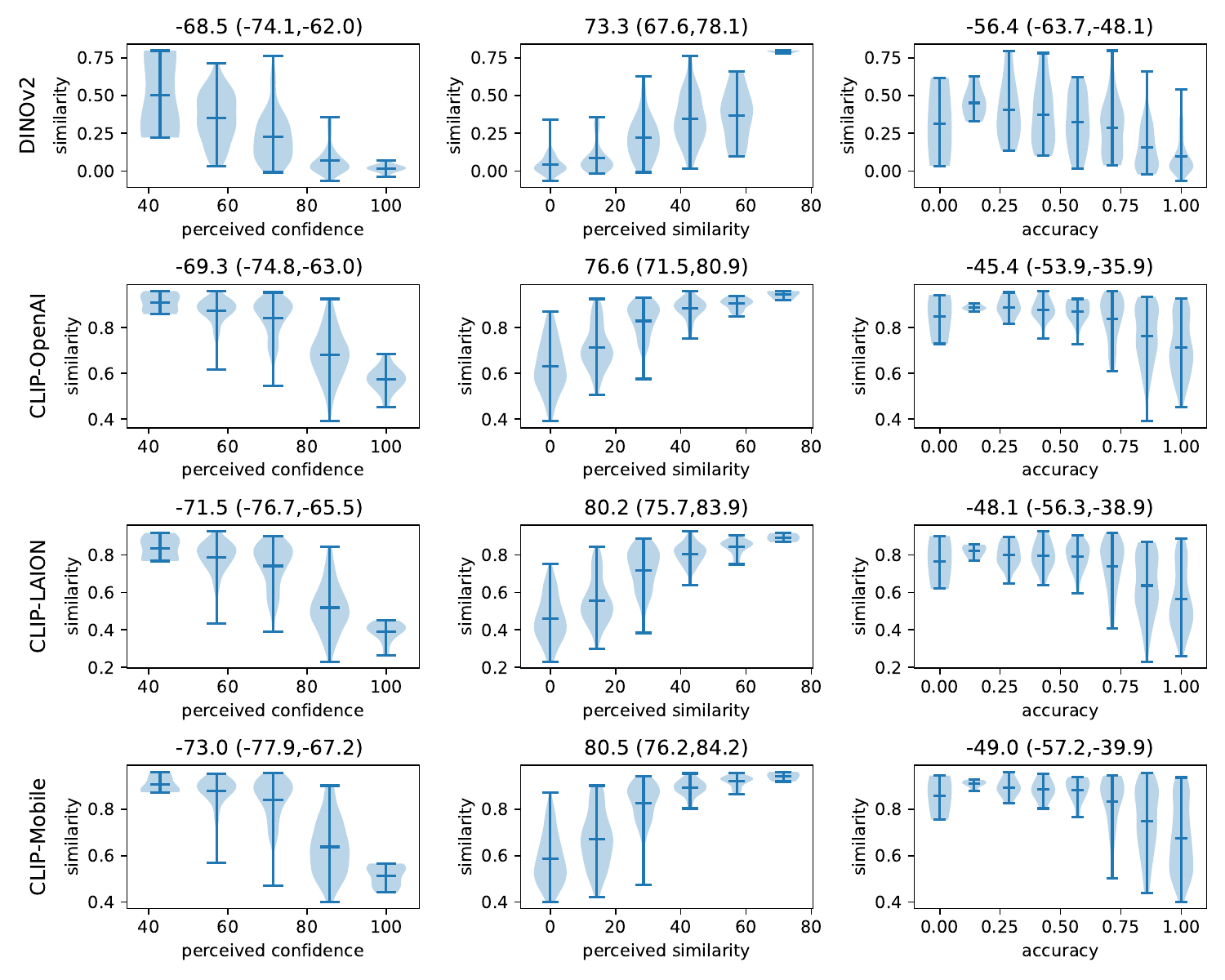}
    \caption{Correlation analysis for the user study regarding ``similarity'', where an odd-one-out task is performed, where participants are asked about their confidence and the similarity of shown concepts. 
    As the setting is controlled,
    we can also compute an accuracy.
    Overall,
    we see high correlations between our concept similarity measure and perceived confidence, similarity and accuracy (given above each violin plot with confidence intervals in parentheses).
    }
    \label{fig:app:interpretability:study:similarity_correlation}
\end{figure}
We have three semantic embeddings (given by a foundation model),
one for each concept set $\mathcal{E}$.
Denoting $\boldsymbol{\vartheta}_\text{odd}$ as the outlier semantic embedding,
and  $\boldsymbol{\vartheta}_\text{a}$ and $\boldsymbol{\vartheta}_\text{b}$ as the embeddings of the pair,
then we compute a similarity score $s$ of 
\begin{equation}
    s = \frac{1}{2} \left(s_\text{cos}(\boldsymbol{\vartheta}_\text{odd}, \boldsymbol{\vartheta}_\text{a}) + s_\text{cos}(\boldsymbol{\vartheta}_\text{odd}, \boldsymbol{\vartheta}_\text{b}) \right)~,
\end{equation}
where $s_\text{cos}$ denotes cosine similarity.

In \cref{fig:app:interpretability:study:similarity_correlation},
the similarity of concept example sets (as measured via foundation models) is shown against perceived confidence (\emph{left}),
perceived similarity (\emph{middle}) and detection accuracy (\emph{right}).
For each case,
Pearson correlation scores are given with a corresponding 95\,\% confidence interval.

\begin{figure}[t]
    \centering
    \includegraphics[width=0.99\textwidth]{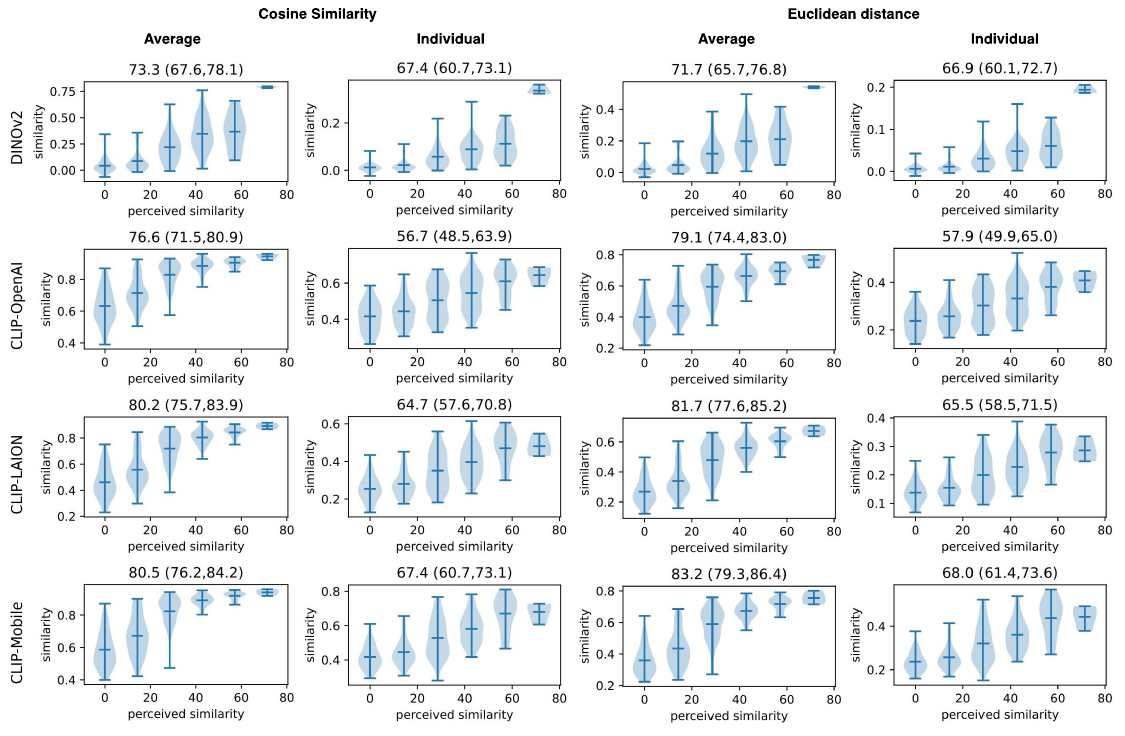}
    \caption{Ablation study via correlation analysis for the user study regarding ``similarity'', where an odd-one-out task is performed, where participants are asked about their confidence and the similarity of shown concepts. 
    In all violin plots,
    correlation between perceived similarity and our concept similarity measures are computed (displayed above each violin plot with confidence intervals in parentheses).
    Here,
    we use on the one hand side either cosine similarity or Euclidean distance to measure similarities,
    where no large differences are seen.
    Performing an averaging of semantic feature vectors (one for a whole set of concept examples $\mathcal{E}$) shows higher alignment, compared to computing similarities across individual feature vectors (for each example in $\mathcal{E}$).
    }
    \label{fig:app:interpretability:study:similarity_correlation_ablation}
\end{figure}
In \cref{fig:app:interpretability:study:similarity_correlation_ablation},
we compare the correlation between computed similarity and perceived similarity using either cosine similarity or Euclidean distance as a distance measure.
Additionally,
we either compute distances across each concept examples of a set, or, as per default, compare the mean embeddings (see \cref{sec:methods:interpretability}).
Whereas there is so significant difference between distance measures,
using the mean embedding instead of individual embeddings for each concept example set results in higher correlation scores.

\begin{figure}[t]
    \centering
    \includegraphics[width=0.99\textwidth]{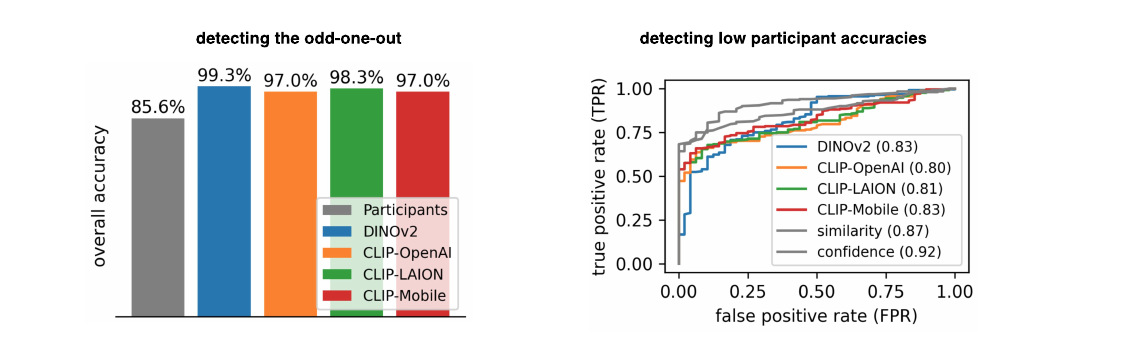}
    \caption{Further results for the user study for measuring alignment of computed concept similarity with human perception via an odd-one-out experiment.
    \emph{Left}: Foundation models can considerably better detect the odd-one-out compared to humans.
    \emph{Right}: When aiming to detect cases where participants have an accuracy below 65\,\%,
    the perceived similarity or confidence is a better indicator than the computed similarity between outlier and correct concept example embeddings.
    }
    \label{fig:app:interpretability:study:similarity_study_additional_results}
\end{figure}
Overall,
the participants found in 85,6\,\% of cases the correct outlier.
This stands in contrast to using the similarities of semantic embeddings,
where we receive over 97\,\% of accuracy when assuming that the correct pair corresponds to the one with the highest pairwise similarity,
as shown in \cref{fig:app:interpretability:study:similarity_study_additional_results} (\emph{left}).
This indicates,
that foundation models might be able to perceive differences or similarities better than humans.
However,
notably, participants on Amazon Mechanical Turk do not represent ideal performance,
as participants usually solve multiple surveys per day and are aiming to finish studies as fast as possible.

We further investigate if the similarity score $s$ as in \cref{eq:app:explore:compare:concept_comparison} is useful to detect if participants have difficulty with finding the correct outlier (accuracy below 0.65).
Compared to using the perceived similarity and confidence values,
the similarity score $s$ is not as indicative. 
We hypothesize that humans will not only have problems when $s$ is high, but also when the similarity between the embeddings of the correct pair is small. 

\subsubsection{Clarity Measure}
We in the following provide details \wrt study design, data filtering and results.

\begin{figure}[t]
    \centering
    \includegraphics[width=0.99\textwidth]{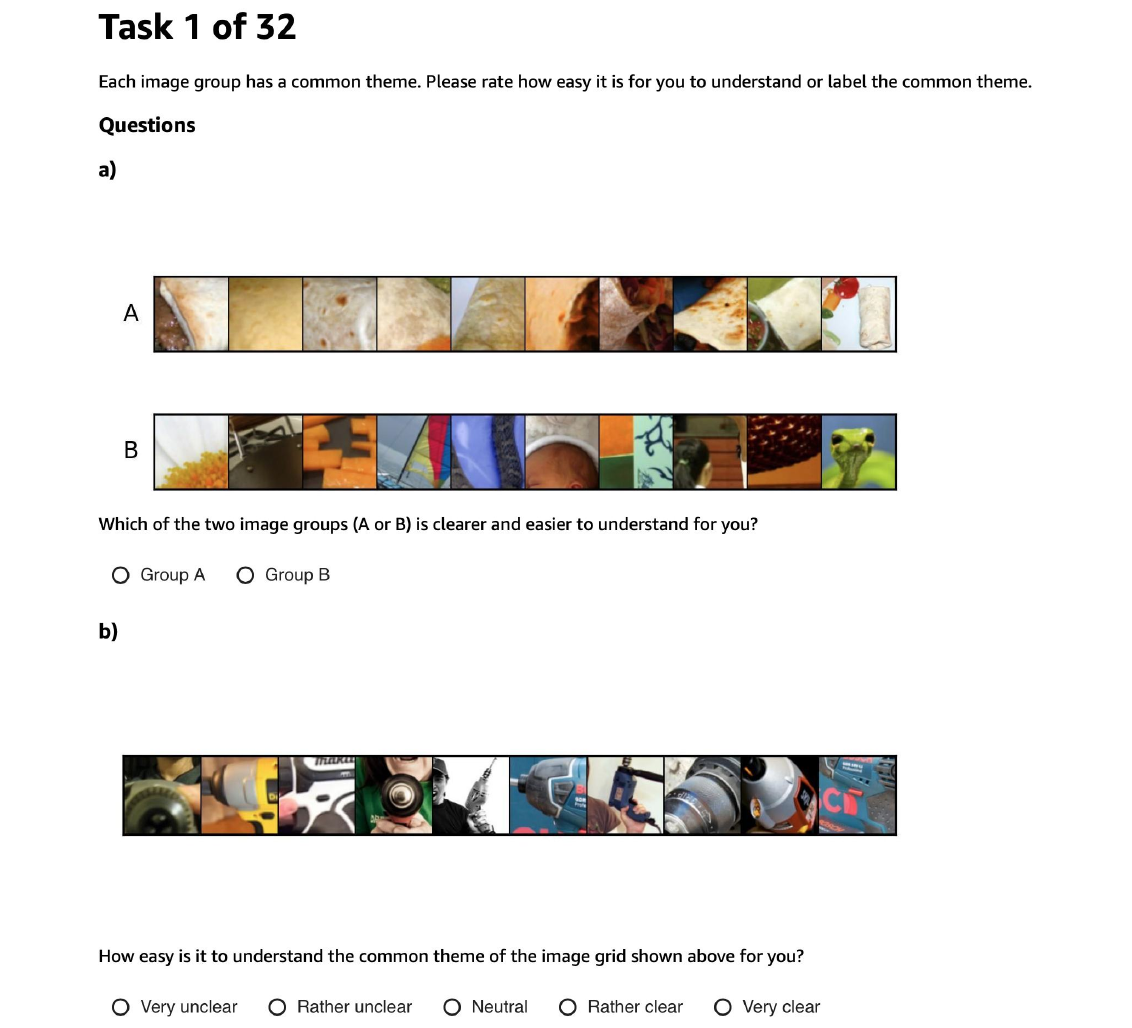}
    \caption{Study design to evaluate alignment of our clarity measure with human perception.
    We ask participants to choose the clearer one of two concept example sets (A and B),
    or for a qualitative clarity rating based on a Likert scale.
    }
    \label{fig:app:interpretability:study:clarity_study}
\end{figure}

\textbf{Study design:}
We evaluate concept clarity (as perceived by humans) via a combination of qualitative and quantitative tests.
Concretely, a study consists of 32 tasks,
where each task is divided into three questions,
as illustrated in \cref{fig:app:interpretability:study:clarity_study}.
For the first and last question,
participants are asked to choose the ``clearer'', easier to understand set of two sets of concept examples (top ten most activating image patches).
Here the first question differs from the last in the fact that one set of concept examples is sampled randomly from all neurons of a model, \eg, each concept examples corresponds to a different neuron.
As such, the first question represents a controlled setting,
where one concept group is artificially made unclear and difficult to understand.
For the second question,
we ask participants to rate the clarity of a single concept example set based on a Likert scale.

All in all,
we create six experiments with 32 tasks each for the ResNet50v2 (four times), ResNet50 and ResNet50.a1, using neurons of the penultimate layer.

\textbf{Compensation:}
The compensation for participation is divided into two parts: a base payment, and a bonus. 
Firstly, participation will receive a base payment of $2.0$ US dollars (estimated time per study are 15 minutes) if their performance was above 66\,\% (represents statistically significantly better than random performance according to a binomial test with 32 draws and success probability of 50\,\%). 
Performance is measured by the number of correctly recognized image grids with a clearer common theme. 
The participants are further warned that responses
corresponding to random guessing are rejected, leading to no payout. 
In addition, participants are motivated to perform well by
 giving a bonus of 1.0 US dollar when performance was above 92\,\%.

\textbf{Data filtering:}
We remove study results where the overall detection accuracy is below 70\% (for question one),
which corresponds to a statistically significant result above random performance ($p<0.03$) according to a binomial test.
Overall, we receive 74 of 90 experimental results after filtering.

\textbf{Results:}
\begin{figure}[t]
    \centering
    \includegraphics[width=0.6\textwidth]{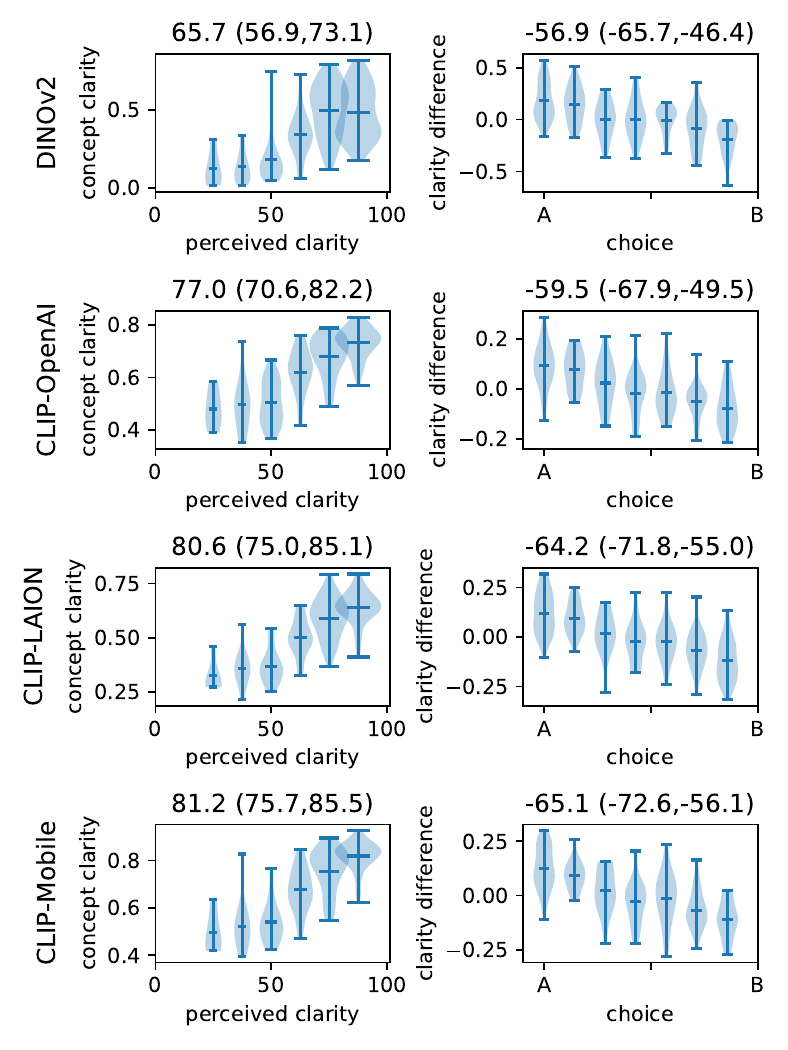}
    \caption{Correlation analysis for the user study regarding ``clarity'', where we ask for a qualitative rating (\emph{left}) or for the clearer concept example set of two choices A and B (\emph{right}). 
    Overall,
    we see high correlations between our concept clarity measure and perceived clarity as well as the choice (when computing the clarity difference between A and B (correlation scores given above each violin plot with 95\,\% confidence intervals in parentheses).
    }
    \label{fig:app:interpretability:study:clarity_correlation}
\end{figure}
For question two,
we receive two clarity scores,
one for each concept set $\mathcal{E}$.
Denoting $I_{\texttt{clarity}}^\text{a}$ and $I_{\texttt{clarity}}^\text{b}$ as the clarity of concept examples according to \cref{eq:methods:efficient-clarity}, respectively,
we receive the difference in clarity scores as
\begin{equation}
    \Delta~I_{\texttt{clarity}} = I_{\texttt{clarity}}^\text{a} - I_{\texttt{clarity}}^\text{b}~.
\end{equation}

In \cref{fig:app:interpretability:study:clarity_correlation},
the clarity of concept example sets (as measured via foundation models) is shown against perceived clarity (answer of second question) (\emph{left}),
and the difference of clarity scores $\Delta~I_{\texttt{clarity}}$ against the average choice (\emph{right}).
For each case,
Pearson correlation scores are given with a corresponding 95\,\% confidence interval.

\subsubsection{Polysemanticity Measure}
We in the following provide details \wrt study design, data filtering and results.

\textbf{Study design:}
\begin{figure}[t]
    \centering
    \includegraphics[width=0.99\textwidth]{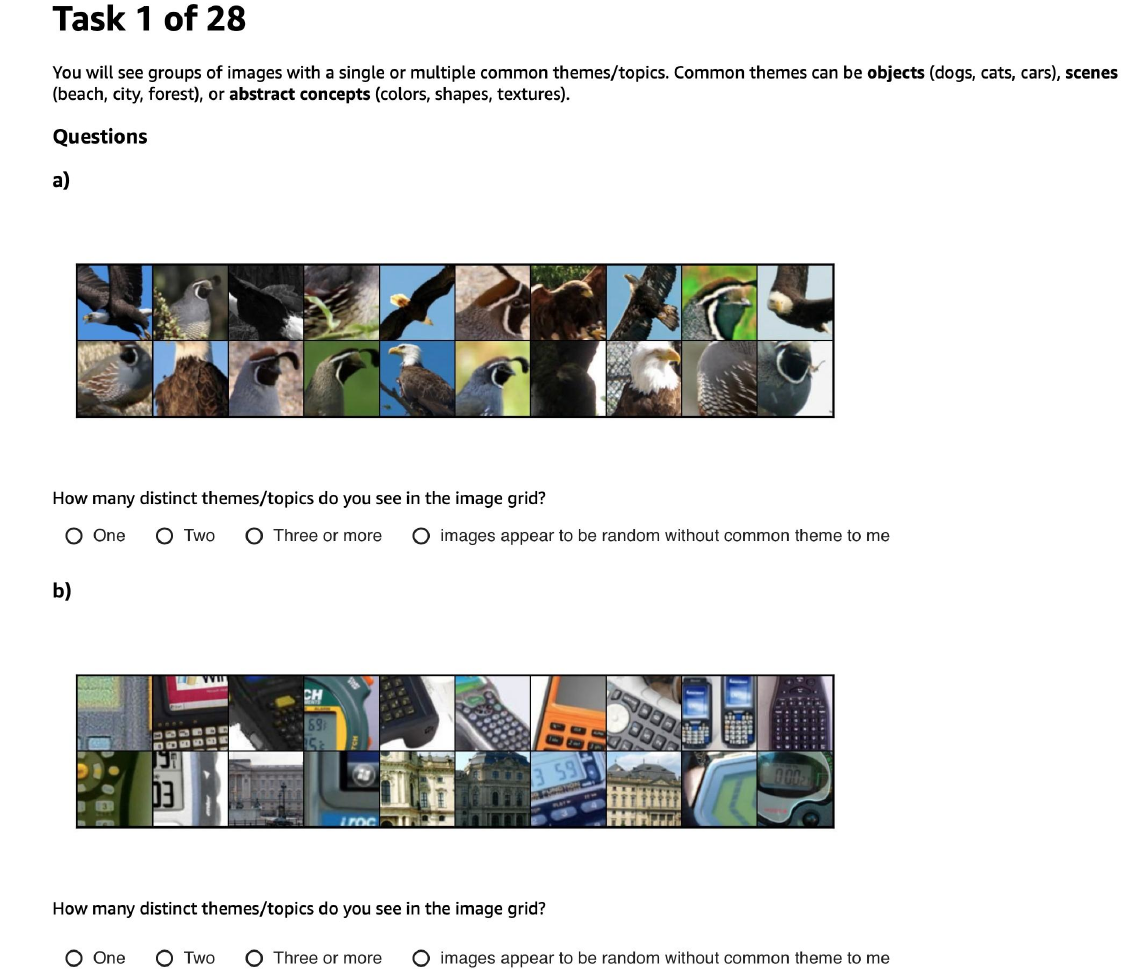}
    \caption{Study design to evaluate alignment of our polysemanticity measure with human perception.
    A study consists of 28 tasks,
    where each task is divided into two questions.
    For individual concept example sets (top 20 most activating image patches),
    we ask participants whether they see ``one'', ``two'', ``three or more'', or ``no common theme at all'' in the concept examples.
    }
    \label{fig:app:interpretability:study:poly_study}
\end{figure}
We evaluate neuron polysemanticity (as perceived by humans) via a combination of qualitative and quantitative tests.
Concretely, a study consists of 28 tasks,
where each task is divided into two questions,
as illustrated in \cref{fig:app:interpretability:study:poly_study}.
For both questions,  we show participants a single concept example set (top 20 most activating image patches),
and ask whether they see ``one'', ``two'', ``three or more'', or ``no common theme at all'' in the concept examples.
A choice of ``one'' is rated as monosemantic according to a human (polysemanticity $p=0$),
whereas the options ``two'' and ``three or more'' are rated as polysemantic ($p=1$).
Here ``no common theme at all'' is valued as neither, receiving a score in-between ($p=0.5$).

Importantly,
for question one, either reference samples of a \emph{single} (rather) monosemantic neuron is shown (corresponding to neurons of a model where clarity is maximal),
or, alternatively (chosen randomly with probability 0.5), of \emph{two} (rather) monosemantic neurons (where reference images are mixed).
As such corresponds question one to a controlled setting,
which is used to filter data.

All in all,
we create six experiments with 28 tasks each for the ResNet50v2 (two sets), ResNet101v2 (two sets) and ResNet50.a1 (two sets), using neurons of the penultimate layer.

\textbf{Compensation:}
The compensation for participation is divided into two parts: a base payment, and a bonus. 
Firstly, participation will receive a base payment of $2.0$ US dollars (estimated time per study are 15 minutes) if their performance was above 39\,\%  (represents statistically significantly better than random performance according to a binomial test with 28 draws and success probability of 25\,\%). 
Performance is measured by the number of correctly recognized outliers in the odd-one-out task. 
The participants are further warned that responses
corresponding to random guessing are rejected, leading to no payout. 
In addition, participants are motivated to perform well by giving a bonus of 1.0 US dollar when performance was above 92\,\%.

\textbf{Data filtering:}
We remove study results where the overall detection accuracy (according to the first question) is below 72\%,
which corresponds to a statistically significant result above random performance ($p<0.001$) according to a binomial test.
Overall, we receive 85 of 108 experimental results (at least 11 results per study) after filtering.

\textbf{Results:}
\begin{figure}[t]
    \centering
    \includegraphics[width=0.99\textwidth]{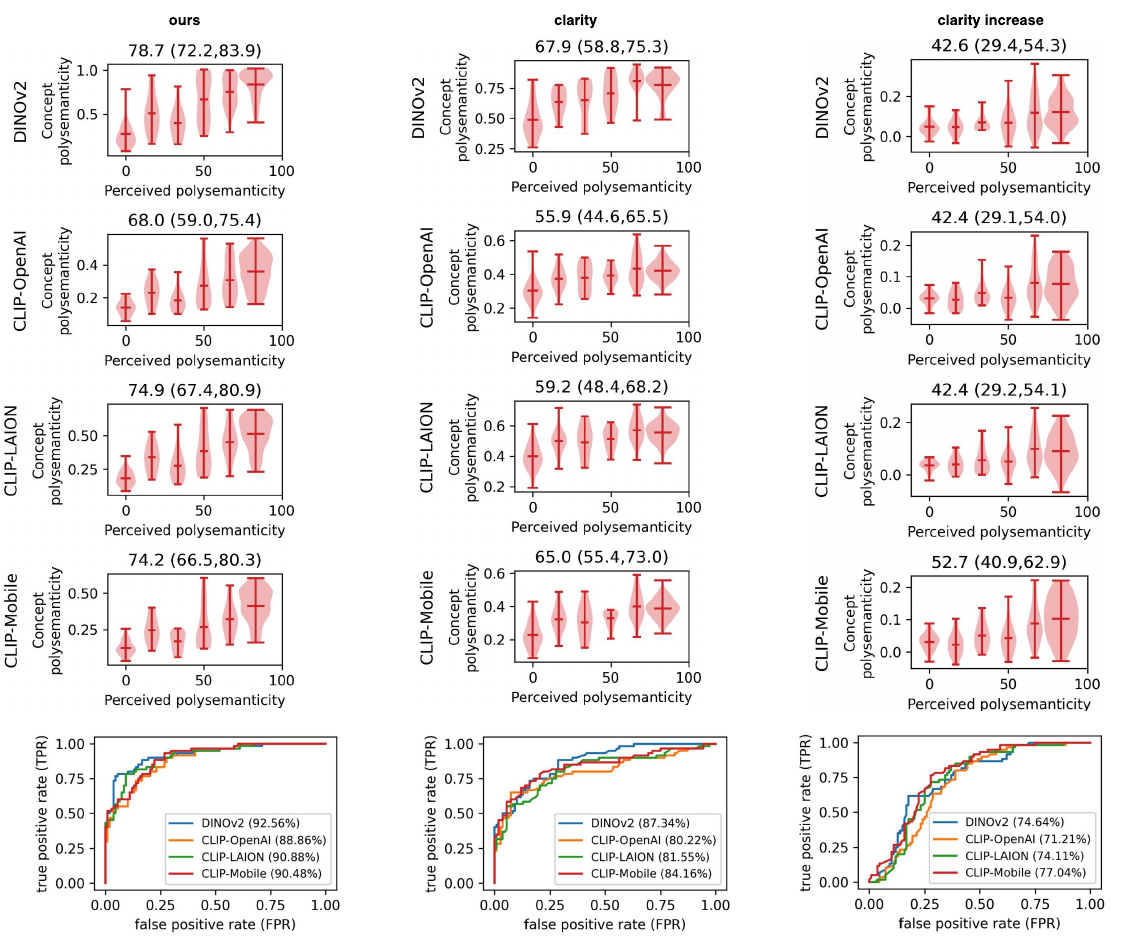}
    \caption{Ablation study via correlation analysis for the user study 
     regarding ``polysemanticity'', where participants are asked about their qualitative opinion \wrt polysemanticity in shown concept example sets. 
    In all violin plots,
    correlation between perceived polysemanticity and our concept polysemanticity measures are depicted (displayed above each violin plot with confidence intervals in parentheses).
    Here,
    we use on the one hand side either our measure (\emph{left}), the clarity measure (\emph{middle}), or the clarity increase (after clustering of concept examples inside one set).
    For all cases,
    we further computed AUC scores for detecting polysemantic neurons (according to perceived polysemanticity $p>0.5$.
    }
    \label{fig:app:interpretability:study:poly_ablation}
\end{figure}
In \cref{fig:app:interpretability:study:similarity_correlation},
the polysemanticity of concept example sets (as measured via foundation models) is shown against perceived polysemanticity (answer of second question) (\emph{left}).
We further provide a correlation analysis for using the clarity measure instead or compute the increase in concept clarity after clustering, as motivated by~\cite{dreyer2024pure}.
For each case,
Pearson correlation scores are given with a corresponding 95\,\% confidence interval.
It is apparent that our polysemanticity measure as introduced in \cref{sec:methods:interpretability} performs generally significantly better than using overall clarity or the increase in clarity as a measure.

Alternatively,
we aim to detect polysemantic units (according to the participants) where the average perceived polysemanticity is smaller than 50\,\%.
The corresponding AUC scores when using the three different measures to detect the polysemantic units are shown in \cref{fig:app:interpretability:study:similarity_correlation} (\emph{bottom}).
The results are related to the correlation analysis,
indicating that our polysemanticity measure is most indicative for detecting polysemantic units.

\subsection{Optimizing Interpretability}
\label{app:interpretability:training}

This section provides more details and additional experiments to \cref{sec:results:evaluate:optimize}.
The interpretability measures of clarity, polysemanticity and redundancy allow to investigate the effect of various hyperparameter choices \wrt model architecture or training on the latent interpretability.
We in the following present results for the training hyperparameters of:
dropout regularization, sparsity regularization, task complexity, data augmentations and number of epochs.

We therefore train a ResNet50, ResNet18, VGG-13 and VGG-13 (BN) on a subset of ImageNet for 100 epochs with an initial learning rate of $0.01$ which is decreased by a factor of 10 after 80 epochs.
The subset consists of 13 animal classes ('Staffordshire bullterrier, Staffordshire bull terrier', 'goldfish, Carassius auratus', 'hen', 'bulbul', 'box turtle, box tortoise', 'water snake', 'bee eater', 'jellyfish', 'hermit crab', 'flamingo', 'pelican', 'beagle', 'soft-coated wheaten terrier', 'Rottweiler', 'Siberian husky', 'tabby, tabby cat', 'brown bear, bruin, Ursus arctos', 'mantis, mantid', 'hamster') and 13 object classes ('castle', 'canoe', 'crane', 'digital clock', 'dome', 'harp', 'lawn mower, mower', 'moped', 'monastery', 'palace', 'pole', 'snorkel', 'snowmobile', 'wok', 'comic book', 'bubble', 'acorn', 'toilet tissue, toilet paper, bathroom tissue').

The semantic embeddings are computed using DINOv2 class token feature vectors of the 50 most activating samples as concept examples for each neuron in the penultimate layer.
We further filter out dead neurons by thresholding the maximal relevance on the test set to be above 0.005 (0.5\,\%).

\paragraph{Dropout regularization}
Dropout regularization randomly removes neurons from the computational graph by setting their activation to zero.
In our study,
we randomly set 40\,\% of neurons to zero.
The resulting interpretability scores are given in \cref{tab:app:interptretability:optimizing_drop_out}.
It is apparent that the ResNet models are not strongly effected by drop-out \wrt interpretability.
However,
the VGG models have overall a higher clarity and redundancy.
\begin{table}[t]
\centering
\caption{Studying the effect of hyperparameters on model interpretability (given by clarity, polysemanticity and redundancy scores) using \textbf{dropout} during training. All models are trained for 100 epochs.}
\label{tab:app:interptretability:optimizing_drop_out}
\begin{tabular}{cccccc}
\toprule
model name & dropout & clarity (\%)& polysemanticity (\%)& redundancy (\%)\\
\midrule
ResNet34 & False & 45.0 & 91.9 & 84.3 \\
ResNet34 & True & 44.0 & 92.3 & 83.4 \\
ResNet50 & False & 47.1 & 89.6 & 88.9 \\
ResNet50 & True & 47.9 & 89.9 & 89.6 \\
VGG-13 & False & 41.8 & 83.3 & 81.6 \\
VGG-13 & True & 47.1 & 83.7 & 85.1 \\
VGG-13 (BN) & False & 35.5 & 84.7 & 79.1 \\
VGG-13 (BN) & True & 39.6 & 78.4 & 81.6 \\
\bottomrule
\end{tabular}
\end{table}

\begin{figure}[t]
    \centering
    \includegraphics[width=0.99\textwidth]{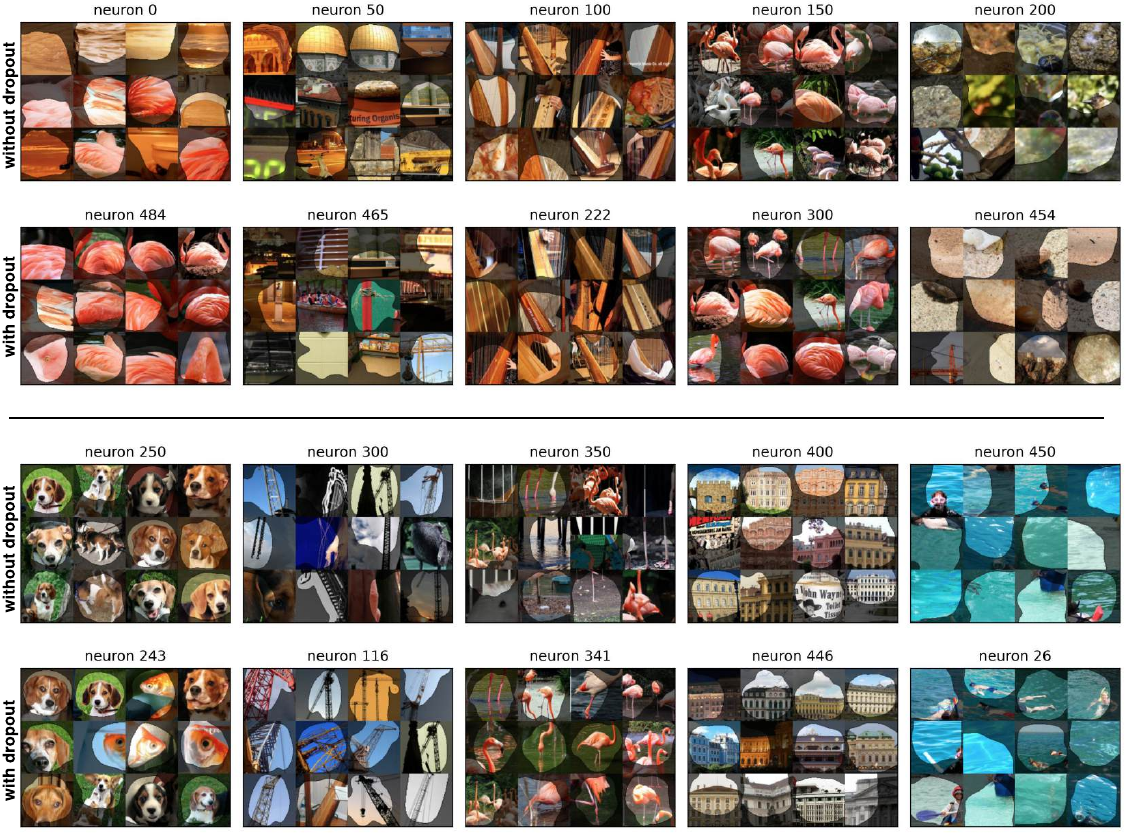}
    \caption{Qualitative comparison of concepts in the penultimate layer of a VGG-13 model when dropout is either applied or not applied.
    For each concept of a neuron in the VGG without dropout,
    we retrieve the most similar concept of the VGG with dropout based on highest cosine similarity values of semantic embeddings.
    }
    \label{fig:app:interptretability:optimizing_drop_out}
\end{figure}
In \cref{fig:app:interptretability:optimizing_drop_out} we present examples of how concepts change when dropout is applied.
Specifically,
we show the concepts of neurons \texttt{\#0}, \texttt{\#50}, \texttt{\#100}, \texttt{\#150}, \texttt{\#200}, \texttt{\#250}, \texttt{\#300}, \texttt{\#350}, \texttt{\#400}, \texttt{\#450} and \texttt{\#500} of the VGG-13 model without dropout,
and the most similar concept of the VGG with dropout based on highest cosine similarity values of semantic embeddings.
Here,
it is observable,
that most neurons become more class-specific and clear,
as \eg, for \texttt{\#484} that only shows flamingo feather instead of \texttt{\#0} that encodes also for sunset besides feathers.

\paragraph{Sparsity Loss}
With the aim to reduce redundancies and polysemanticity,
we apply a sparsity loss $L_\text{sparse}$ on the activations $Z\in\mathbb{R}^k$ of the penultimate layer's features ($k$ neurons) during training.
Concretely, we apply the additional sparsity loss with strength $\lambda$ besides the standard cross entropy loss $L_\text{CE}$
\begin{equation}
    L = L_\text{CE} + \lambda L_\text{sparse}~,
\end{equation}
where the sparsity loss is given as the L1-norm over all (non-zero) activations:
\begin{equation}
    L_\text{sparse} = \frac{1}{k^+}\sum_{i=0}^k |Z_i|~,
\end{equation}
where $k^+$ is the number of non-zero activations.

The resulting interpretability scores are given in \cref{tab:app:interptretability:optimizing_sparsity}.
In almost all cases,
redundancy is decreased, clarity increased and polysemanticity reduced, leading to an overall improved interpretability.
\begin{table}[t]
\centering
\caption{Studying the effect of hyperparameters on model interpretability (given by clarity, polysemanticity and redundancy scores) using a \textbf{sparsity loss} with strength $\lambda$. All models are trained for 100 epochs.}
\label{tab:app:interptretability:optimizing_sparsity}
\begin{tabular}{ccccc}
\toprule
regularization ($\lambda$) & model name & clarity (\%)& polysemanticity (\%)& redundancy (\%)\\
\midrule
0 & ResNet34 & 45.0 & 91.5 & 84.4 \\
0 & ResNet50 & 47.7 & 89.3 & 89.1 \\
0 & VGG-13 & 41.8 & 83.3 & 81.6 \\
0 & VGG-13 (BN) & 35.6 & 84.8 & 79.1 \\
80 & ResNet34 & 60.2 & 67.0 & 44.8 \\
80 & ResNet50 & 56.4 & 87.7 & 70.7 \\
80 & VGG-13 & 52.1 & 83.2 & 52.9 \\
80 & VGG-13 (BN) & 48.0 & 71.8 & 82.8 \\
\bottomrule
\end{tabular}   
\end{table}

\paragraph{Task complexity}
Further,
we investigate the effect of low or higher task complexity on interpretability.
We would expect that very easy tasks lead to more redundancies and less specific neurons, which correspond to lower interpretability.
In order to test the effect of task complexity,
we train in two scenarios: a simple binary task (animal vs. objects) and more complex multi-class scenario (all 23 animal and object classes).
The resulting interpretability scores are given in \cref{tab:app:interptretability:optimizing_complexity}.
Here, overall, higher complexity seems to result in higher interpretability,
\eg, clarity increases, and redundancy decreases.
However,
polysemanticity is observed to increase.

\begin{table}[t]
\centering
\caption{Studying the effect of hyperparameters on model interpretability (given by clarity, polysemanticity and redundancy scores) varying the \textbf{task complexity} (binary: two classes, class-specific: 26 classes). All models are trained for 100 epochs.}
\label{tab:app:interptretability:optimizing_complexity}
\begin{tabular}{ccccc}
\toprule
model name & task & clarity (\%)& polysemanticity (\%)& redundancy (\%)\\
\midrule
ResNet34 & binary & 37.4 & 85.3 & 95.0 \\
ResNet34 & class-specific & 44.3 & 91.9 & 83.6 \\
ResNet50 & binary & 29.8 & 87.1 & 93.8 \\
ResNet50 & class-specific & 44.3 & 89.2 & 88.3 \\
VGG-13 & binary & 35.1 & 85.3 & 82.2 \\
VGG-13 & class-specific & 42.2 & 83.1 & 81.5 \\
VGG-13 (BN) & binary & 33.5 & 82.7 & 83.8 \\
VGG-13 (BN) & class-specific & 35.0 & 84.6 & 78.9 \\
\bottomrule
\end{tabular}
\end{table}

\paragraph{Data Augmentation}
The next training hyperparameter that we vary is data augmentation.
Here,
we apply three sets of data augmentation:
\begin{itemize}
    \item \textbf{none}
    \item \textbf{default}: random crop to $224\times224$ pixels, random horizontal flip (probability of 50\,\%)
    \item \textbf{strong}: random rotation of up to 10 degrees, random crop to $224\times224$ pixels, random horizontal flip (probability of 50\,\%), random sharpness decrease by 0.2 (probability of 50\,\%), random sharpness increase by 0.2 (probability of 50\,\%).
\end{itemize}

The resulting interpretability scores are given in \cref{tab:app:interptretability:optimizing_augmentation}.
Although there are changes in interpretability such as the VGG-13 for which clarity increases,
there are usually no clear trends in interpretability.
\begin{table}[t]
\centering
\caption{Studying the effect of hyperparameters on model interpretability (given by clarity, polysemanticity and redundancy scores) varying the \textbf{data augmentations} 
(
Rot: random rotation of up to 10 degrees,
Crop: random crop to 224 pixels height and width,
xFlip: random horizontal flip,
Sharpness: random sharpness variations). 
All models are trained for 100 epochs.}
\label{tab:app:interptretability:optimizing_augmentation}
\begin{tabular}{cccccc}
\toprule
model name & augmentation & clarity & polysemanticity & redundancy \\
\midrule
ResNet34 & Rot,Crop,xFlip,Sharpness & 44.5 & 91.5 & 83.8 \\
ResNet34 & Crop,xFlip & 45.0 & 91.4 & 84.3 \\
ResNet34 & none & 45.3 & 89.7 & 86.1 \\
ResNet50 & Rot,Crop,xFlip,Sharpness & 46.7 & 88.9 & 90.3 \\
ResNet50 & Crop,xFlip & 44.9 & 90.3 & 88.8 \\
ResNet50 & none & 46.9 & 85.5 & 89.7 \\
VGG-13 & Rot,Crop,xFlip,Sharpness & 42.1 & 83.7 & 81.6 \\
VGG-13 & Crop,xFlip & 41.8 & 83.0 & 81.8 \\
VGG-13 & none & 34.3 & 84.6 & 79.3 \\
VGG-13 (BN) & Rot,Crop,xFlip,Sharpness & 35.2 & 85.0 & 78.9 \\
VGG-13 (BN) & Crop,xFlip & 35.7 & 84.9 & 79.3 \\
VGG-13 (BN) & none & 35.6 & 84.2 & 80.8 \\
\bottomrule
\end{tabular}
\end{table}

\paragraph{Number of Epochs}
In \cref{sec:results:evaluate} we have observed that more extensively trained ImageNet models such as the ResNet50v2 result in higher interpretability compared to the ResNet50 for the last feature layers neurons.
Thus, we lastly investigate
the effect of the number of epochs on latent interpretability.
Concretely,
we train models for 400 epochs on the ImageNet subset, 
and decrease learning rate by 10 after 300 and 370 epochs, respectively.
The resulting interpretability scores are given in \cref{tab:app:interptretability:optimizing_numer_of_epochs}.
It is apparent,
that interpretability changes, especially for the first 100 epochs.
Between 100 and 400 epochs, large changes are not visible any more.
Generally,
the longer the training, the less redundancies are observed.
Further,
clarity is reduced, indicating the formation of more complex concepts,
which is contrary to the observation in \cref{sec:results:evaluate},
where better trained models show higher clarity.
This example shows that many effects can take place and have an effect on interpretability.
It might be that for the subset of ImageNet,
models are more likely to learn the data (overfit) compared to the whole ImageNet dataset,
where models are more likely to generalize.

\renewcommand{\arraystretch}{1.3}
\begin{table}[t]
\centering
\caption{Studying the effect of hyperparameters on model interpretability (given by clarity, polysemanticity and redundancy scores) varying the \textbf{number of epochs}.}
\label{tab:app:interptretability:optimizing_numer_of_epochs}
\begin{tabular}{cccccc}
\toprule
epoch & model name & clarity & polysemanticity & redundancy \\
\midrule
39 & ResNet34 & 45.7 & 91.5 & 85.3 \\
79 & ResNet34 & 44.8 & 91.9 & 84.3 \\
119 & ResNet34 & 44.7 & 91.8 & 84.0 \\
159 & ResNet34 & 44.0 & 92.1 & 83.5 \\
199 & ResNet34 & 43.8 & 92.2 & 83.3 \\
239 & ResNet34 & 44.1 & 92.1 & 83.1 \\
279 & ResNet34 & 43.5 & 92.3 & 83.0 \\
319 & ResNet34 & 43.4 & 92.2 & 82.5 \\
359 & ResNet34 & 43.3 & 92.2 & 82.4 \\
399 & ResNet34 & 43.3 & 92.2 & 82.4 \\
39 & ResNet50 & 51.2 & 88.6 & 91.6 \\
79 & ResNet50 & 43.5 & 90.1 & 88.0 \\
119 & ResNet50 & 47.3 & 88.3 & 88.8 \\
159 & ResNet50 & 46.8 & 89.2 & 88.5 \\
199 & ResNet50 & 46.6 & 85.9 & 88.4 \\
39 & VGG-13 & 42.2 & 83.1 & 82.8 \\
79 & VGG-13 & 41.8 & 83.3 & 81.6 \\
119 & VGG-13 & 41.9 & 83.1 & 81.3 \\
159 & VGG-13 & 42.2 & 82.9 & 81.4 \\
199 & VGG-13 & 42.2 & 83.2 & 81.2 \\
239 & VGG-13 & 42.1 & 82.9 & 81.2 \\
279 & VGG-13 & 42.0 & 83.1 & 81.1 \\
319 & VGG-13 & 42.6 & 82.6 & 81.2 \\
359 & VGG-13 & 42.6 & 82.7 & 81.2 \\
399 & VGG-13 & 42.7 & 82.9 & 81.3 \\
39 & VGG-13 (BN) & 36.4 & 84.1 & 82.0 \\
79 & VGG-13 (BN) & 35.2 & 84.5 & 79.5 \\
119 & VGG-13 (BN) & 34.3 & 84.7 & 78.2 \\
159 & VGG-13 (BN) & 34.2 & 84.5 & 77.7 \\
199 & VGG-13 (BN) & 34.0 & 84.8 & 77.4 \\
239 & VGG-13 (BN) & 33.7 & 84.8 & 76.9 \\
279 & VGG-13 (BN) & 33.8 & 84.8 & 77.1 \\
319 & VGG-13 (BN) & 33.9 & 84.8 & 76.9 \\
359 & VGG-13 (BN) & 33.8 & 85.0 & 76.8 \\
399 & VGG-13 (BN) & 33.8 & 85.0 & 76.7 \\
\bottomrule
\end{tabular}
\end{table}
\renewcommand{\arraystretch}{1.5}

\setcounter{figure}{0}
\setcounter{table}{0}\setcounter{equation}{0}

\section{Method}
\label{app:methods}
\begin{figure}[t]
    \centering
    \includegraphics[width=0.99\textwidth]{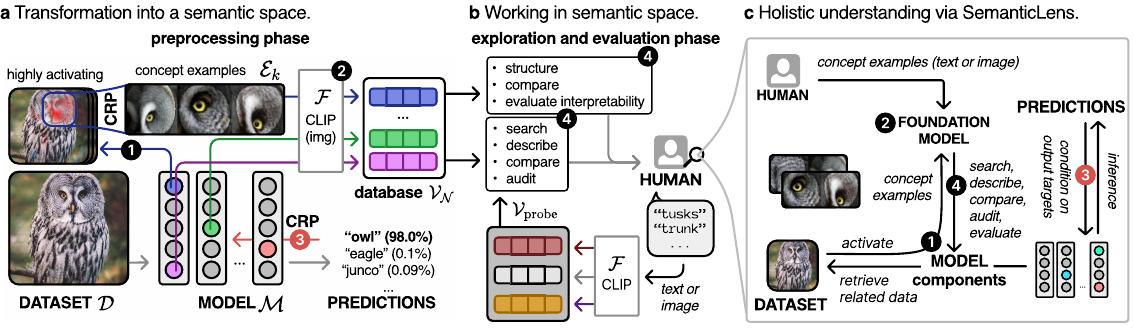}
    \caption{Methodological overview over \gls{ours}.
    \textbf{a})
    In an automated preprocessing step,
    \protect\circledsmall{1} concept examples $\mathcal{E}_k$ for each model component $k$ are computed
    by cropping highly activating samples to the relevant part using \gls{crp}~\cite{achtibat2023attribution}.
    Further,
    \protect\circledsmall{2} a multimodal foundation model such as CLIP is used to embed the concept examples in a semantic representation \gls{cone}$_k\in\mathcal{V}_\mathcal{M}$.
    Optionally,
    \protect\circledrsmall{3} \gls{crp} allows to compute relevances of components \wrt specific output targets.
    \textbf{b})
    \protect\circledsmall{4}
    The preprocessing phase results in a vector database,
    which can be utilized to compare models or evaluate human-interpretability.
    Alternatively,
    a user might search, describe and audit the model by defining a set of concepts via text or image data that is also embedded by the foundation model.
    \textbf{c})
    Through the foundation model, a user can search representations (\protect\circledsmall{4}), filter them according to output relevances (\protect\circledrsmall{3}), and find related (highly activating) data samples.
    }
    \label{fig:app:methods:methods_in_brief}
\end{figure}

We in the following provide more details regarding our methodology and its hyperparameters.
Hereby,
\cref{fig:app:methods:methods_in_brief} illustrates the methodology which is discussed in \cref{sec:methods}.

\subsection{Transformation into Semantic Space}
In order to embed model components in the semantic space of a foundation model,
we first collect examples images for a components concept,
and pool the feature vectors of the foundation model when operating on the examples.

\paragraph{Concept Examples}
The concept examples are collected by searching for the most activating samples of the test dataset.
Here,
we perform average pooling over spatial dimensions, as, \eg, given for transformer blocks or convolutional layers.
Alternatively, max-pooling could be applied, which we also experiment with (and evaluate) in \cref{sec:app:explore:describe:evaluation}.

Throughout the paper,
we show concept examples that are cropped \emph{and} masked. 
However,
we refrain from masking concept examples when encoded by a foundation model due to potentially being out of the data distribution.

In order to generate cropped samples,
we use the zennit-crp framework~\cite{achtibat2023zennit},
which computes a neuron-specific heatmap (corresponding to a heatmap for the largest spatial activation value for convolutional channels),
which is further smoothed with a Gaussian blur (kernel size 51).
Subsequently,
the images are cropped to include all values where the heatmap is higher than 1,0\,\% of the largest attribution value.
The black masks (semi-transparent with 0.4 opacity) are corresponding to all values smaller than 2,0\,\% of the maximal attribution value.

Notably,
\gls{crp} is not applicable to \gls{vit} yet.
Thus,
we use the upsampled spatial transformer tokens as heatmaps to localize concepts, as, \eg, inspired by \cite{bau2017network}.

\paragraph{Pooling Semantic Feature Vectors}
The pooling operation we perform on semantic vectors is a straightforward averaging of latent feature vectors (corresponding to the class token) of \gls{vit}-based foundation models.
For future work,
we leave to compare different pooling operations, such as median or a (activation)-weighted average.

\paragraph{Templates}
Our default templates are ``\texttt{<concept>}'', ``\texttt{<concept>}-like'', ``\texttt{a <concept>}'' and ``an image of a close up of \texttt{<concept>}''
which have been shown to lead to more faithful labels in \cref{sec:app:explore:describe:evaluation}.

\subsection{Concept Relevance Scores}
\label{sec:app:methods:concept_relevances}
We use \gls{crp}~\cite{achtibat2023attribution} to compute latent importance scores of neurons and neuron-specific heatmaps via the zennit-crp package~\cite{achtibat2023zennit}.
\gls{crp} is hereby based on the feature attribution method of \gls{lrp}~\cite{bach2015pixel} that works by back-propagating relevance scores from the output back to the input through the model and its latent components.
We use the \gls{lrp} composite used in the work of \gls{crp}, namely the $\varepsilon z^+ \flat$-rule.

\subsection{Human-Interpretability of Concepts}
\label{app:methods:interpretability}
The compact form of our \texttt{clarity} score, presented in \cref{eq:methods:efficient-clarity}, is derived directly from the following equality:
\begin{equation}
    \Big\|\sum_{i=1}^n v_i\Big\|_2^2  = \Big\langle{\sum_{i=1}^nv_i,\sum_{j=1}^nv_j}\Big\rangle =  \sum_{i,j=1}^n\skp{v_i,v_j}
\end{equation}
which holds for any selection of vectors $v_1,...,v_n{\in\mathbb{R}^d}$. For $v_1,{...,}v_n\in S^{d-1}$ of unit length, we obtain
\begin{align}
    \sum_{i=1}^n \sum_{i\neq j} \skp{v_i,v_j} 
    &= 
    \sum_{i=1}^n \Big(\sum_{j=1}^n \skp{v_i,v_j}-1\Big)
    =
    \big\|
    \textstyle\sum_{i=1}^n v_i
    \big\|_2^2
    -n.
\end{align}
Applying this result to \cref{eq:methods:efficient-clarity:line1} yields the desired compact formula.

\subsection{Workflows for SemanticLens}
\label{sec:app:methods:workflow}
This section details possible workflows (and the involved steps) for answering the questions presented in \cref{tab:questions} of the main manuscript.
\begin{table}[th]
    \caption{Overview of questions which can be answered by \gls{ours} and workflow steps detailed in \cref{sec:app:methods:workflow}.}
    \centering 
    \begin{tabular}{l|p{8.5cm}|p{2cm}}
        Type & Question to the model $\mathcal{M}$ & Steps\\
        \toprule
        \multirow{3}{*}{search} & \textit{``Has my model learned to encode a specific concept?''} via convenient  ``search-engine''-type text or image descriptions & \circledw{1}, \circledw{2} \\
         & \textit{``Which components have encoded a concept, how is it used, and which data is responsible?''}  & 
        \circledw{1}, \circledw{2}, \circledw{4} \\
        \hline
        \multirow{5}{*}{describe} & \textit{``What concepts has my model learned?''} in a structured, condensed and understandable manner via textual descriptions & \circledw{1}, \circledw{3}\\
         & \textit{``What and how are concepts contributing to a decision?''} by visualizing concept interactions throughout the model & \circledw{1}, \circledw{3}, \circledw{4}\\
         & \textit{``What do I \emph{not} yet understand of my model?''}, offering to understand the unexpected concepts and their role for the model and origin in data & \circledw{1}, \circledw{3}, \circledw{4}, \circledw{5}\\
        \hline
        \multirow{4}{*}{compare} & \textit{``What has one model learned but not the other?''} by comparing learned concepts qualitatively and quantitatively & \circledw{1}, (\circledw{3}), \circledw{6}\\
         & \textit{``How do my model's concepts change when changing the architecture or training?''} by comparing and tracking semantics of components & \circledw{1}, (\circledw{3}), \circledw{6}\\
        \hline
        \multirow{2}{*}{audit} & \textit{``Is my model relying on valid information only?''} by separating learned concepts into \emph{valid}, \emph{spurious} or \emph{unexpected} knowledge & \circledw{1}, \circledw{3}, \circledw{4}, \circledw{5}\\
        \hline
        \multirow{3}{*}{evaluate} & \textit{``How interpretable is my model?''} with easy to compute measures & \circledw{1}, \circledw{7}\\
         & \textit{``How can I improve interpretability of my model?''} by evaluating interpretability measures when changing model architecture or training procedure & \circledw{1}, \circledw{7}\\
        \bottomrule
    \end{tabular}
    \label{app:tab:questions_workflow}
\end{table}
An overview over questions and involved steps is given in \cref{app:tab:questions_workflow}.

\textbf{\circledw{1} Semantic embedding of model components:}
The first step consists in the embedding of model components into the semantic representation,
as detailed in \cref{sec:methods:examples,sec:methods:transformation}.
Here,
for each component of model $\mathcal{M}$, concept examples $\mathcal{E}$ are collected,
that are subsequently transformed into a single semantic embedding \gls{cone} through a multimodal foundation model $\mathcal{F}$.

\textbf{\circledw{2} Search using a single probing embedding:}
As outlined in \cref{sec:methods:search},
to search the semantic embeddings \gls{cone} of a model $\mathcal{M}$,
we require a set of concept examples $\mathcal{E}$ for the concept of interest.
These concept examples can be of the same data domain as the model $\mathcal{M}$,
or another (\eg, text for CLIP).
As in \cref{sec:methods:examples,sec:methods:transformation},
the concept examples are transformed into a semantic probing embedding \gls{pcone} using the foundation model.
With this probing embedding,
we can retrieve via cosine similarity the most aligned semantic embeddings of the model.

\textbf{\circledw{3} Annotate using a set of probing embeddings:}
As outlined in \cref{sec:methods:search},
to label the semantic embeddings \gls{cone} of a model $\mathcal{M}$,
we require a set of expected concepts.
The expected concept set can further contain parent categories (which allows further grouping of concepts),
or a distinction between valid and spurious features (useful for auditing).

For each concept,
we further define a set of concept examples $\mathcal{E}$.
These concept examples can be of the same data domain as the model $\mathcal{M}$,
or another (\eg, text for CLIP).
As in \cref{sec:methods:examples,sec:methods:transformation},
the concept example sets are transformed into semantic probing embeddings \gls{pcone} using the foundation model.
For each model component,
we can now compute the most aligned expected concept (label).

\textbf{\circledw{4} Compute concept relevance scores:}
As detailed in \cref{sec:app:methods:concept_relevances},
\gls{crp} allows to compute relevance scores for individual neurons or neuron groups on the test set efficiently.
These relevance scores can be used individually (for each sample),
or an average or maximum value can be computed to understand global importance, \eg, the relevance of a neuron (group)/concept for the prediction of a specific output class.
Whereas individual relevance scores can be used to filter out data examples where a concept is present (and relevant),
a global relevance score allows, \eg, to filter out all irrelevant components.

Alternatively to computing relevance scores \wrt the prediction output,
one can also compute relevance scores of components for the activation of an upper-level layer component,
which ultimately allows to compute an attribution graph that visualizes the relevance flow (hierarchical dependencies of neurons) throughout the network, as also depicted in \cref{fig:results:describe}c.

\textbf{\circledw{5} Audit alignment:}
\cref{sec:methods:audit}
Having defined a set of expected concepts (see \circledw{3}),
as well as filtered out all overall irrelevant or class-irrelevant components (see \circledw{4}),
we can begin with inspecting the alignment of a model with our expectation, as also detailed in \cref{sec:methods:audit}.

In a first step,
one can study the relevant components (relevances from \circledw{4}),
that do not align with any expected concept.
These components can correspond to unexpected valid or spurious concepts.
In order to understand the components better,
it is useful to retrieve the output classes or samples where the component is relevant.

Secondly,
the aligned components and their concepts can be studied more closely.
For example,
with parent categories for concepts are available (\eg, spurious or valid),
one can quantify the number of components for each group, or the total relevance of components per group.

\textbf{\circledw{6} Compare embeddings:}
The learned knowledge of model can be compared by measuring similarities, as detailed in \cref{sec:methods:search}.
Here either pairs of neurons can be compared via cosine similarity,
or groups of neurons, \eg, via average maximum similarity.

Furthermore,
two models can be compared with labels available from \circledw{3},
where in an additional optional step \circledw{4} we can filter components to be relevant for specific classes.
In the manner of Network Dissection, \eg, one can compare the amount of neurons that were assigned to a specific concept.

\textbf{\circledw{7} Evaluate Interpretability:}
In order to evaluate interpretability,
we proposed three measures in \cref{sec:methods:interpretability} corresponding to ``clarity'', ``polysemanticity'' and redundancy.
In order to evaluate whole models, only step \circledw{1} is required and the measures in \cref{sec:methods:interpretability} computed.
As such, they provide the means to also change hyperparameters of the model during training or hyperparameters of the training procedure and investigate the effect on interpretability.

\setcounter{figure}{0}
\setcounter{table}{0}\setcounter{equation}{0}
\section{Limitations and Future Work}
\label{app:limitations}
One limitation of \gls{ours} lies in its reliance on the expertise of the employed foundation model of the considered data domain. If the foundation model was not sufficiently trained on the specific data domain, or itself relies on spurious correlation and biases~\cite{tanjim2024discovering}, \gls{ours}'s ability to correctly/faithfully analyse the network's knowledge is affected. 
Therefore the foundation model employed by \gls{ours} needs to be carefully selected with respect to the studied data domain, as done with WhyLesionCLIP in \cref{sec:results:medical}. However, in very niche or specific data domains, no foundation model might be available. 
In such cases, 
the investigated model itself could be used as an alternative, although this may not provide an optimal semantic space~\cite{park2023concept}.

Secondly,
if the components of the model are not describable via concept examples,
or when concept examples are not meaningful,
an investigation with \gls{ours} is not effective.
A potential solution lies here with post-hoc architecture changes such as \gls{sae}~\cite{huben2023sparse} or activation factorization techniques~\cite{fel2024holistic} that lead to more interpretable components.
Alternatively, instead of post-hoc adaptations, interpretability can be integrated into the hyperparameter selection process. Our proposed novel interpretability scores can be a helpful tool for this, as demonstrated with the examples of dropout and activation-sparsity regularization in \cref{sec:results:evaluate,app:interpretability:training}. These scores facilitate the study of the effects of hyperparameter choices and post-hoc model augmentations on the interpretability of the final model, representing a promising direction for future research.

Whereas we provide three latent interpretability measures,
there is still potential for measures that describe other aspects of concept-based explanations.
For example,
it might be easier to understand concept use for singular instances if they are spatially localized.
Further,
it will be easier to understand explanations,
if as few as possible distinct concepts are used by model.

In this work,
we demonstrate \gls{ours} on image data.
However,
we \gls{ours} can be applied to any data domain where foundation models are available,
such as text, audio or video, where the application of \gls{ours} corresponds to future work.

Lastly,
we demonstrated in \cref{sec:methods:audit} how to apply \gls{ours} to audit models.
Hereby,
first steps have been taken to quantify alignment to expectation, \eg, quantify the relative share of neurons that align to valid features. 
For future work,
further quantification measures could be developed, compared and the correlation to actual misbehaviour tested.

\end{appendix}

\end{document}